%% file: main.tex
\documentclass[final]{l4dc2025}

\usepackage[utf8]{inputenc} 
\usepackage[T1]{fontenc}    
\usepackage{hyperref}       
\usepackage{url}            
\usepackage{booktabs}       
\usepackage{nicefrac}       
\usepackage{microtype}      
\usepackage{xcolor}         

\usepackage{graphicx}
\usepackage{amsmath}
\usepackage{pifont}
\usepackage{bm}
\usepackage{multirow}
\usepackage{algorithm}
\usepackage{algorithmic}
\usepackage{amssymb}
\usepackage{amsfonts}       
\usepackage{pbox}
\usepackage{cleveref}
\usepackage{subcaption}
\usepackage{caption}
\usepackage{rotating}
\usepackage{colortbl}
\usepackage{wrapfig}
\usepackage{adjustbox}
\usepackage{comment}

\usepackage{times}
\usepackage{array}
\usepackage{enumitem}

\input{math_commands}

\newcommand{\xmark}{\ding{55}}

\definecolor{navyblue}{rgb}{0.0, 0.0, 0.5}

\newtheorem{asm}{Assumption}

\Crefname{asm}{Assumption}{Assumption}
\newcommand{\mathbbm}[1]{\text{\usefont{U}{bbm}{m}{n}#1}}

\begin{document}

\title[]{Learn With Imagination: \\Safe Set Guided State-wise Constrained Policy Optimization}

\author{%
 \Name{Yifan Sun*} \Email{yifansu2@andrew.cmu.edu}\\
 \Name{Feihan Li*} \Email{feihanl@andrew.cmu.edu}\\
 \Name{Weiye Zhao*} \Email{weiyezha@andrew.cmu.edu} \\
 \Name{Rui Chen} \Email{ruic3@andrew.cmu.edu} \\
 \Name{Tianhao Wei} \Email{twei2@andrew.cmu.edu} \\
 \Name{Changliu Liu} \Email{cliu6@andrew.cmu.edu} \\
 \addr Robotics Institute, Carnegie Mellon University, Pittsburgh, PA 15213, USA \thanks{ * Equal Contribution}
}

\maketitle

\input{0_abstruct}
\input{1_introduction_v1}

\input{2_problem_formulation}

\input{3_prior_works}
\input{4_safety_index_guided_SCPO}

\input{5_practical_implementation}
\input{6_theoretical_results}
\input{7_experiments}
\input{8_conclusion}
\input{9_acknowledgement}

\bibliography{main.bib}
\clearpage

\appendix

\input{appendix_algo}
\input{appendix_proof}

\input{appendix_experiments}

\end{document}

%% file: math_commands.tex
\newcommand{\cM}{\mathcal{M}}

\newcommand{\cS}{{\mathcal{S}}}

\newcommand{\EE}{\mathbb{E}}

\newcommand{\argmax}{\mathop{\mathbf{argmax}}}
\newcommand{\maximize}{\mathop{\mathbf{max}}}

\newcommand{\maximizewrt}[1]{\mathop{\underset{#1}{\maximize}}}

\newcommand{\argmaxwrt}[1]{\mathop{\underset{#1}{\argmax}}}

\newcommand{\st}{\mathop{\mathbf{s.t.}}}

\newcommand{\hs}{\hat{s}}

%% file: 0_abstruct.tex
\begin{abstract}

    Deep reinforcement learning (RL) has achieved remarkable success across various control tasks. However, its reliance on exploration through trial and error often results in safety violations during training. To mitigate this, safety filters are commonly employed to correct unsafe actions generated by the RL policy. 
    Yet, a key challenge remains: how to enable safety-filter-guided learning to produce a policy that remains optimally safe even after the filter is removed.
    In this paper, we propose Safe Set Guided State-wise Constrained Policy Optimization (S-3PO) — a novel algorithm designed to generate optimal safe RL policies with zero training violations while maintaining safety during evaluation even without any safety filter. S-3PO integrates a safety-oriented monitor operating on black-box dynamics to ensure safe exploration and introduces an imaginary cost mechanism that guides the safe RL agent toward optimal behavior under safety constraints. This imaginary cost inherits the interpretability of the safety filter while outperforming conventional imitation-based cost designs.
    Equipped with state-of-the-art components, S-3PO demonstrates superior performance on high-dimensional robotic control tasks, effectively handling expected state-wise constraints and ensuring safety throughout the training process.

\end{abstract}

%% file: 1_introduction_v1.tex
\section{Introduction}
Safe Reinforcement Learning (safe RL) has emerged as a powerful approach in domains such as games and robotic control, where ensuring safety during or after training is critical. While objective-based methods aim to optimize reward, they often lack formal guarantees on safety performance~\cite{bohez2019value}. To address this, many approaches enforce hard constraints~\cite{bouvier2024policedrllearningclosedloop, bouvier2024learningprovablysatisfyhigh}; however, these methods are typically effective only in low-dimensional systems. More recent advances~\cite{zhao2023scpo, zhao2024absolutestatewiseconstrainedpolicy} leverage trust-region techniques combined with Maximum Markov Decision Processes (MMDP) to enforce simultaneous improvement of worst-case performance and adherence to cost constraints.

Despite these developments, RL-based methods fundamentally depend on trial-and-error exploration, making it difficult to guarantee safety throughout the training process. A common strategy to mitigate this issue is the use of safety filters~\cite{alshiekh2018safe}, which is designed to correct unsafe actions generated by the RL policy. These safety filters are often constructed using principles from safe control theories~\cite{shao2021reachability}, where energy function-based methods remain the most widely adopted approach~\cite{khatib1986real, ames2014control, liu2014control, gracia2013reactive, wei2019safe}. 
Although safety filters can ensure safety during training by overriding unsafe actions, they also prevent the policy from learning how to avoid unsafe behaviors on its own. This creates a fundamental dilemma: \textit{how can an agent learn to avoid unsafe scenarios if it is always shielded from experiencing them}?

To enable the policy to learn from the safety filter and generate safe actions by itself, rather than only being protected by the safety filter,  
~\cite{cheng2019end} use Gaussian Process (GP) models to estimate unknown system dynamics and construct a safety filter that implicitly guides the policy updates based on the history of safety filter interventions.  
To explicitly imitate the effect of the safety filter at each step, other works have considered using reward penalties to learn from safety filters. ~\cite{krasowski2022provably} introduce a constant penalty when the safety filter is activated. Yet, this approach does not account for how unsafe the proposed action was. In contrast, ~\cite{wabersich2021predictive} penalize the reward with magnitude of the action correction applied by the safety filter. However, the magnitude of the correction alone may not accurately capture the true impact of the action on system safety, and the reward penalty can only impose a soft constraint during the learning process.

\begin{figure}[t]
\vspace{-30pt}
\centering
\begin{adjustbox}{center}
\includegraphics[width=0.7\columnwidth]{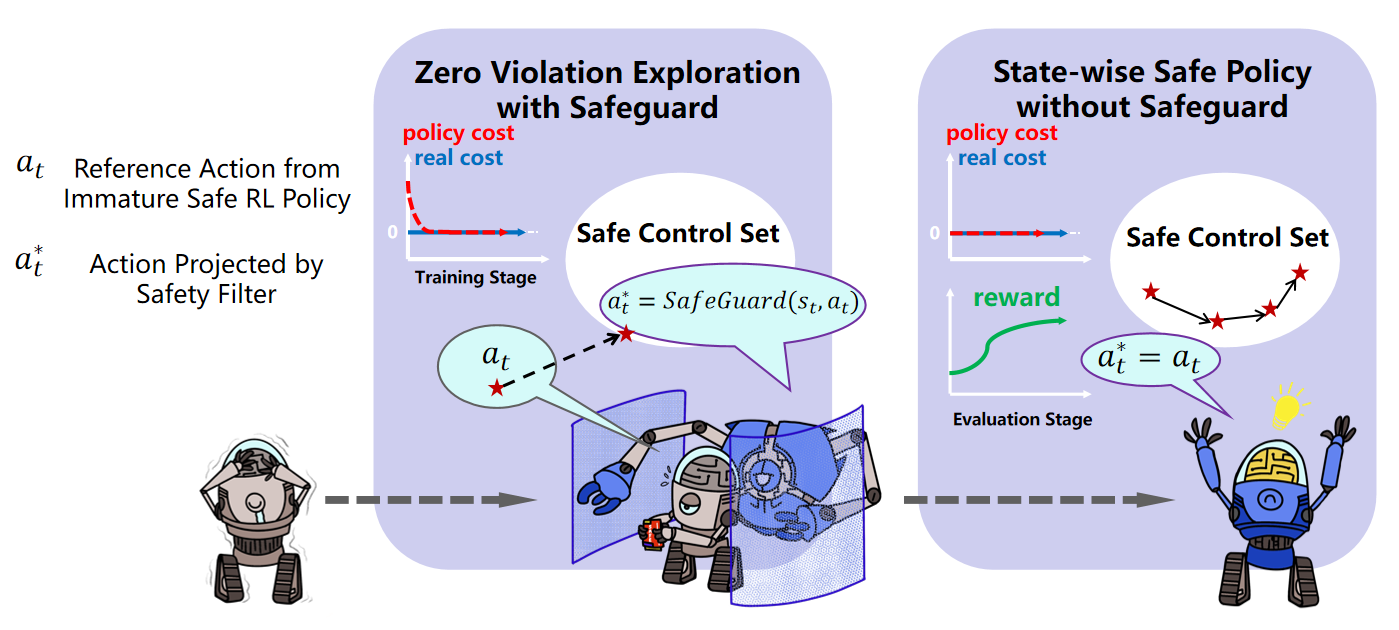}
\end{adjustbox}
\caption{Overview of the principles of the S-3PO algorithm.}
\label{fig:1}
\vspace{-20pt}
\end{figure}

These limitations highlight a critical gap: \textbf{to enable a policy to maintain safety after the safety filter is removed, it is essential to introduce a cost term that explicitly measures how the policy’s actions influence the system's safety level if the safe filter is removed}.
To this end, we introduce 
\textit{\textbf{S}afe \textbf{S}et Guided \textbf{S}tate-wise Constrained \textbf{P}olicy \textbf{O}ptimization} (S-3PO). S-3PO safeguards the exploration of immature policies through a black-box safe control mechanism and formulates a novel constrained optimization framework where RL learns an optimal safe policy by constraining \textit{imaginary safety violations}—violations that would have occurred without the filter.

%% file: 2_problem_formulation.tex
\section{Problem Formulation}
\subsection{Assumptions}
\label{subsec: probform}
\paragraph*{Dynamics}

We consider a robot system described by its state $s_t \in \mathcal{S}\subset \mathbb{R}^{n_s}$ at time step $t$, with $n_s$ denoting the dimension of the state space $\mathcal{S}$, and its action input $a_t \in \mathcal{A}\subset \mathbb{R}^{n_a}$ at time step $t$, where $n_a$ represents the dimension of the control space $\mathcal{A}$. The system dynamics are defined as follows:
\begin{equation}\label{eq:dynamics fn}
\begin{split}
    &s_{t+1} = f(s_t,a_t), \\
\end{split}
\end{equation}
where $f: \mathcal{S} \times \mathcal{A} \rightarrow \mathcal{S}$ is a deterministic function that maps the current robot state and control to the robot's state in the next time step.

To maintain simplicity, our approach focuses on deterministic dynamics, although the proposed method can be easily extended to stochastic dynamics~\cite{zhao2021model}. Additionally, we assume the access to the dynamics model $f$ is only in the training phase and restricted to an black-box form, such as an implicit digital twin simulator or a deep neural network model~\cite{zhao2021model}. We also assume there is no model mismatch. Model mismatch can be addressed by robust safe control~\cite{wei2022persistently} and is left for future work. 
Post training, the knowledge of the dynamics model is concealed—a benefit of using "imaginary cost"—aligning with practical scenarios where digital twins of real-world environments are too costly to access during deployment.
Based on these, our core target is to figure out how can safety-filter-guided learning be used to produce a policy that remains optimally safe even without the filter.

\paragraph{Markov Decision Process}

In this research, our primary focus lies in ensuring safety for episodic tasks, which falls within the purview of finite-horizon Markov Decision Processes (MDP). An MDP is defined by a tuple $(\mathcal{S}, \mathcal{A}, \gamma, R, P, \mu)$. The reward function is denoted by $R: \mathcal{S} \times \mathcal{A} \times \mathcal{S} \rightarrow \mathbb{R}$, the discount factor by $0 \leq \gamma < 1$, the initial state distribution by $\mu: \mathcal{S} \rightarrow \mathbb{R}$, and the transition probability function by $P: \mathcal{S} \times \mathcal{A} \times \mathcal{S} \rightarrow \mathbb{R}$.

The transition probability $P(s'|s,a)$ represents the likelihood of transitioning to state $s'$ when the previous state was $s$, and the agent executed action $a$ at state $s$. This paper assumes deterministic dynamics, implying that $P(s_{t+1}|s_t,a_t) = 1$ when $s_{t+1} = f(s_t,a_t)$. We denote the set of all stationary policies as $\Pi$, and we further denote $\pi_\theta$ as a policy parameterized by the parameter $\theta$.

In the context of an MDP, our ultimate objective is to learn a policy $\pi$ that maximizes a performance measure $\mathcal{J}(\pi)$, computed via the discounted sum of rewards, as follows:
\begin{align}
\label{eq: reward function}
    \mathcal{J}(\pi) = \mathbb{E}_{\tau \sim \pi}\left[\sum_{t=0}^H \gamma^t R(s_t, a_t, s_{t+1})\right],
\end{align}
where $H \in \mathbb{N}$ denotes the horizon, $\tau = [s_0, a_0, s_1, \cdots]$, and $\tau \sim \pi$ indicates that the distribution over trajectories depends on $\pi$, i.e., $s_0 \sim \mu$, $a_t \sim \pi(\cdot | s_t)$, and $s_{t+1} \sim P(\cdot|s_t, a_t)$.

\paragraph*{Safety Specification}

The safety specification requires that the system state remains within a closed subset in the state space, denoted as the ``safe set'' $\mathcal{S}_S$. This safe set is defined by the zero-sublevel set of a continuous and piecewise smooth function $\phi_0: \mathbb{R}^{n_s} \rightarrow \mathbb{R}$, where $\mathcal{S}_S = \{s \mid \phi_0(s) \leq 0\}$, usually specified by users. 
For instance, for collision avoidance, $\phi_0$ can be specified as $d_{min}-d$ where $d$ is the closest distance between the robot and environmental obstacles and $d_{min}$ is the distance margin.

\subsection{Problem}


We are interested in the safety imperative of averting collisions for mobile robots navigating 2D planes.
We aim to persistently satisfy safety specifications at every time step while solving MDP, following the intuition of State-wise Constrained Markov Decision Process (SCMDP)~\cite{zhao2023state}. Formally,
the set of feasible stationary policies for SCMDP is defined as
\begin{align}
\label{eq:scmdp}
    \bar{\Pi}_{C} = \{ \pi \in \Pi \big | ~\forall s_t \sim \tau, s_t \in \mathcal{S}_S\},
\end{align}
where $\tau \sim \pi$. Then, the objective for SCMDP is to find a feasible stationary policy from $\bar{\Pi}_{C}$ that maximizes the performance measure. Formally, 
\begin{align}
\label{eq: fundamental problem}
    \max_\theta \mathcal{J}(\pi_\theta), \text{ s.t. } \pi_\theta \in \bar{\Pi}_{C}.
\end{align}

\paragraph{State-wise Safe Policy with Zero Violation Training}
The primary focus of this paper centers on solving \eqref{eq: fundamental problem}, i.e., ensuring no safety violation during the training process, while achieving convergence of the policy to the optimal solution of \eqref{eq: fundamental problem}. 


\section{Preliminary}
\subsection{Implicit Safe Set Algorithm}

As a deterministic safety filter, Implicit Safe Set Algorithm (ISSA) \cite{zhao2021model, zhao2024discreteISSA} ensures the persistent satisfaction of safety specifications for systems with black-box dynamics (e.g., digital twins or neural networks) through energy function-based optimization. 
Leveraging energy function $\phi = \phi_0^* + k_1\dot{\phi}_0 + \cdots + k_n\phi_0^{(n)}$ and theoretical results from SSA \cite{liu2014control}, ISSA synthesizes a safety index to guarantee that the safe control set $\mathcal{A}_S(s):=\{a\in\mathcal{A}\mid \dot\phi \leq -\eta(\phi)\}$ is nonempty. And then the set $\bar{\mathcal{S}} :=\{s\mid\phi(s)\leq 0\}\cap \{s\mid \phi_0(s)\leq 0\}$ is forward invariant under \Cref{asm:ruleassumption}.
Here $\eta(\phi)$ is designed to be a positive constant when $\phi\geq0$ and $-\infty$ when $\phi<0$. 

\begin{asm}
\label{asm:ruleassumption}
The system~\eqref{eq:dynamics fn} is a second-order mobile platform that avoids 2D obstacles. 
1) The state space is bounded, and the relative acceleration $w$ and angular velocity $z$ to the obstacle are bounded and both can achieve zeros, i.e., $w \in [w_{min}, w_{max}]$ for $w_{min} \leq 0 \leq w_{max}$ and $z \in [z_{min}, z_{max}]$ for $z_{min} \leq 0\leq z_{max}$; 
2) For all possible values of $z$ and $w$, there always exists a control $a$ to realize such $z$ and $w$;
3) The discrete-time system time step $dt \to 0$;
4) At any given time, there can at most be one obstacle becoming safety critical (Sparse Obstacle Environment).
\end{asm}
\begin{remark}
    The bounds in the first assumption will be directly used to synthesize $\phi$. The second assumption enables us to turn the question on whether there exists a feasible control in $\mathcal{A}_S^D$ to the question on whether there exists $z$ and $w$ to decrease $\phi$. The third assumption ensures that the discrete time approximation error is small. The last assumption ensures that the safety index design rule is applicable to multiple moving obstacles.
\end{remark}

The deterministic ISSA could be used as a safety filter in the discrete-time MDP and is treated as part of the environment during training. We define the discrete-time safe control set as $\mathcal{A}_S^D(s):=\{a\in \mathcal{A}\mid \phi(f(s, a)) \leq \max\{\phi(s)-\eta, 0\} \}$.
The ISSA mechanism ensures safety by projecting the nominal control action $a_t$, proposed by the RL policy $\pi_\theta$, onto the safe control set $\mathcal{A}_S^D(s_t)$ by solving the optimization problem:
\begin{equation}\label{eq:adamba_discrete}
\begin{split}
    &\min_{a_t^*\in\mathcal{A}} \| a_t^* - a_t\|^2\\
    & \text{s.t. } \phi(f(s_t, a_t)) \leq \max\{\phi(s_t)-\eta, 0\}.
\end{split}
\end{equation}

\subsection{State-wise Constrained Policy Optimization}

Safe RL algorithms under the framework of Constrained Markov Decision Process (CMDP) do not consider state-wise constraints. To address this gap, State-wise Constrained Policy Optimization (SCPO) was proposed \cite{zhao2023scpo} to provide guarantees for state-wise constraint satisfaction in expectation, which is under the framework of State-wise CMDP (SCMDP). To achieve this, SCPO directly constrain the expected maximum state-wise cost along the trajectory. And they introduced Maximum MDP (MMDP). In this setup, a running maximum cost value is associated with each state, and a non-discounted finite MDP is utilized to track and accumulate non-negative increments in cost. The format of MMDP will be introduced in \Cref{sec: 4}.

%% file: 4_safety_index_guided_SCPO.tex
\section{Safety Index Guided State-wise Constrained Policy Optimization}
\label{sec: 4}

\begin{figure}[t]
\vspace{-30pt}
\centering
\begin{adjustbox}{center}
\includegraphics[width=1.0\columnwidth]{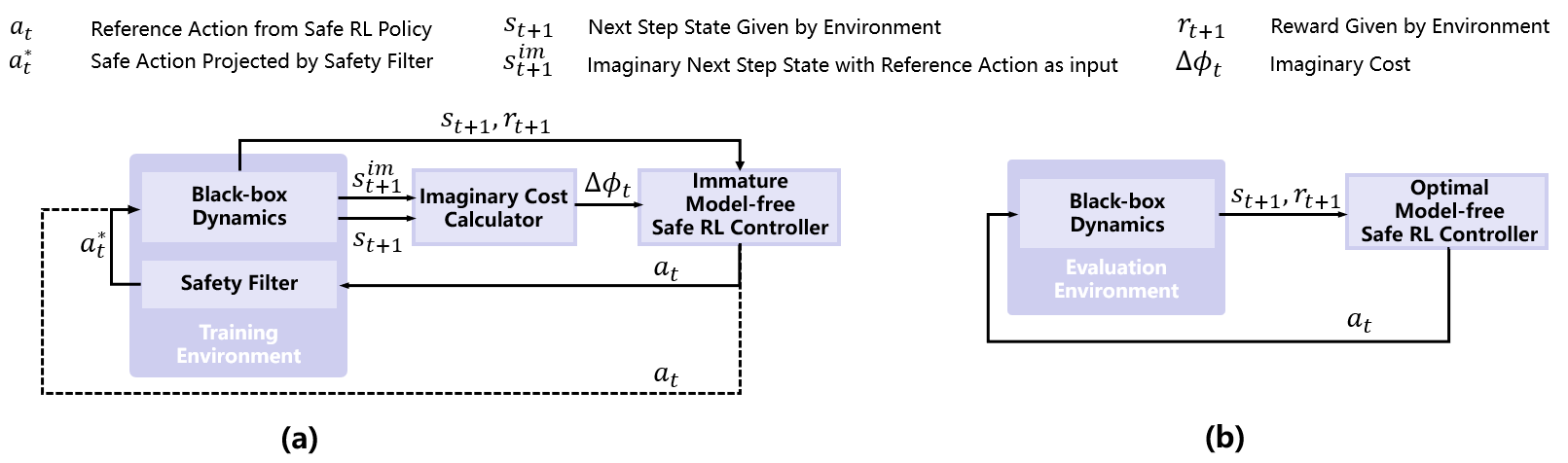}
\end{adjustbox}
\caption{(a) S-3PO pipeline during training. (b) S-3PO pipeline during evaluation.}
\label{fig:framwork}
\end{figure}

The core idea of S-3PO is to enforce zero safety violations during training by projecting unsafe actions to the safe set, and then constraining the "imaginary" safety violations to ensure convergence of the policy to an optimal safe policy. As shown in \Cref{fig:framwork}, the safety filter is part of the training environment, while the evaluation environment does not have the safety filter.

\subsection{Learn with Imaginary Cost}

\paragraph{Zero Violation Exploration}
To ensure zero violation exploration, we adopt ISSA as the safety filter and safeguard nominal control via solving \eqref{eq:adamba_discrete} at every time step during policy training. 
With the safety index synthesis rule in \cite{zhao2021model}, ISSA is guaranteed to find a feasible solution of \eqref{eq:adamba_discrete}, making the system forward invariance in the set $\mathcal{\bar S}$. It is worth mentioning that any energy-function-based method that ensures forward invariance could be used as the safety filter, and the scalar energy function could be used to evaluate the imaginary cost. The integration with other energy-function-based methods will be left for future work. 

\paragraph{Learning Safety Measures Safely}
While eliminating safety violations during training is beneficial, it also poses challenges for RL policy training, as RL relies on a trial-and-error process. To address this, our key insight is that instead of directly encountering unsafe states ($s \notin \mathcal{S}_S$), the policy can leverage an "imaginary cost" to learn about unsafe scenarios without actually experiencing them. 

\paragraph{Imaginary Cost
}
\label{obs: delta phi}
Define "imaginary cost" as $\Delta \phi_t = \Delta \phi(s_t, a_t, s_{t+1}) \doteq \phi(f(s_t, a_t)) - \phi(f(s_t, a_t^*))$, i.e. the degree of required correction to safeguard $a_t$. Here $a_t^*$ is the projected action by ISSA. Therefore, $\Delta \phi_t$ can be treated as an imagination on how unsafe the reference action would be, where $\Delta \phi_t \leq 0$ means $a_t \in \mathcal{A}_S^D(s_t)$.

Following the definition, 
\Cref{eq: fundamental problem} can be translated to:
\begin{align}
\label{eq: new problem}
    \max_\theta \mathcal{J}(\pi_\theta), \text{ s.t. } \pi_\theta \in \{ \pi \in \Pi \big | ~\forall \Delta \phi_t \sim \tau, \Delta \phi_t \leq 0 \}.
\end{align}

\begin{remark}
    Policies satisfying \eqref{eq: new problem} ensure there is no imaginary safety violation in expectation for any possible $a_t$, making $\pi_\theta$ a safe policy as required by \eqref{eq: fundamental problem}, to be proved by lemma \ref{prop: s-3po safety convergence}.
\end{remark}

\subsection{Transfrom State-wise Constraint into Maximum Constraint}
For \eqref{eq: new problem}, each state-action transition pair corresponds to a constraint, which is intractable to solve. Inspired by \cite{zhao2023state}, we constrain the expected maximum state-wise $\Delta \phi$ along the trajectory instead of individual state-action transition $\Delta \phi$. 

Next, by treating $\Delta \phi_t$ as an ``imaginary'' cost, we define a MMDP~\cite{zhao2023state} by introducing (i) an up-to-now maximum state-wise cost $M$ within $\cM \subset \mathbb{R}$, and (ii) a "cost increment" function $D$, where $D:(\mathcal{S}, \cM) \times \mathcal{A} \times (\mathcal{S}, \cM) \rightarrow [0, \mathbb{R}^+]$ maps the augmented state-action transition tuple to non-negative cost increments. 
We define the augmented state $\hs = (s, M) \in (\mathcal{S},\cM) \doteq \hat{\cS}$, where $\hat{\cS}$ is the augmented state space.
Formally, 
\begin{align}
    D\big(\hs_t, a_t, \hs_{t+1}\big) = \max\{\Delta \phi(s_t, a_t, s_{t+1}) - M, 0\}.
\end{align}
By setting $D\big(\hs_0, a_0, \hs_{1}\big) = \Delta \phi(s_0,a_0,s_1)$, we have $M = \sum_{k=0}^{t-1} D\big(\hs_k, a_k, \hs_{k+1}\big)$ for $t \geq 1 $. Hence, we define \textit{expected maximum state-wise cost} (or $D$-return) for S-3PO policy $\pi$: 
\begin{align}
\label{eq: sum of max cost}
    \mathcal{J}_{D}(\pi) =   \mathbb{E}_{\tau \thicksim \pi} \Bigg[
    \sum_{t=0}^{H} D\big(\hs_t, a_t, \hs_{t+1}\big)\Bigg].
\end{align}
With \eqref{eq: sum of max cost}, \eqref{eq: new problem} can be rewritten as:
\begin{align}
    \label{eq: fundamental problem s3po}
    \maximizewrt{\pi} \mathcal{J}(\pi),~\st \mathcal{J}_{D}(\pi) \leq 0,
\end{align}
where $\mathcal{J}(\pi) = \mathbb{E}_{\tau \sim \pi}\left[\sum_{t=0}^H \gamma^t R(\hs_t, a_t, \hs_{t+1})\right]$ and $R(\hs,a,\hs')\doteq R(s, a, s')$.
With $R(\tau)$ being the discounted return of a trajectory, we define the on-policy value function as $V^\pi(\hs) \doteq \mathbb{E}_{\tau \sim \pi}[R(\tau) | \hs_0 = \hs]$, the on-policy action-value function as $Q^\pi(\hs,a) \doteq \mathbb{E}_{\tau \sim \pi}[R(\tau) | \hs_0 = \hs, a_0=a]$, and the advantage function as $A^\pi(\hs,a) \doteq Q^\pi(\hs, a) - V^\pi(\hs)$. 

Lastly, we define on-policy value functions, action-value functions, and advantage functions for the cost increments in analogy to $V^\pi$, $Q^\pi$, and $A^\pi$, with $D$ replacing $R$, respectively. 
We denote those by $V_{D}^\pi$, $Q_{D}^\pi$ and $A_{D}^\pi$.

\begin{remark}
    \Cref{eq: new problem} is difficult to solve since there are as many constraints as the size of trajectory $\tau$.
    With \eqref{eq: fundamental problem s3po}, we turn all constraints in \eqref{eq: new problem} into only a single constraint on the maximal $\Delta\phi$ along the trajectory, yielding a practically solvable problem.
\end{remark}

\subsection{S-3PO}

To solve \eqref{eq: fundamental problem s3po}, we propose S-3PO under the framework of trust region optimization methods~\cite{schulman2015trust}. S-3PO uses KL divergence distance to restrict the policy search in \eqref{eq: fundamental problem s3po} within a trust region around the most recent policy $\pi_k$. Moreover, S-3PO uses surrogate functions for the objective and constraints, which can be easily estimated from sample trajectories by $\pi_k$. Mathematically, S-3PO updates policy via solving the following optimization:
\begin{align}
\label{eq: s3po optimization final}
    & \pi_{k+1} = \argmaxwrt{\pi \in \Pi_\theta} \underset{\substack{\hs \sim d^{\pi_k} \\ a\sim \pi}}{\mathbb{E}} [A^{\pi_k}(\hs,a)] \\ \nonumber 
    & \st  ~~ \mathbb{E}_{\hs \sim \bar d^{\pi_k}}[\mathcal{D}_{KL}(\pi \| \pi_k)[\hs]]\leq \delta, \\ \nonumber
    & ~~~~\mathcal{J}_{D}(\pi_k) + \underset{\substack{\hs \sim \bar d^{\pi_k} \\ a\sim {\pi}}}{\mathbb{E}}\Bigg[ A^{\pi_k}_{D}(\hs,a) \Bigg] + 2(H+1)\epsilon_{D}^{\pi} \sqrt{\frac{1}{2} \delta}  \leq 0.
\end{align}
where $\mathcal{D}_{KL}(\pi' \| \pi)[\hs]$ is KL divergence between two policy $(\pi', \pi)$ at state $\hat s$,
the set $\{\pi \in \Pi_\theta ~: \\ ~ \mathbb{E}_{\hs \sim \bar d^{\pi_k}}[\mathcal{D}_{KL}(\pi \| \pi_k)[\hs]] \leq \delta\}$ is called \textit{trust region}, $d^{\pi_k} \doteq (1-\gamma)\sum_{t=0}^H\gamma^t P(\hs_t=\hs|{\pi_k})$, 
$\bar d^{\pi_k} \doteq \sum_{t=0}^H P(\hs_t=\hs|{\pi_k})$
and $\epsilon^{{\pi}}_{D} \doteq \mathbf{max}_{\hs}|\EE_{a\sim {\pi}}[A^{\pi_k}_{D}(\hs,a)]|$.

\begin{remark}
    Despite the complex forms, the objective and constraints in \eqref{eq: s3po optimization final} can be interpreted in two steps.
    First, maximizing the objective (expected reward advantage) within the trust region (marked by the KL divergence constraint) theoretically guarantees the worst performance degradation.
    Second, $J_D(\pi)$ can not be computed at step $k+1$ since the state $s_{k+1}$ is inaccessible, thus we leverage a surrogate function to upper bound the $J_D(\pi)$ to guarantee the worst-case ``imagionary'' cost is non-positive at all steps as in \eqref{eq: new problem}. 
\end{remark}


\subsection{Practical Implementation}
\label{sec: practical}
The pseudocode of S-3PO is give as \cref{alg:s3po_main}. Here we summarize two techniques that helps with S-3PO's practical performance. (i) \textbf{Weighted loss for cost value targets}: A critical step in S-3PO involves fitting the cost increment value function, $V_{D}^\pi(\hs_t)$, which represents the maximum future cost increment relative to the highest state-wise cost observed so far. This function follows a non-increasing staircase pattern along the trajectory. Thus, we adopt a weighted loss function, $L_{weight}$, to penalize predictions that violate the non-increasing property: $L_{weight} = L(\hat{y}_t - y_t) * (1 + w * \mathbbm{1}[(\hat{y}_t - y_{t - 1}) > 0])$, where $L$ denotes Mean Squared Error, $\hat{y}_t$ is the prediction, ${y}_t$ is the fitting target and ${w}$ is the penalty weight. (ii) \textbf{Line Search scheduling}: Constraints in \eqref{eq: s3po optimization final} might become infeasible due to approximation errors. In this case, we perform a recovery update, enforcing the cost advantage $A_D^\pi$ to decrease from early training steps $k_\mathrm{safe}$ while focusing on reward improvements of $A^\pi$ towards the end of training, prioritizing safety first and reward performance later. Check \Cref{sec: practical detail} for furthur details.

%% file: 6_theoretical_results.tex
\section{Theoretical Results}
\label{sec: theory}

In this section, we first present the lemma to show the equivalence between constraining the \textit{imaginary cost} and constraining the safety violation. Then we present the main conclusion for S3PO.


\begin{lemma}[Safety Equivalence under Imaginary Cost]
\label{prop: s-3po safety convergence}
For a given policy $\pi$ and any initially safe state $s_0$ ($\phi(s_0) \leq 0$), the following two conditions are equivalent: 1) the corresponding trajectory in the training environment has zero imaginary cost $\max_{t}\Delta \phi_t\leq 0$; 2) the corresponding trajectory in the evaluation environment has zero safety violation $\max_t{\phi_t}\leq 0$. And if either condition holds, the two trajectories are the same.  
\end{lemma}

\begin{theorem}[Safety and Optimality of S-3PO]
S-3PO will converge in the training environment to a policy $\pi$ with no imaginary cost in expectation, and bounded worst case reward performance. In particular, if $\pi_{k}$ and $\pi_{k+1}$ are related by applying S-3PO \eqref{eq: s3po optimization final}, then with $\epsilon^{\pi_{k+1}} \doteq \mathbf{max}_{\hs}|\EE_{a\sim\pi_{k+1}}[A^{\pi_k}(\hs,a)]|$,  the performance of $\pi_{k+1}$ in the training environment satisfies:
\begin{align}
\mathcal{J}(\pi_{k+1}) - \mathcal{J}(\pi_k) \geq - \frac{    \sqrt{2\delta}\gamma \epsilon^{\pi_{k+1}}}{1-\gamma}. \nonumber
\end{align}
\end{theorem}
Since the training environment is deterministic and all the assumptions in the theoretical results for SCPO are satisfied, the proof for the theorem directly follows from Proposition 2 from SCPO~\cite{zhao2023scpo}.  

By lemma \ref{prop: s-3po safety convergence}, no imaginary cost in the training environment implies no safety violation in the evaluation environment. The theorem then implies that the converged policy could achieve zero safety violation in expectation in the evaluation environment. Nevertheless, formally establishing their equivalence in the probability space (e.g., in expectation) will be left for future work. 




%% file: 7_experiments.tex
\section{Experiments}

In our experiments, we aim to answer:
\textbf{Q1:} Does S-3PO achieve zero-violation during the training? 
\textbf{Q2:} How does S-3PO without safeguard compare with other advanced safe RL methods? 
\textbf{Q3:} Does S-3PO learn to act without safeguard? 
\textbf{Q4:} How does weighted loss trick impact the performance of S-3PO? 
\textbf{Q5:} Is ``imaginary'' cost necessary to make the RL policy learn to achieve zero violation by itself?
\textbf{Q6:} How does S-3PO scale to high dimensional robots?
\subsection{Experiments Setup}
To answer these questions, we conducted experiments on the safe reinforcement learning benchmark GUARD~\cite{zhao2023guard} which is based on Mujoco and Gym interface.

\textbf{Environment Setting}  We design experimental environments with different task types, constraint types, constraint numbers, and constraint sizes. We name these environments as \{Robot\}\_\{Constraint Number\}\{Constraint Type\}. All of the environments are based on \texttt{Goal} where the robot must navigate to a goal. Three different robots that can be categorized into two types are included in our experiments: (i) Wheel Robot: \textbf{Point}: (\cref{fig: Point}) A point robot ($\mathcal{A} \subseteq \mathbb{R}^{2}$) that maintains seamless interaction with the environment. (ii) Link Robot: (a) \textbf{Swimmer}: (\cref{fig: Swimmer}) A three-link robot ($\mathcal{A} \subseteq \mathbb{R}^{2}$) that interacts intermittently with the surroundings. (b) \textbf{Ant}: (\cref{fig: Ant}) A quadrupedal robot ($\mathcal{A} \subseteq \mathbb{R}^{8}$).
Two different types of constraints are considered. (i) \textbf{Hazard}: (\cref{fig: hazard}) Trespassable circles on the ground. (ii) \textbf{Pillar}: (\cref{fig: pillar}) Fixed obstacles. All tasks are trained over 200 epochs, with each epoch consisting of 30,000 steps.
More details about the experiments are discussed in \Cref{appendix:Environment Settings}.

\begin{figure*}[t]
    \centering
    \subfigure[Point\_1Hazard]{
        \centering
        \parbox{0.23\textwidth}{
            \includegraphics[width=0.24\textwidth]{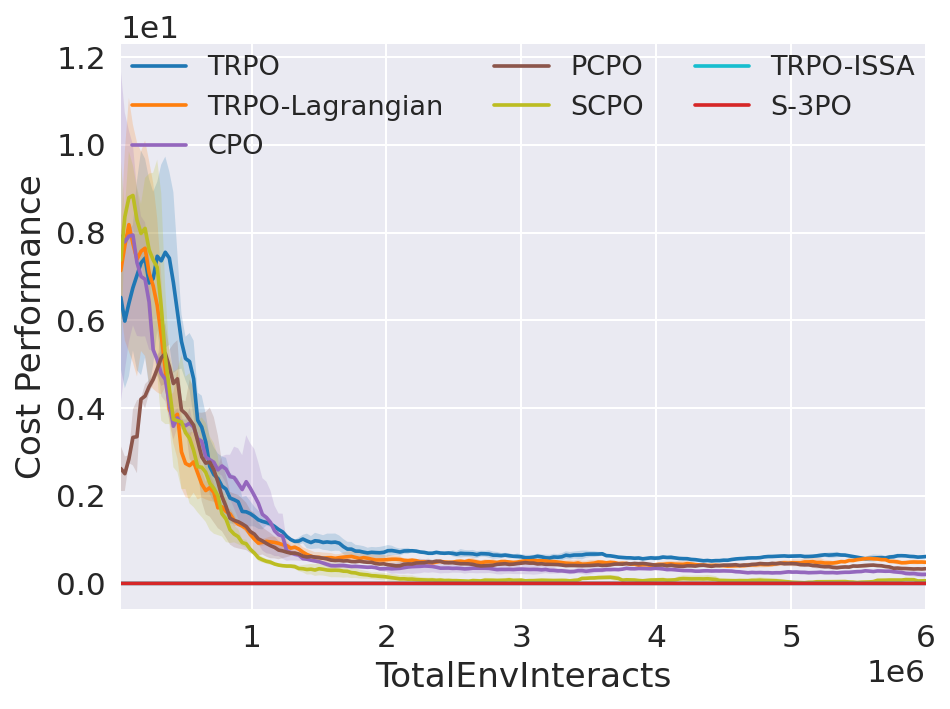} \\
            \includegraphics[width=0.24\textwidth]{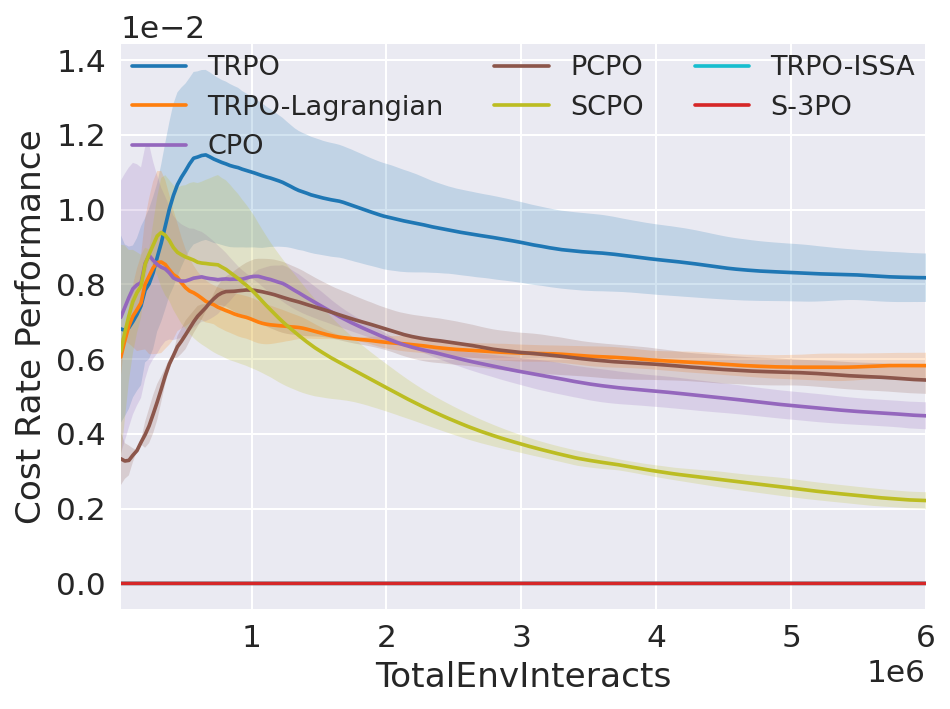}
        }
    }
    \subfigure[Point\_8Hazard]{
        \centering
        \parbox{0.23\textwidth}{
            \includegraphics[width=0.24\textwidth]{fig/goal1_Cost_Performance_Train.png} \\
            \includegraphics[width=0.24\textwidth]{fig/goal1_Cost_Rate_Performance_Train.png}
        }
    }
    \subfigure[Point\_1Pillar]{
        \centering
        \parbox{0.23\textwidth}{
            \includegraphics[width=0.24\textwidth]{fig/goal1_Cost_Performance_Train.png} \\
            \includegraphics[width=0.24\textwidth]{fig/goal1_Cost_Rate_Performance_Train.png}
        }
    }
    \subfigure[Swimmer\_1Hazard]{
        \centering
        \parbox{0.23\textwidth}{
            \includegraphics[width=0.24\textwidth]{fig/goal1_Cost_Performance_Train.png} \\
            \includegraphics[width=0.24\textwidth]{fig/goal1_Cost_Rate_Performance_Train.png}
        }
    }    
    \vspace{-10pt}
    \caption{Illustration of training-time cost performance from four representative test suites.}
    \label{fig:zero violation performance}
    \vspace{-10pt}
\end{figure*}

\textbf{Comparison Group}  The methods in the comparison group include: (i) unconstrained RL algorithm TRPO~\citep{schulman2015trust} and TRPO-ISSA.  (ii) end-to-end constrained safe RL algorithms CPO~\citep{achiam2017cpo}, TRPO-Lagrangian~\citep{bohez2019value}, PCPO~\citep{yang2020projection}, SCPO~\citep{zhao2023scpo}. (iii) 
We select TRPO as our baseline method since it already has safety-constrained derivatives that can be tested off-the-shelf.
The full list of parameters of all methods compared can be found in \Cref{appendix:Policy Settings}.

\textbf{Metrics}  For comparison, we evaluate algorithm performance based on (i) reward performance, (ii) average episode cost, and (iii) cost rate. More details are provided in \Cref{appendix:Metrics Comparison}. 
We set the limit of cost to 0 for all safe RL algorithms since no violation of the constraints is allowed. 

\begin{figure*}[t]
    \centering
    \subfigure[Point\_1Hazard]{
        \centering
        \parbox{0.225\textwidth}{
            \includegraphics[width=0.24\textwidth]{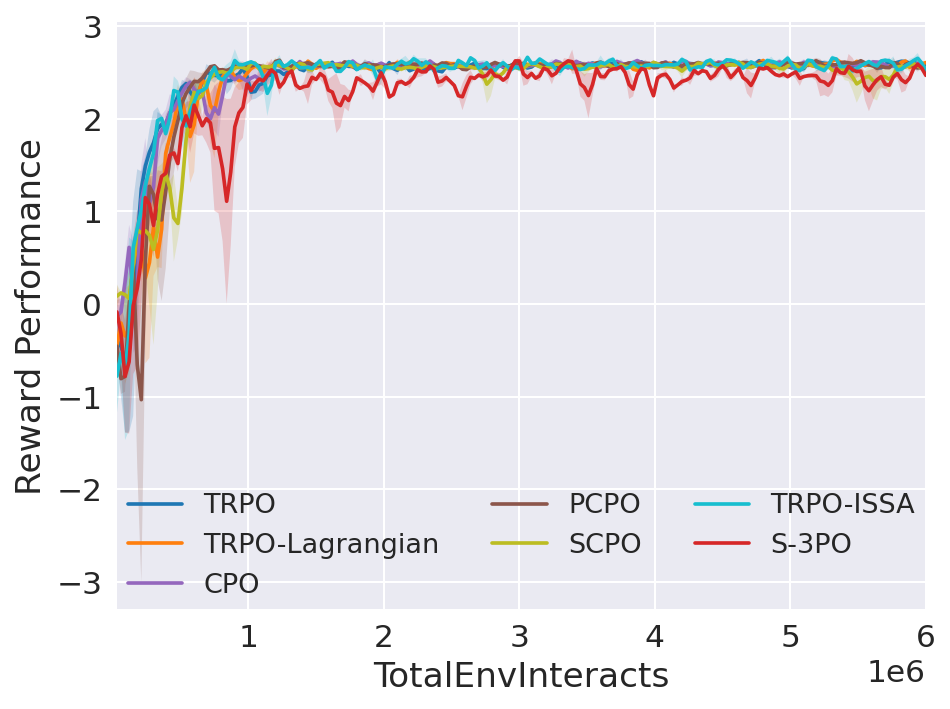} \\
            \includegraphics[width=0.24\textwidth]{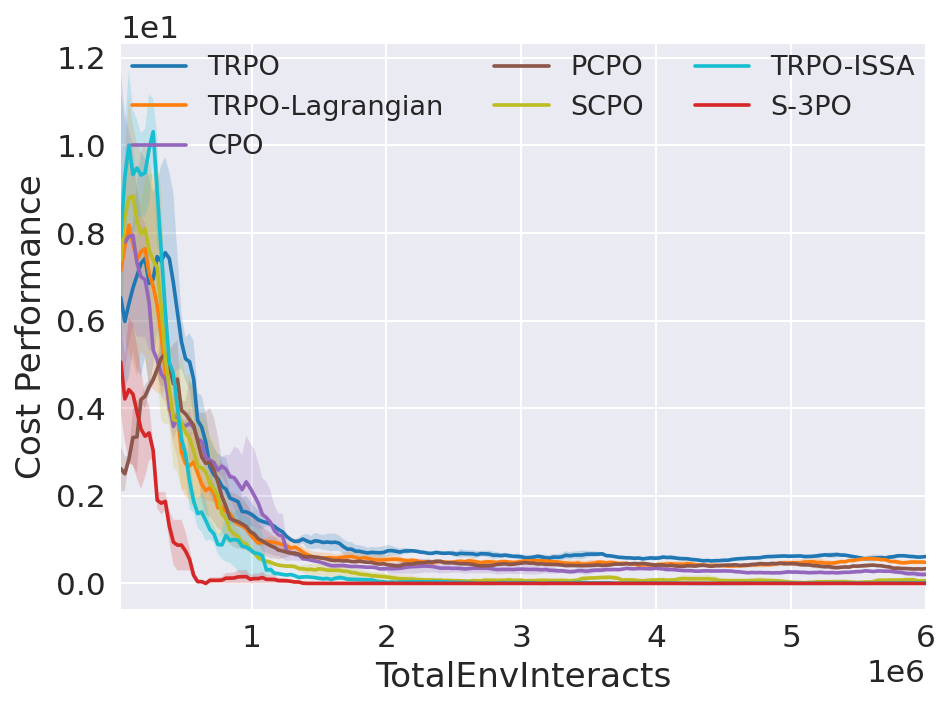} \\
            \includegraphics[width=0.24\textwidth]{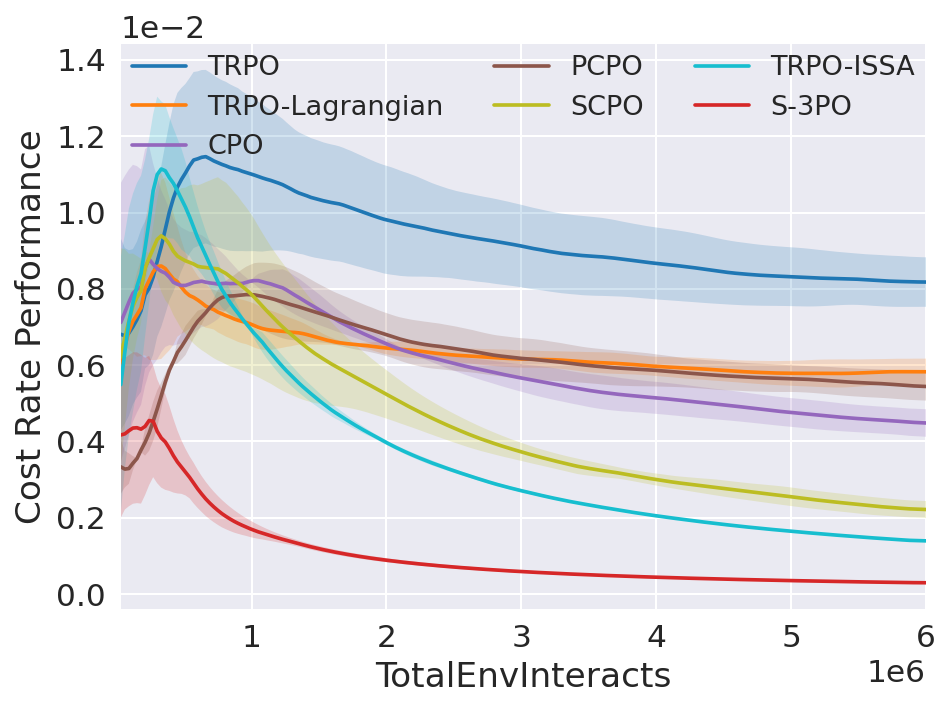} 
        }
        \label{fig:goal-point-1hazard-main}
    }
    \subfigure[Point\_8Hazard]{
        \centering
        \parbox{0.225\textwidth}{
            \includegraphics[width=0.24\textwidth]{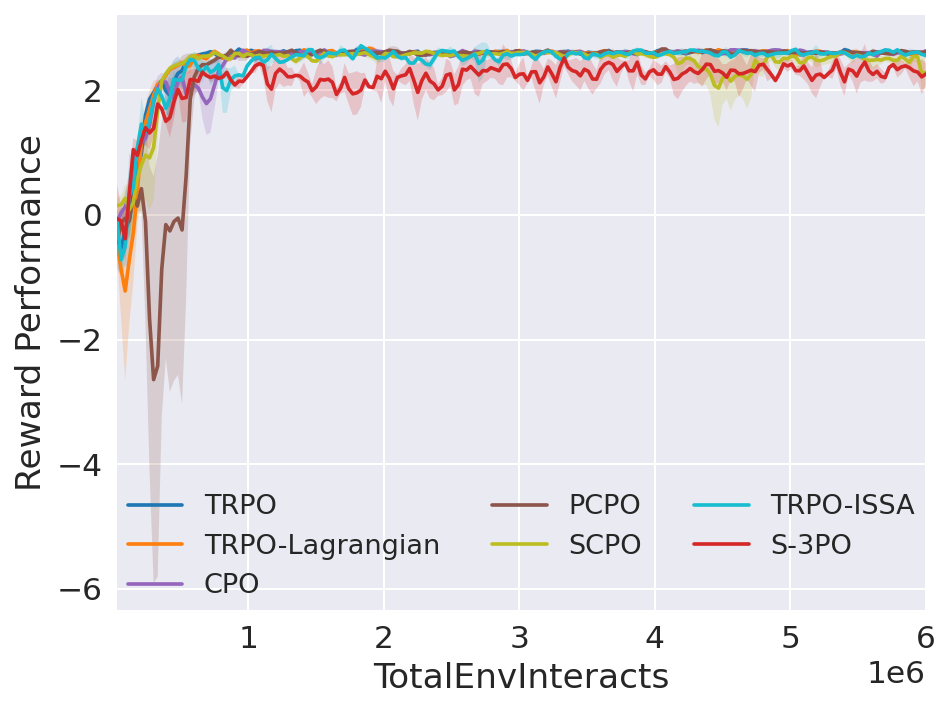} \\
            \includegraphics[width=0.24\textwidth]{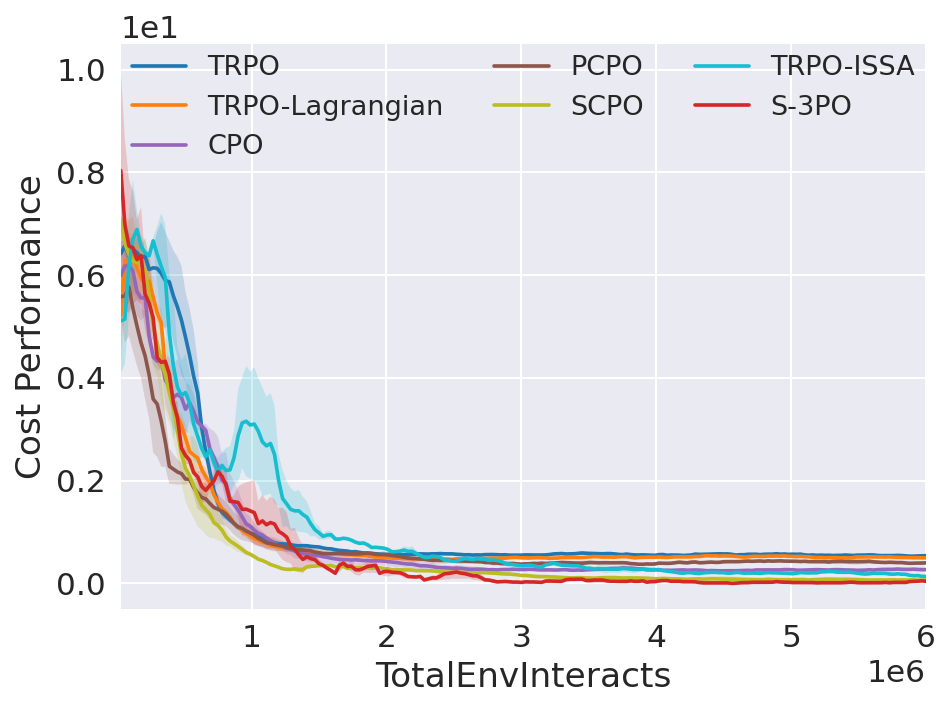} \\
            \includegraphics[width=0.24\textwidth]{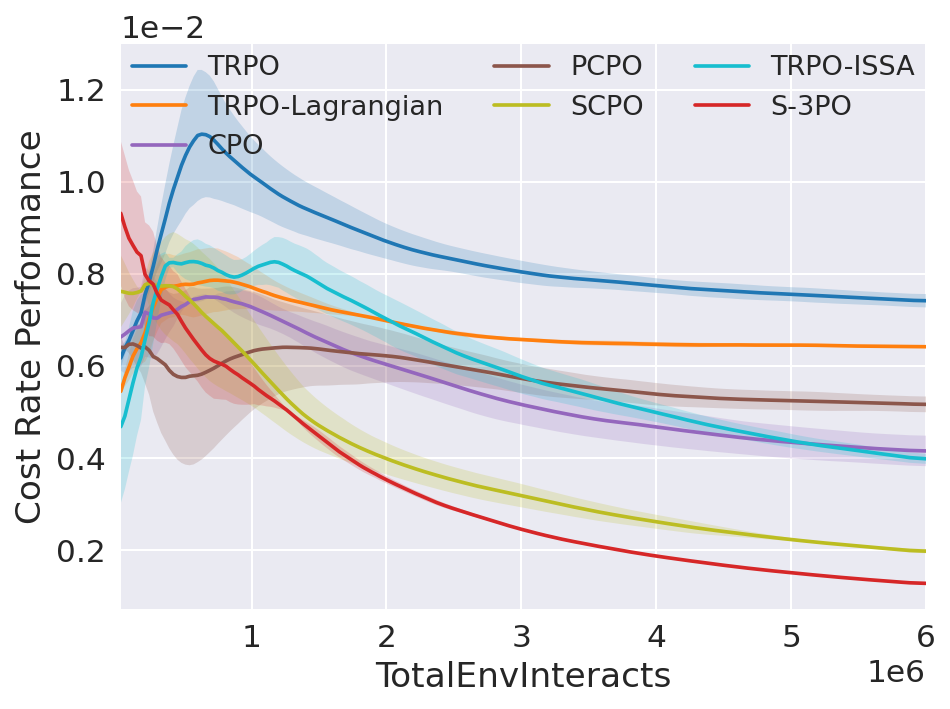}
        }
        \label{fig:goal-point-8hazard-main}
    }
    \subfigure[Point\_1Pillar]{
        \centering
        \parbox{0.225\textwidth}{
            \includegraphics[width=0.24\textwidth]{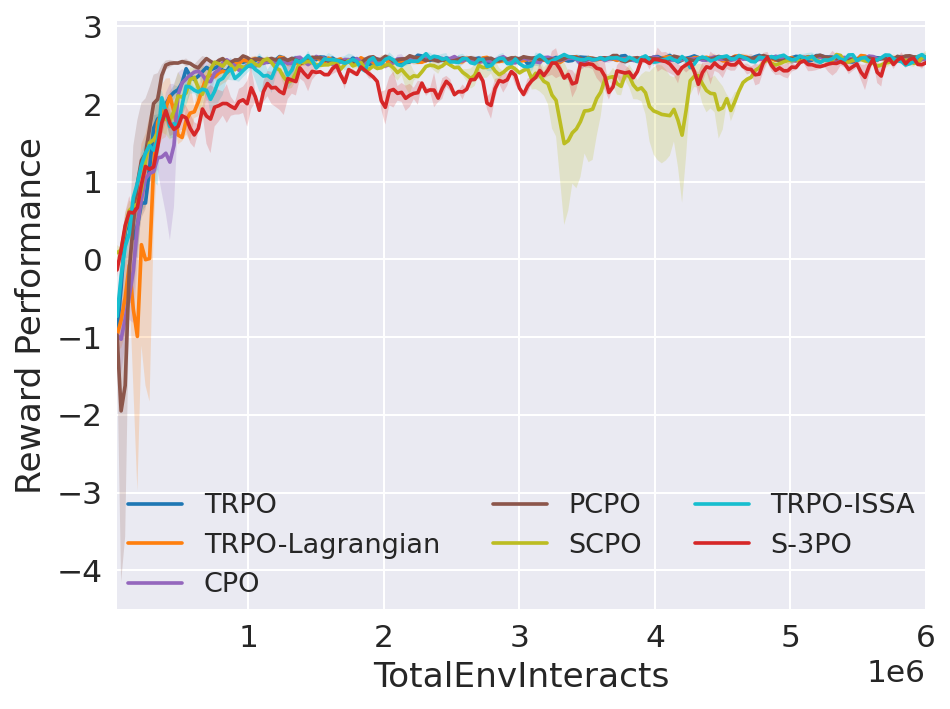} \\
            \includegraphics[width=0.24\textwidth]{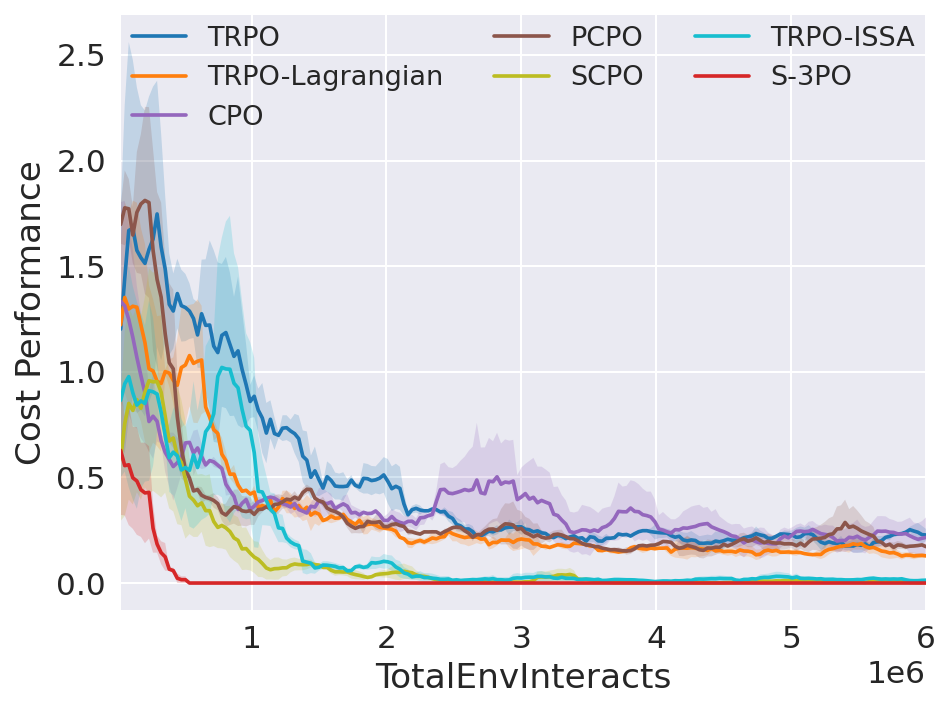} \\
            \includegraphics[width=0.24\textwidth]{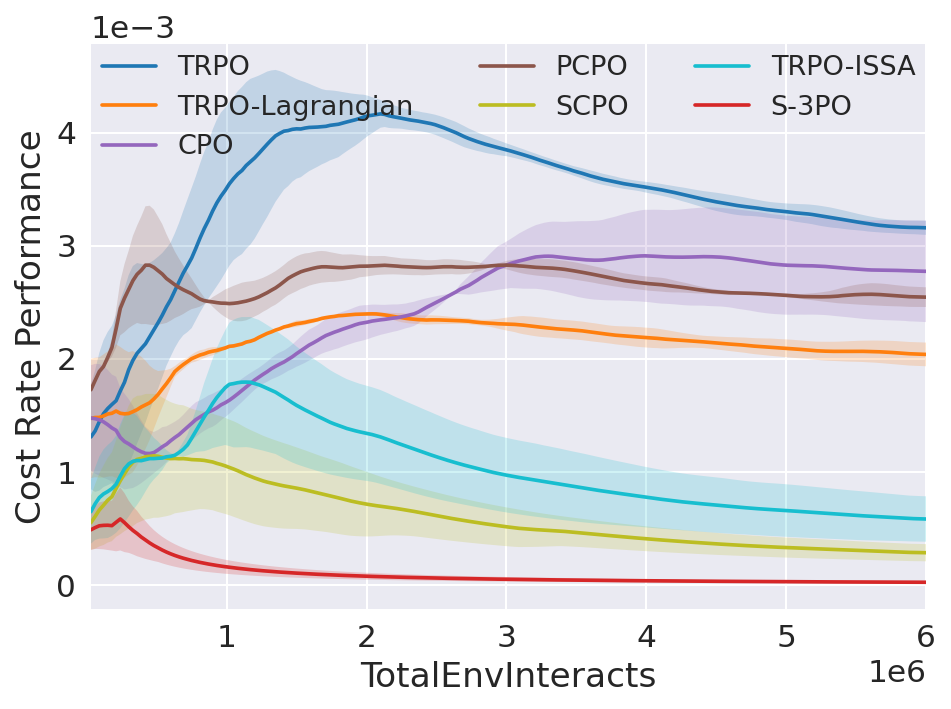} 
        }
        \label{fig:goal-point-1pillar-main}
    }
    \subfigure[Swimmer\_1Hazard]{
        \centering
        \parbox{0.225\textwidth}{
            \includegraphics[width=0.24\textwidth]{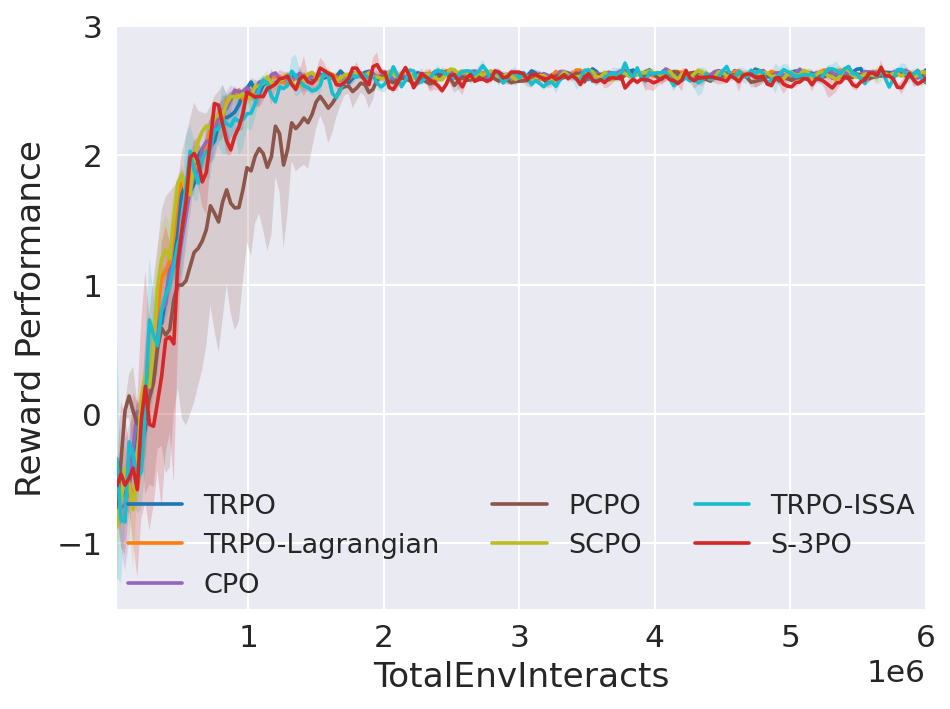} \\
            \includegraphics[width=0.24\textwidth]{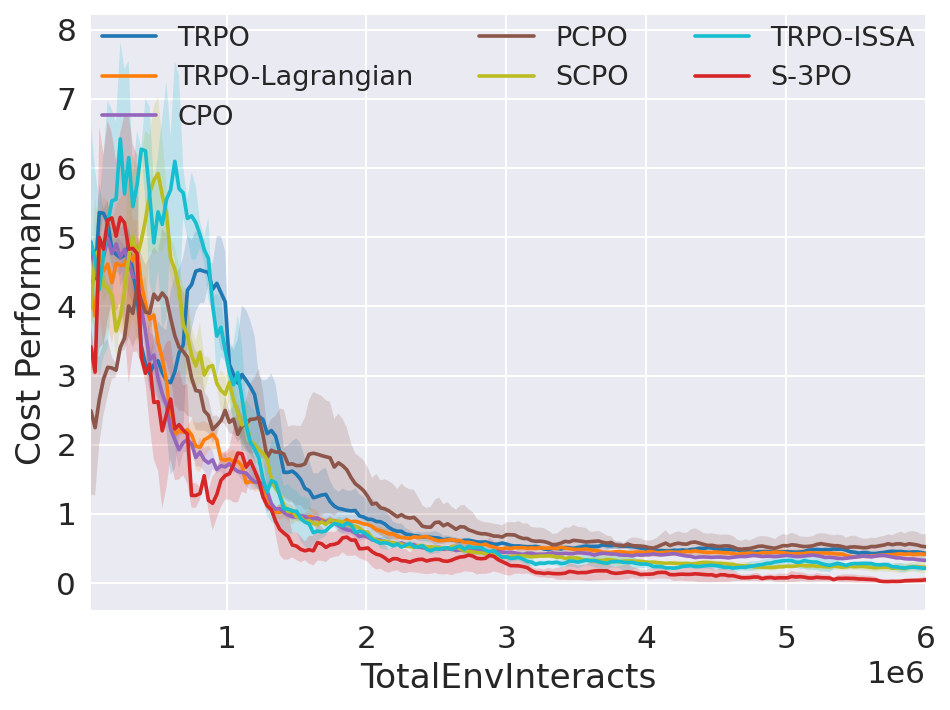} \\
            \includegraphics[width=0.24\textwidth]{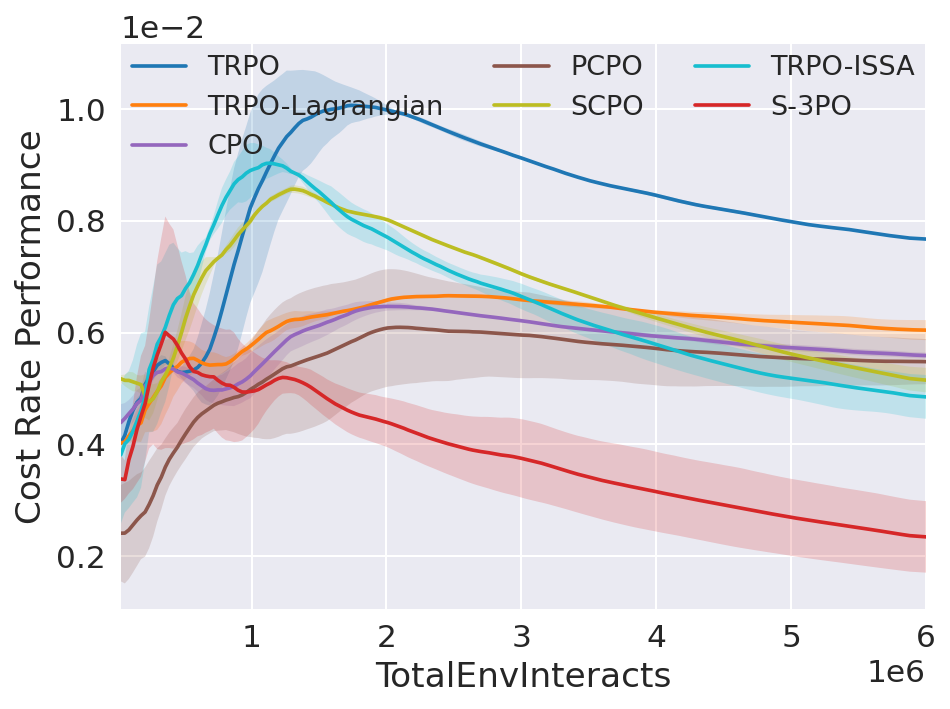}
        }
        \label{fig:goal-swimmer-1hazard-main}
    }
    \vspace{-10pt}
    \caption{Results from four representative test suites (evaluated without the safeguard).}
    \label{fig:comparison results-main}
    \vspace{-10pt}
\end{figure*}

\begin{figure*}[t]
    \centering
    \subfigure[Point\_1Hazard]{
        \centering
        \parbox{0.225\textwidth}{
            \includegraphics[width=0.24\textwidth]{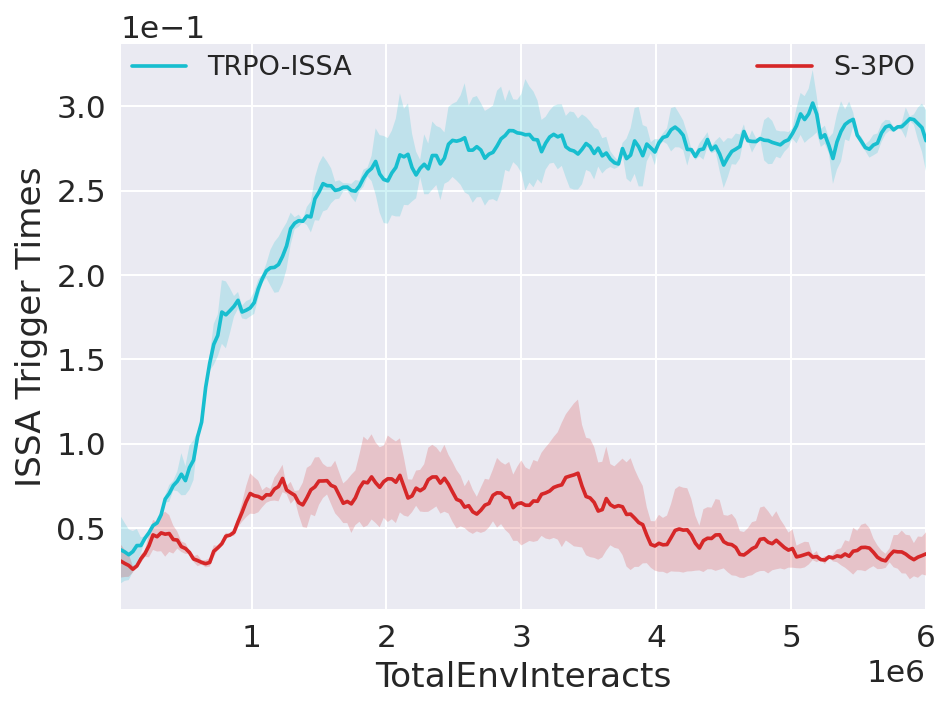}
        }
    \label{fig:goal-point-1hazard-ISSA-main}
    }
    \subfigure[Point\_8Hazard]{
        \centering
        \parbox{0.225\textwidth}{
            \includegraphics[width=0.24\textwidth]{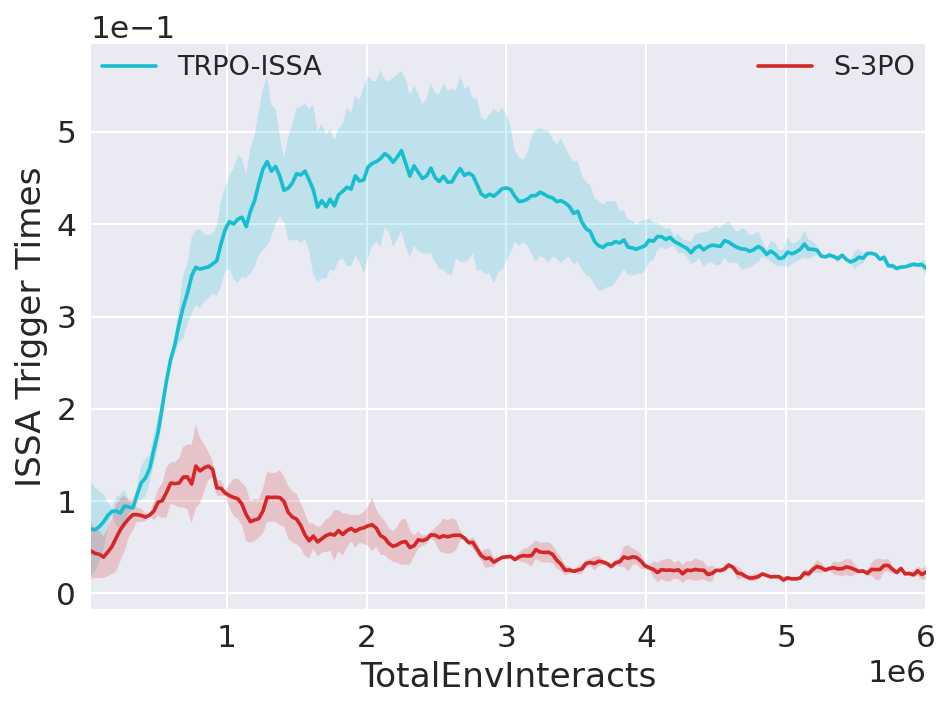}
        }
    \label{fig:goal-point-8hazard-ISSA-main}
    }
    \subfigure[Point\_1Pillar]{
        \centering
        \parbox{0.225\textwidth}{
            \includegraphics[width=0.24\textwidth]{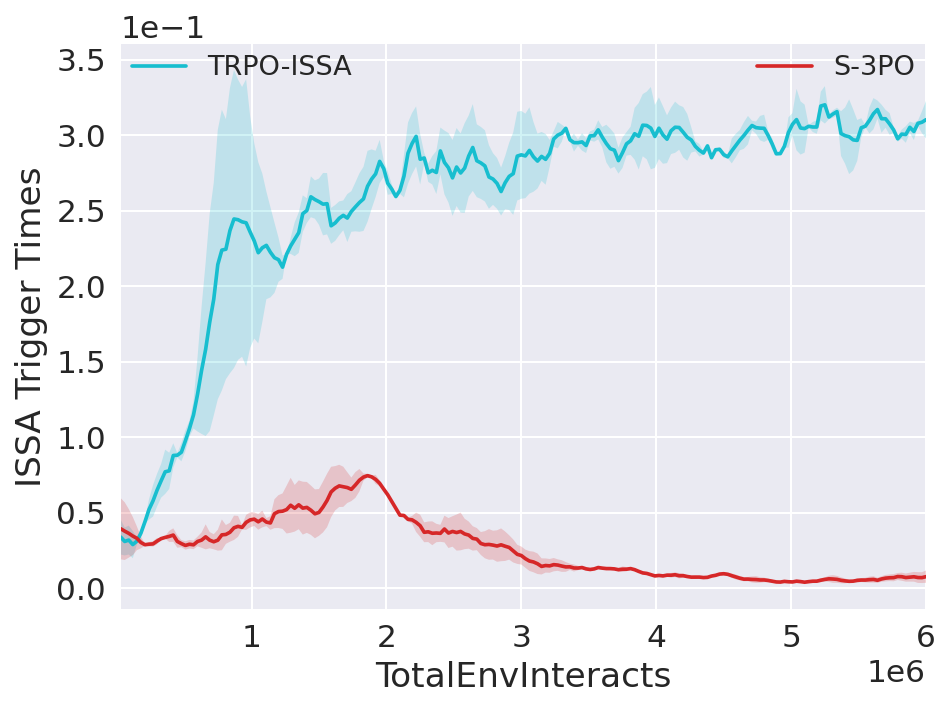}
        }
    \label{fig:goal-point-1pillar-ISSA-main}
    }
    \subfigure[Swimmer\_1Hazard]{
        \centering
        \parbox{0.225\textwidth}{
            \includegraphics[width=0.24\textwidth]{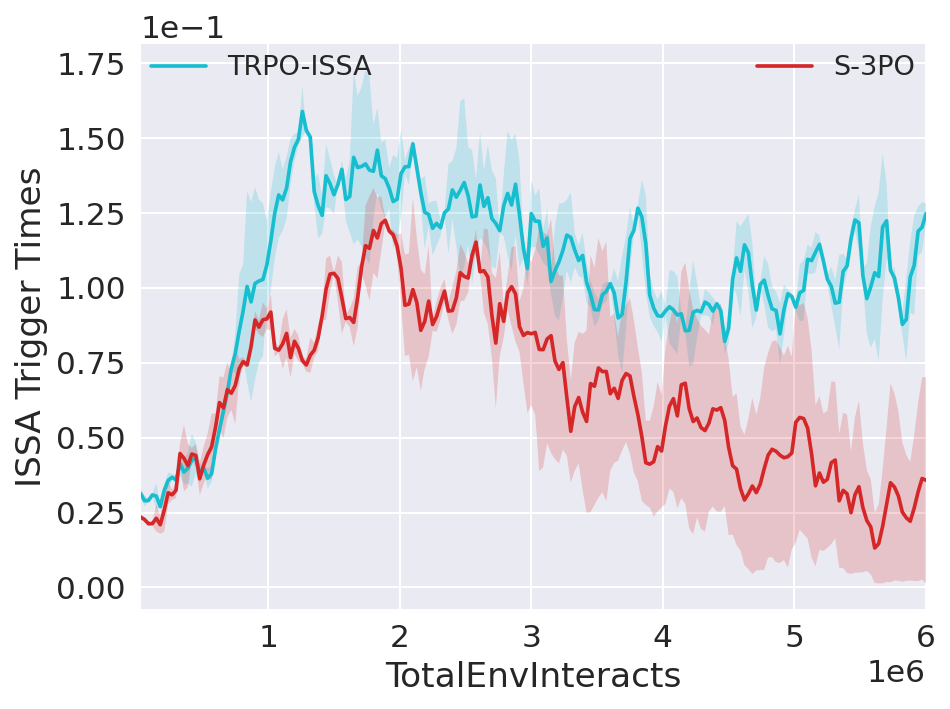}
        }
    \label{fig:goal-swimmer-1hazard-ISSA-main}
    }
    \vspace{-10pt}
    \caption{Triggering frequency of the safeguard from four representative test suites.}
    \label{fig:comparison results ISSA-main}
    \vspace{-15pt}
\end{figure*}

\subsection{Evaluating S-3PO and Comparison Analysis}

\textbf{~~~~~~Zero Violation During Training}  The training performance of four representative test suites are summarized in \Cref{fig:zero violation performance}, where the S-3PO algorithm clearly outperforms other baseline methods by achieving zero violations, consistent with the safety guarantee outlined in \cite{zhao2021model}. For more experiments, please check \Cref{sec: experiment details}. This superior performance is attributed to the safeguard mechanism within the S-3PO framework, which effectively corrects unsafe actions at every step, particularly during training. 
Furthermore, as demonstrated in \Cref{fig:comparison results-main}, the reward performance remains comparable to advanced baselines. This distinct capability of S-3PO ensures safe reinforcement learning with zero safety violations, addressing \textbf{Q1}. 

\textbf{State-wise Safety Without Safety Monitor}  
At the end of each epoch, the S-3PO policy is tested over 10,000 steps without the safeguard. This allows us to determine whether S-3PO effectively learns a state-wise safe policy through the guidance of the safe set-guided cost. As shown in \Cref{fig:comparison results-main}, S-3PO demonstrates superior performance even without the safeguard, achieving (i) near-zero average episode cost and (ii) significantly reduced cost rates, all while maintaining competitive reward performance. These findings 
highlights that by minimizing imaginary safety violations, the policy rapidly learns to act safely, which addresses \textbf{Q2}.

\textbf{Learn to Act without Safeguard} As highlighted in \Cref{obs: delta phi}, the key concept behind penalizing imaginary safety violations is to minimize the activation of the safeguard, thereby significantly reducing its computational complexity and enabling real-time implementation. To illustrate this, we visualize the average number of times the ISSA-based safeguard is triggered per step in \Cref{fig:comparison results ISSA-main}. For comparison, TRPO-ISSA is included as a baseline, which relies continuously on the safeguard to maintain safe control. \Cref{fig:comparison results ISSA-main} shows that S-3PO dramatically reduces the frequency of safeguard activations, approaching zero, indicating that a state-wise safe policy has been effectively learned, thus addressing \textbf{Q3}.

\begin{wrapfigure}{hr}{0.55\linewidth}
    \vspace{-20pt}
    \centering
    \subfigure[Comparison of cost rate performance with 6 different weights.]{
        \centering
        \includegraphics[width=0.24\textwidth]{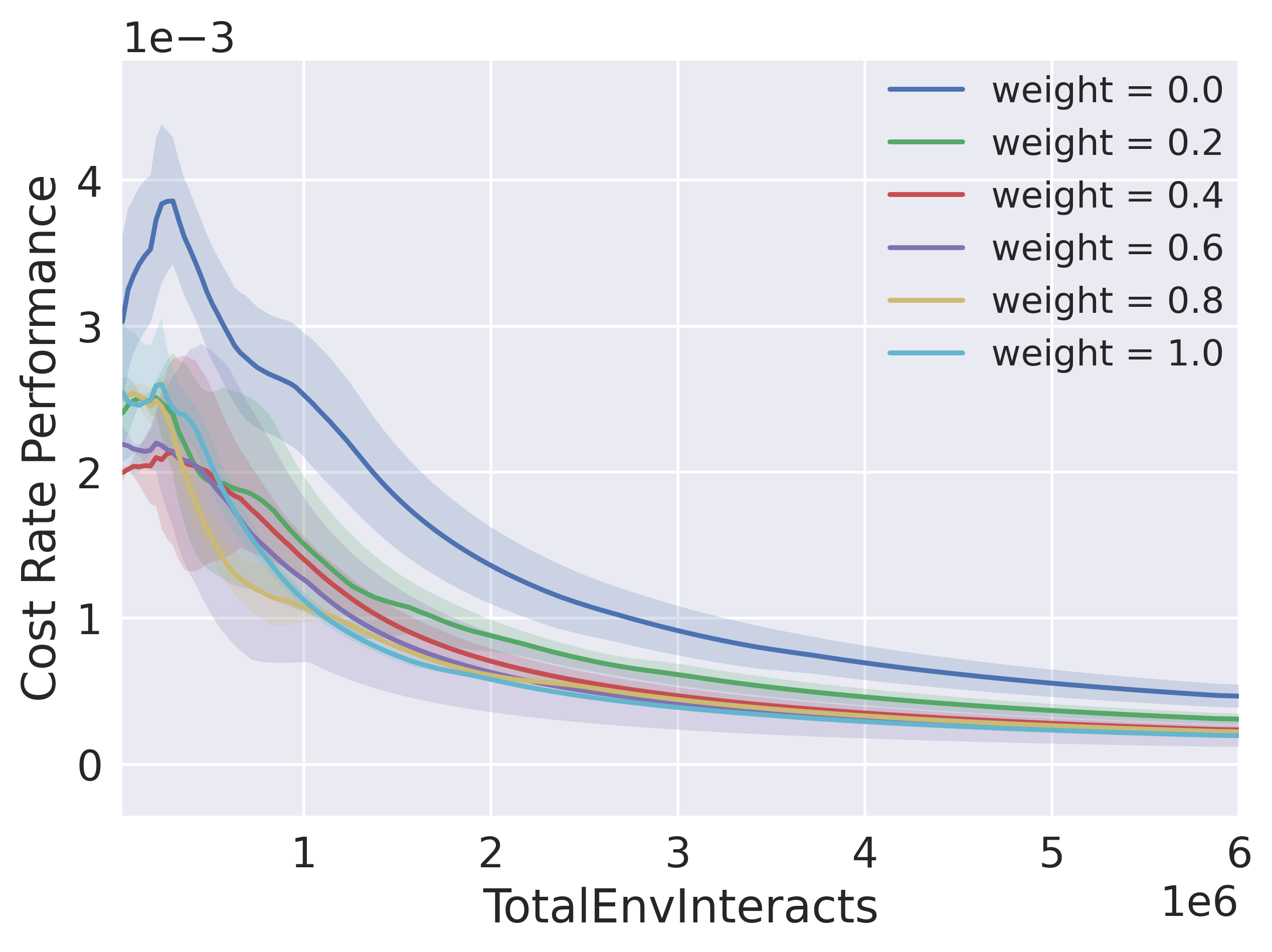}
        \label{fig:weight performance}
    }
    \hfill
    \subfigure[Comparison between ``imaginary'' cost and action correction cost.]{
        \centering
        \includegraphics[width=0.24\textwidth]{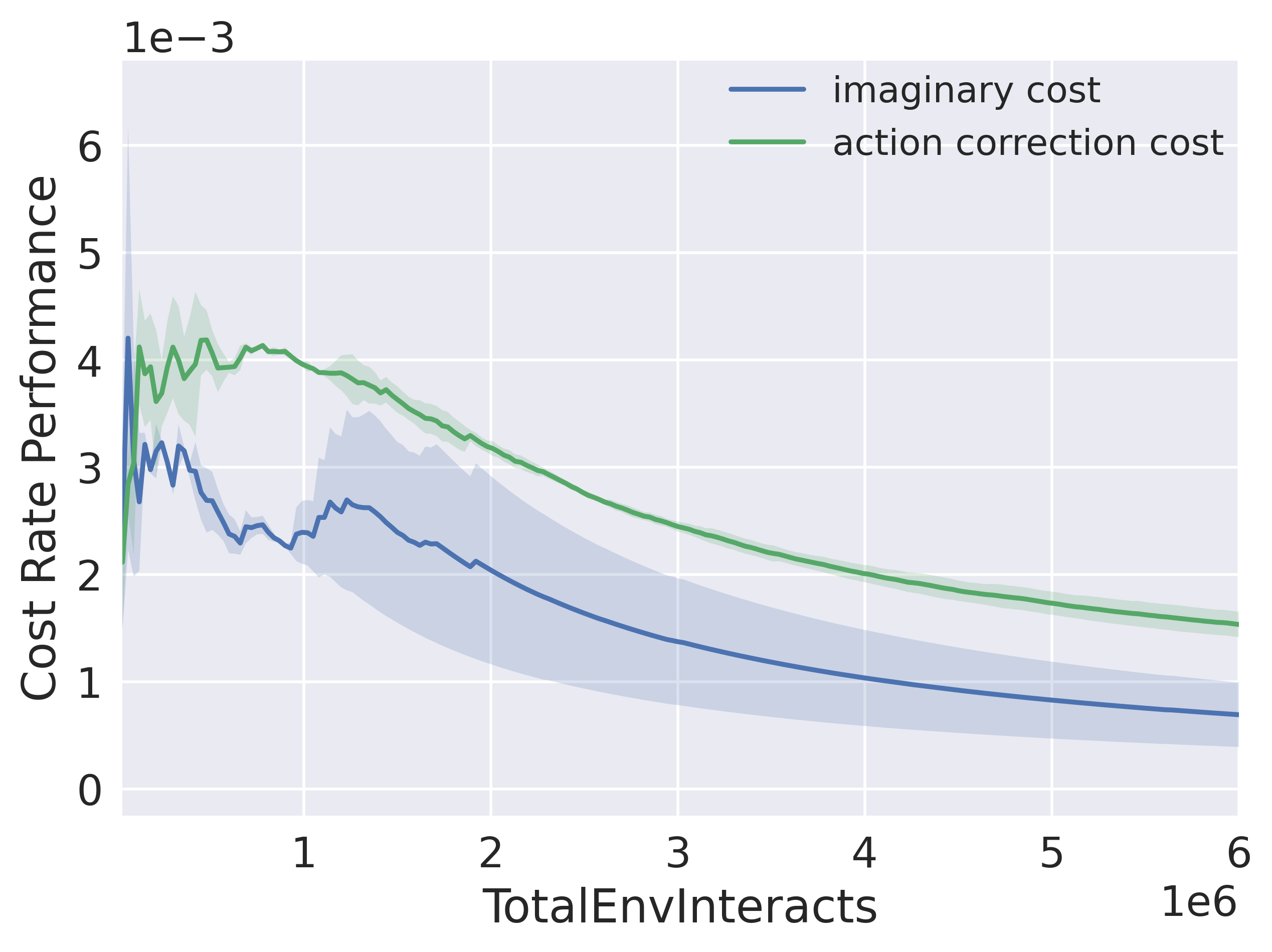}
        \label{fig:action correction performance}
    }
    \vspace{-5pt}
    \label{fig: comparison}
    \caption{Comparison results of S-3PO}
    \vspace{-15pt}
\end{wrapfigure}

\textbf{Ablation on Weighted Loss for Fitting Cost Increment Value Targets}  As pointed in \Cref{sec: practical}, fitting $V_{D_i}(\hs_t)$ is a critical step towards solving S-3PO, which is challenging due to non-increasing stair shape of the target sequence.
To elucidate the necessity of weighted loss for solving this challenge, we evaluate the cost rate of S-3PO under six distinct weight settings (0.0, 0.2, 0.4, 0.6, 0.8, 1.0) on \texttt{Point\_4Hazard} test suite. The results shown in \Cref{fig:weight performance} validates that a larger weight (hence higher penalty on predictions that violate the characteristics of value targets) results in better cost rate performance. This ablation study answers \textbf{Q4}. 

\textbf{Necessity of ``Imaginary'' Cost}  To understand the importance of the "imaginary" cost within the S-3PO framework, we compare it to another cost based on the magnitude of action correction~\cite{chen2021safe}. This empirical analysis is conducted using the \texttt{Point\_4Hazard} test suite. As shown in \Cref{fig:action correction performance}, the "imaginary" cost yields superior cost rate performance. This suggests that the "imaginary" cost offers deeper insights into the complex dynamics between the robot and its environment, thereby addressing \textbf{Q5}.

\begin{wrapfigure}{hr}{0.55\linewidth}
    \vspace{-20pt}
    \centering
    \includegraphics[width=0.24\textwidth]{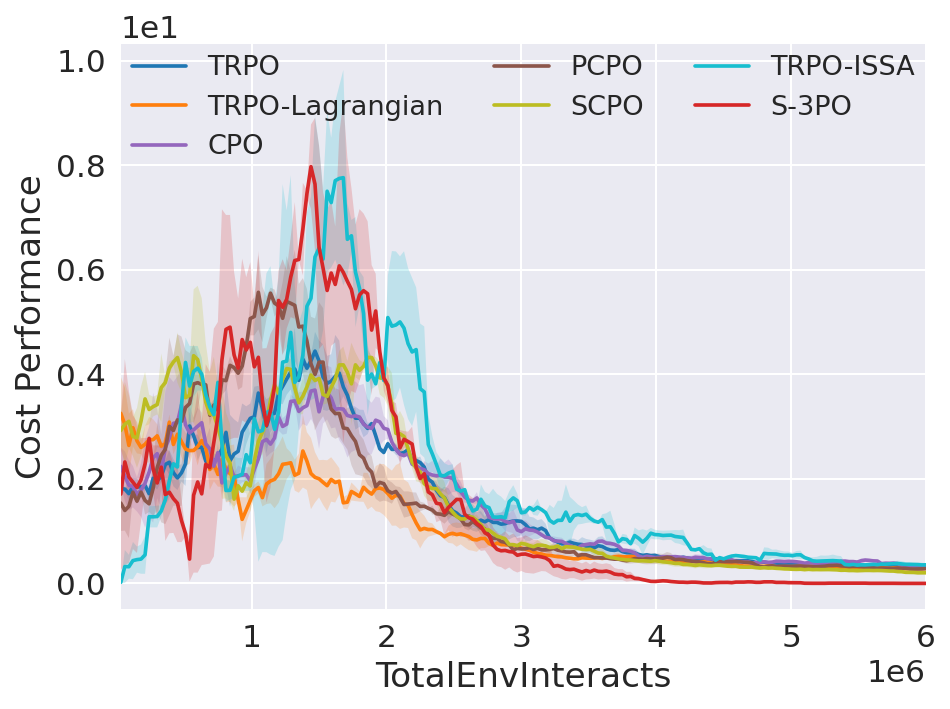}
    \includegraphics[width=0.24\textwidth]{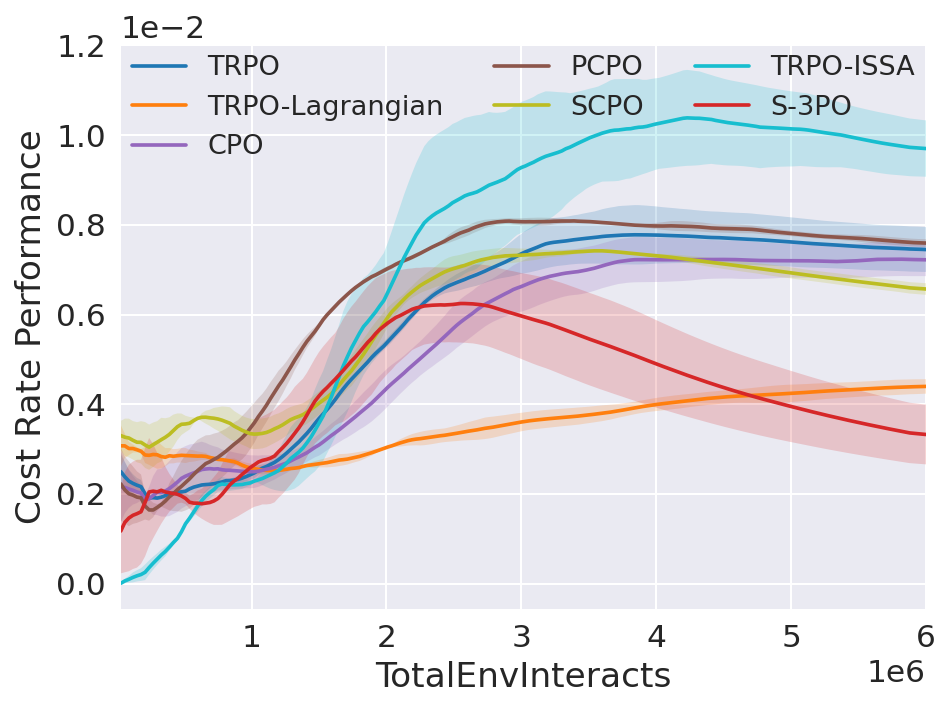}
    \vspace{-5pt}
    \caption{Cost performance of Ant\_1Hazard.}
    \label{fig:complex-link-robot-main}
    \vspace{-15pt}
\end{wrapfigure}

\textbf{Scale S-3PO to High-Dimensional Link Robots}  To showcase S-3PO's scalability and performance with complex, high-dimensional link robots, we conducted additional tests on \texttt{Ant\_1Hazard} featuring 8 dimensional control spaces. As shown in \Cref{fig:complex-link-robot-main}, S-3PO effectively drives the cost to zero and rapidly reduces the cost rate, showcasing its superiority in high-dimensional safety policy learning and highlighting its exceptional scalability to more complex systems, which answers \textbf{Q6}.

%% file: 8_conclusion.tex
\section{Conclusion and Future Prospectus}
In this study, we introduce Safe Set Guided State-wise Constrained Policy Optimization (S-3PO), a novel algorithm pioneering state-wise safe optimal policies. This distinction is underlined by the absence of training violations, signifying an error-free learning paradigm. S-3PO employs a safeguard anchored in black-box dynamics to ensure secure exploration. Subsequently, it integrates a novel ``imaginary'' safety cost to guide the RL agent towards optimal safe policies. In the following work, we will try to implement our algorithm in real robot implementation.


%% file: 9_acknowledgement.tex
\section*{Acknowledgement}

This work is supported by the National Science Foundation under grant No. 2144489.

%% file: appendix_algo.tex
\section{Algorithms}\label{appdx:algs}
\subsection{S-3PO Algorithm}
\begin{algorithm}
\caption{S-3PO}\label{alg:s3po_main}
\begin{algorithmic}
\STATE \textbf{Input:} Initial policy $\pi_0\in\Pi_\theta$.
\FOR{$k=0,1,2,\dots$}
\FOR{$t=0,1,2,\dots$}
\STATE Sample nominal action $a_t\sim\pi_k(s_t)$
\STATE Compute and execute $a_t^* = \mathrm{ISSA}(s_t,a^r_t)$
\STATE Log $\tau \gets \tau \cup \{(s_t,a_t,r_t,s_{t+1},\Delta \phi_t)\}$
\ENDFOR
\STATE $g \gets \left.\nabla_{\theta}\EE_{\hs,a\sim\tau}\left[A^{\pi}(\hs, a)\right]\right\rvert_{\theta=\theta_k}$
\STATE $b \gets \left.\nabla_{\theta}\EE_{\hs,a\sim\tau}\left[A^{\pi}_{D}(\hs, a)\right]\right\rvert_{\theta=\theta_k}$
\STATE $c \gets \mathcal{J}_{D}(\pi_k) + 2(H+1)\epsilon_{D}^{\pi} \sqrt{\delta/2}$
\STATE  $H \gets \left.\nabla^2_{\theta} \mathbb{E}_{\hs \sim \tau}[\mathcal{D}_{KL}(\pi \| \pi_k)[\hs]]\right\rvert_{\theta=\theta_k}$
\STATE $\theta^*_{k+1} = \argmaxwrt{\theta} ~ g^\top(\theta-\theta_k) ~ \st$
\STATE $~~~\frac{1}{2}(\theta-\theta_k)^\top H (\theta-\theta_k) \leq \delta, c + b^\top(\theta-\theta_k) \leq 0$
\STATE Get search direction $\Delta\theta^* \gets \theta^*_{k+1} - \theta_k$
\FOR{$j=0,1,2,\dots$} 
\STATE $\theta' \gets \theta_{k} + \xi^j\Delta\theta^*$
\IF{
$\mathbb{E}_{\hs \sim \tau}[\mathcal{D}_{KL}(\pi_{\theta'} \| \pi_k)[\hs]] \leq \delta$ 
\textbf{and}\\
$\EE_{\hs,a\sim\tau}\left[A^{\pi_{\theta'}}_{D}(\hs, a)-A^{\pi_{k}}_{D}(\hs, a)\right] \leq \mathrm{max}(-c,0)$ \textbf{and} \\
$(k\leq k_\mathrm{safe}~\mathrm{or}~\EE_{\hs,a\sim\tau}\left[A^{\pi_{\theta'}}(\hs, a)\right] \geq \EE_{\hs,a\sim\tau}\left[A^{\pi_k}(\hs, a)\right])$}
\STATE $\theta_{k+1} \gets \theta'$ \COMMENT{\textcolor{blue}{Update policy}}
\STATE \textbf{break}
\ENDIF
\ENDFOR
\ENDFOR
\end{algorithmic}
\end{algorithm}

\clearpage

%% file: appendix_proof.tex
\section{Proof of Lemma \ref{prop: s-3po safety convergence}}
\label{sec: s-3po safety convergence}

\begin{proof}
To prove that the two conditions are equivalent, we show that condition 2 can be derived from condition 1, and vice versa.

\textbf{(Condition 1 $\Rightarrow$ Condition 2):}

By the property of ISSA, $\Delta \phi > 0$ only when the safety filter is triggered. Under condition 1, we have $\Delta \phi \leq 0$ throughout the trajectory in the training environment, which implies that ISSA is never activated during training. As a result, the trajectory remains entirely within the safe state set, satisfying $\max_t \phi_t \leq 0$.

Moreover, since ISSA is inactive, the applied action coincides with the nominal policy action, i.e., $a_t^* = a_t$ for all $t$. Therefore, the same sequence of actions will be executed in the evaluation environment (which uses the same initial state), producing the same trajectory that stays within the safe set, with $\max_t \phi_t \leq 0$ and without triggering ISSA.

\textbf{(Condition 2 $\Rightarrow$ Condition 1):}

Conversely, if a given policy starting from a certain initial state generates a trajectory in the evaluation environment that never triggers ISSA and satisfies $\max_t \phi_t \leq 0$, the same trajectory will be produced in the training environment when initialized from the same state. Along this trajectory, ISSA remains inactive, implying that the imaginary cost remains zero, i.e., $\Delta \phi_t \leq 0$ for all $t$.

Since both conditions ensure that ISSA is never triggered along the entire trajectory, the actions remain unchanged, $a_t^* = a_t$ for all $t$, and the resulting trajectories are identical in both the training and evaluation environments.

\end{proof}

It is important to note that the guarantee in Lemma \ref{prop: s-3po safety convergence} holds only when both conditions are strictly satisfied over the entire trajectory distribution induced by the policy. In contrast, for the probability space where the conditions are satisfied in expectation, the trajectories in the evaluation environment may deviate from those in the training environment, even when initialized from the same initial state. Such deviation can occur if either the imaginary cost or the safety index $\phi$ becomes positive at any step along the trajectory. The deviation may lead to discrepancies in the policy performance during evaluation, which will be analyzed in future work.

%% file: appendix_experiments.tex
\newpage
\section{Expeiment Details}
\label{sec: experiment details}
\subsection{Environment Settings}
\label{appendix:Environment Settings}

\begin{figure}[ht]
    \centering
    \subfigure[Point]{
        \centering
        \includegraphics[scale=0.1]{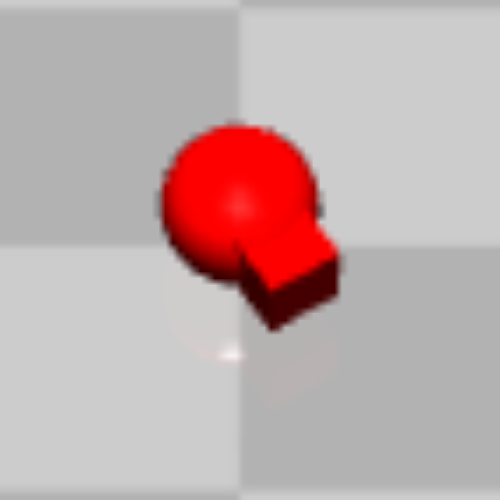}
        \label{fig: Point}
    }
    \hfill
    \subfigure[Swimmer]{
        \centering
        \includegraphics[scale=0.1]{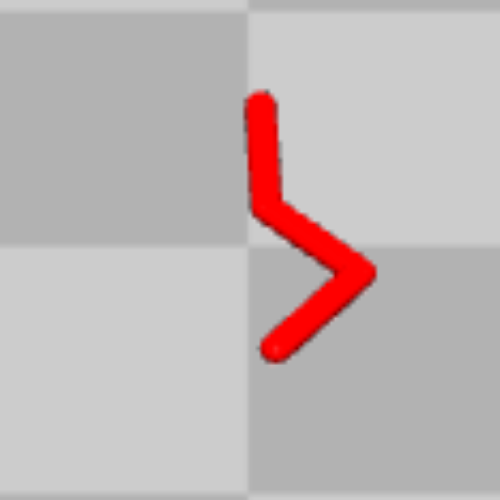}
        \label{fig: Swimmer}
    }
    \hfill
    \subfigure[Ant]{
        \centering
        \includegraphics[scale=0.1]{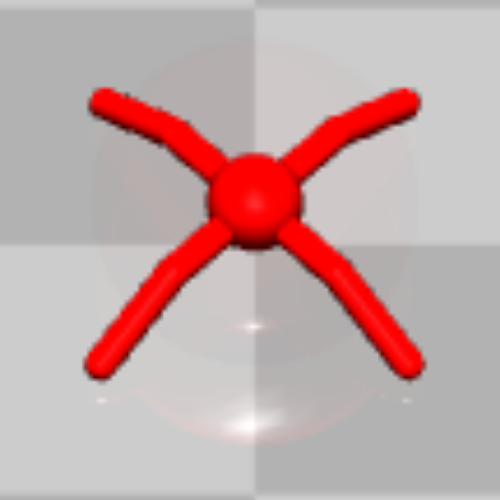}
        \label{fig: Ant}
    }
    \hfill
    \subfigure[Hazard]{
        \centering
        \includegraphics[scale=0.15]{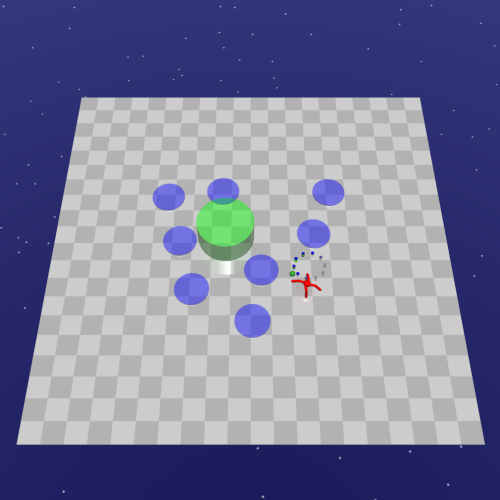}
        \label{fig: hazard}
    }
    \hfill
    \subfigure[Pillar]{
        \centering
        \includegraphics[scale=0.15]{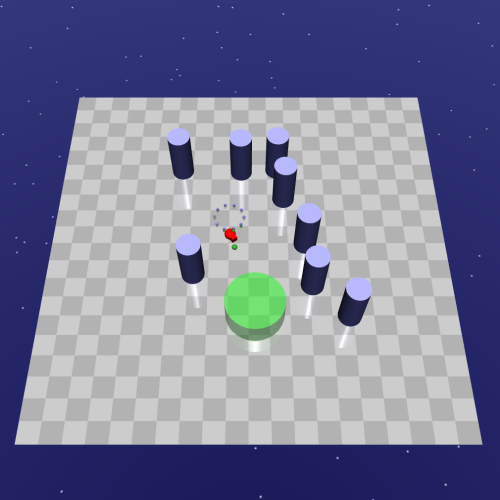}
        \label{fig: pillar}
    }
  \label{fig: constraints}
  \label{fig: robots}
  \caption{Robots and constraints for benchmark problems in our environment.}
  \vspace{-10pt}
\end{figure}

\paragraph{Goal Task}
In the Goal task environments, the reward function is:
\begin{equation}\notag
\begin{split}
    & r(x_t) = d^{g}_{t-1} - d^{g}_{t} + \mathbf{1}[d^g_t < R^g]~,\\
\end{split}
\end{equation}
where $d^g_t$ is the distance from the robot to its closest goal and $R^g$ is the size (radius) of the goal. When a goal is achieved, the goal location is randomly reset to someplace new while keeping the rest of the layout the same. The test suites of our experiments are summarized in \Cref{tab: testing suites}.
\paragraph{Hazard Constraint}
In the Hazard constraint environments, the cost function is:
\begin{equation}\notag
\begin{split}
    & c(x_t) = \max(0, R^h - d^h_t)~,\\
\end{split}
\end{equation}
where $d^h_t$ is the distance to the closest hazard and $R^h$ is the size (radius) of the hazard.
\paragraph{Pillar Constraint}
In the Pillar constraint environments, the cost $c_t = 1$ if the robot contacts with the pillar otherwise $c_t = 0$. 

\begin{figure}[h]
    \centering
    \includegraphics[scale=0.08]{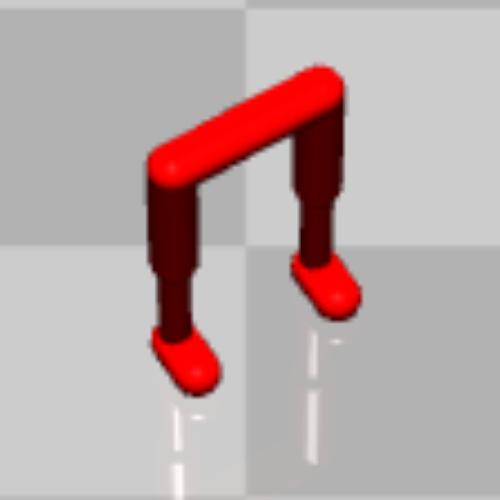}
    \caption{Walker}
    \label{fig: Walker}
\end{figure}

\paragraph{Additional high dimensional link robot}
To scale our method to high dimensional link robots. We additionally adopt \textbf{Walker} shown in \ref{fig: Walker} as a bipedal robot ($\mathcal{A} \subseteq \mathbb{R}^{10}$) in our experiments.
\paragraph{State Space}
The state space is composed of two parts. The internal state spaces describe the state of the robots, which can be obtained from standard robot sensors (accelerometer, gyroscope, magnetometer, velocimeter, joint position sensor, joint velocity sensor and touch sensor). The details of the internal state spaces of the robots in our test suites are summarized in \Cref{tab:internal_state_space}.
The external state spaces are describe the state of the environment observed by the robots, which can be obtained from 2D lidar or 3D lidar (where each lidar sensor perceives objects of a single kind). The state spaces of all the test suites are summarized in \Cref{tab:external_state_space}. Note that Vase and Gremlin are two other constraints in Safety Gym \cite{ray2019benchmarking} and all the returns of vase lidar and gremlin lidar are zero vectors (i.e., $[0, 0, \cdots, 0] \in \mathbb{R}^{16}$) in our experiments since none of our test suites environments has vases.
\begin{table}[h]
\vskip 0.15in
\caption{The test suites environments of our experiments}
\begin{center}
\begin{tabular}{c|cc|cc}
\toprule
\textbf{Task Setting} & \multicolumn{2}{c|}{Low dimension} &\multicolumn{2}{c}{High dimension}\\
\cline{2-5}\\[-1.02em]
&Point  & \multicolumn {1}{c|}{Swimmer} 
& Walker & \multicolumn {1}{c}{Ant}\\
\cline{1-5}\\[-1.02em]
Hazard-1    & \checkmark & \checkmark & \checkmark & \checkmark \\
Hazard-4    & \checkmark &            &            &            \\
Hazard-8    & \checkmark &            &            &            \\
Pillar-1    & \checkmark &            &            &            \\
Pillar-4    & \checkmark &            &            &            \\
Pillar-8    & \checkmark &            &            &            \\
\bottomrule
\end{tabular}
\label{tab: testing suites}
\end{center}
\end{table}

\begin{table}[h]
\vskip 0.15in
\caption{The internal state space components of different test suites environments.}
\begin{center}
\begin{tabular}{c|cccc}
\toprule
\textbf{Internal State Space} & Point  & Swimmer & Walker & Ant \\
\hline
Accelerometer ($\mathbb{R}^3$) & \checkmark & \checkmark & \checkmark & \checkmark \\
Gyroscope ($\mathbb{R}^3$) & \checkmark & \checkmark & \checkmark & \checkmark \\
Magnetometer ($\mathbb{R}^3$) & \checkmark & \checkmark & \checkmark & \checkmark \\
Velocimeter ($\mathbb{R}^{3}$) & \checkmark & \checkmark & \checkmark & \checkmark \\
Joint position sensor ($\mathbb{R}^{n}$) & ${n=0}$ & ${n=2}$ & ${n=10}$ & ${n=8}$ \\
Joint velocity sensor ($\mathbb{R}^{n}$)  & ${n=0}$ & ${n=2}$ & ${n=10}$ & ${n=8}$ \\
Touch sensor ($\mathbb{R}^{n}$) & ${n=0}$ & ${n=4}$ & ${n=2}$ & ${n=8}$ \\
\bottomrule
\end{tabular}
\label{tab:internal_state_space}
\end{center}
\end{table}

\begin{table}[h]
\vskip 0.15in
\caption{The external state space components of different test suites environments.}
\begin{center}
\begin{tabular}{c|ccc}
\toprule
\textbf{External State Space} &  Goal-Hazard & Goal-Pillar\\
\hline\\[-0.8em]
Goal Compass ($\mathbb{R}^{3}$) & \checkmark & \checkmark\\
Goal Lidar ($\mathbb{R}^{16}$) & \checkmark  & \checkmark\\
3D Goal Lidar ($\mathbb{R}^{60}$) & \xmark  & \xmark\\
Hazard Lidar ($\mathbb{R}^{16}$) & \checkmark & \xmark\\
3D Hazard Lidar ($\mathbb{R}^{60}$) & \xmark  & \xmark\\
Pillar Lidar ($\mathbb{R}^{16}$) & \xmark  & \checkmark\\
Vase Lidar ($\mathbb{R}^{16}$) & \checkmark  & \checkmark\\
Gremlin Lidar ($\mathbb{R}^{16}$) & \checkmark & \checkmark\\
\bottomrule
\end{tabular}
\label{tab:external_state_space}
\end{center}
\end{table}

\paragraph{Control Space}
For all the experiments, the control space of all robots are continuous, and linearly scaled to [-1, +1].

\subsection{Policy Settings}
\label{appendix:Policy Settings}
The hyper-parameters used in our experiments are listed in \Cref{tab:policy_setting} as default.

Our experiments use separate multi-layer perception with ${tanh}$ activations for the policy network, value network and cost network. Each network consists of two hidden layers of size (64,64). All of the networks are trained using $Adam$ optimizer with learning rate of 0.01.

We apply an on-policy framework in our experiments. During each epoch the agent interact $B$ times with the environment and then perform a policy update based on the experience collected from the current epoch. The maximum length of the trajectory is set to 1000 and the total epoch number $N$ is set to 200 as default. In our experiments the Walker was trained for 250 epochs due to the high dimension.

The policy update step is based on the scheme of TRPO, which performs up to 100 steps of backtracking with a coefficient of 0.8 for line searching.

For all experiments, we use a discount factor of $\gamma = 0.99$, an advantage discount factor $\lambda =0.95$, and a KL-divergence step size of $\delta_{KL} = 0.02$.

For experiments which consider cost constraints we adopt a target cost $\delta_{c} = 0.0$ to pursue a zero-violation policy.

Other unique hyper-parameters for each algorithms are hand-tuned to attain reasonable performance. 

Each model is trained on a server with a 48-core Intel(R) Xeon(R) Silver 4214 CPU @ 2.2.GHz, Nvidia RTX A4000 GPU with 16GB memory, and Ubuntu 20.04.

\begin{sidewaystable}
\vskip 0.15in
\caption{Important hyper-parameters of different algorithms in our experiments}
\begin{center}
\resizebox{\textwidth}{!}{%
\begin{tabular}{lr|ccccccc}
\toprule
\textbf{Policy Parameter} & & TRPO & TRPO-ISSA & TRPO-Lagrangian & CPO & PCPO &  SCPO & S-3PO\\
\hline\\[-1.0em]
Epochs & $N$ & 200  & 200 & 200 & 200 & 200 & 200 & 200 \\
Steps per epoch & $B$& 30000 & 30000 & 30000 & 30000 & 30000 & 30000 & 30000\\
Maximum length of trajectory & $L$ & 1000 & 1000 & 1000 & 1000 & 1000 & 1000\\
Policy network hidden layers & & (64, 64) & (64, 64) & (64, 64) & (64, 64) & (64, 64) & (64, 64) & (64, 64)\\
Discount factor  &  $\gamma$ & 0.99 & 0.99 & 0.99 & 0.99 & 0.99 & 0.99 & 0.99\\
Advantage discount factor  & $\lambda$ & 0.97 & 0.97 & 0.97 & 0.97 & 0.97  & 0.97 & 0.97\\
TRPO backtracking steps & &100 &100 &100  &100 & - &100 &100\\
TRPO backtracking coefficient & &0.8 &0.8 &0.8 &0.8 & - &0.8 &0.8\\
Target KL & $\delta_{KL}$& 0.02 & 0.02 & 0.02 & 0.02 & 0.02 & 0.02 & 0.02\\

Value network hidden layers & & (64, 64) & (64, 64) & (64, 64) & (64, 64) & (64, 64) & (64, 64) & (64, 64)\\
Value network iteration & & 80 & 80 & 80 & 80 & 80 & 80 & 80 \\
Value network optimizer & & Adam & Adam & Adam & Adam & Adam & Adam & Adam\\
Value learning rate & & 0.001 & 0.001 & 0.001 & 0.001 & 0.001 & 0.001 & 0.001\\

Cost network hidden layers & & - & -& (64, 64) & (64, 64) & (64, 64) & (64, 64) & (64, 64)\\
Cost network iteration & & - & - & 80 & 80 & 80 & 80 & 80\\
Cost network optimizer & & - & -& Adam & Adam & Adam & Adam & Adam\\
Cost learning rate & & - & - & 0.001  & 0.001 & 0.001 & 0.001 & 0.001\\
Target Cost & $\delta_{c}$& - & -& 0.0 & 0.0 & 0.0 & 0.0 & 0.0\\

Lagrangian learning rate & & -  &- & 0.005 & - & - & - & -\\

Cost reduction & & - & - & - & 0.0 & - & 0.0 & 0.0\\
\bottomrule
\end{tabular}
}
\label{tab:policy_setting}
\end{center}
\end{sidewaystable}

\subsection{Practical Implementation Details}
\label{sec: practical detail}

In this section, we summarize implementation techniques that helps with S-3PO's practical performance.
The pseudocode of S-3PO is give as \cref{alg:s3po_main}.
\paragraph{Weighted loss for cost value targets}

\label{weighted loss}

\begin{wrapfigure}{r}{0.3\textwidth}
  \vspace{-10pt}
\centering
  \includegraphics[width=0.3\textwidth]{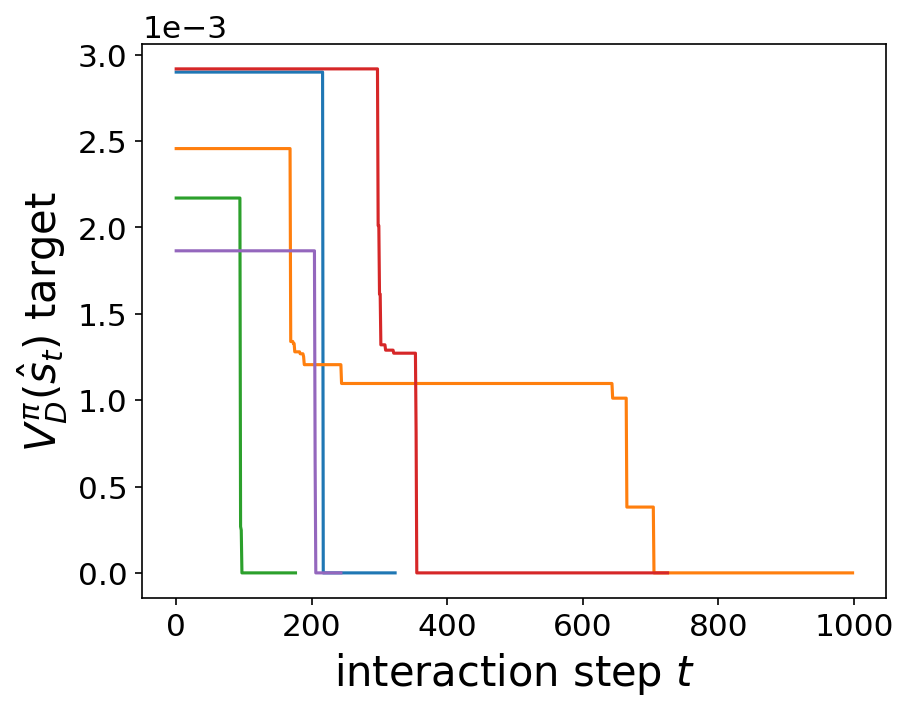}
  \caption{$V_{[D_i]\pi}(\hs)$ target of five sampled episodes.}
  \label{fig:vc_target}
  \vspace{-60pt}
\end{wrapfigure}

A critical step in S-3PO requires fitting of the cost increment value functions, denoted as $V_{D}^\pi(\hs_t)$. 
By definition, $V_{D}^\pi(\hs_t)$ is equal to the maximum cost increment in any future state over the maximal state-wise cost so far. In other words, $V_{D}^\pi(\hs_t)$ forms a non-increasing stair shape along the trajectory. Here we visualize an example of $V_{D}^\pi(\hs_t)$ in \Cref{fig:vc_target}.
To enhance the accuracy of fitting this stair shape function, a weighted loss strategy is adopted, capitalizing on its monotonic property.
Specifically, we define a weighted loss $L_{weight}$:
\begin{equation}\notag
\begin{split}
L_{weight} &= L(\hat{y}_t - y_t) * (1 + w * \mathbbm{1}[(\hat{y}_t - y_{t - 1}) > 0])
\end{split}
\end{equation}

where $L$ denotes Mean Squared Error (MSE), $\hat{y}_t$ is the prediction, ${y}_t$ is the fitting target and ${w}$ is the penalty weight. To account for the initial step (${t = 0}$), we set ${y_{t - 1}}$ to sufficiently large, thereby disregarding the weighted term associated with the first step. In essence, the rationale is to penalize any prediction that violates the non-increasing characteristics of the target sequence, thereby leading to an improved fitting quality.

\paragraph{Line search scheduling}
Note that in \eqref{eq: s3po optimization final}, there are two constraints: (a) the trust region and (b) the bound  on expected advantages.
In practice, due to approximation errors, constraints in \eqref{eq: s3po optimization final} might become infeasible.
In that case, we perform a recovery update that only enforces the cost advantage $A_D^\pi$ to decrease starting from early training steps (in first $k_\mathrm{safe}$ updates), and starts to enforce reward improvements of $A^\pi$ towards the end of training.
This is different from \cite{zhao2023scpo}, where the reward improvements are enforced at all times.
This is because SCPO only guarantees safety (constraint satisfaction) after convergence, while S-3PO prioritizes constraining imaginary safety violation.
With our line search scheduling, S-3PO is able to first grasp a safe policy, and then improve the reward performance.
In that way, S-3PO achieves zero safety violation both during training and in testing with a worst-case performance degradation guarantee.

\subsection{Metrics Comparison}
\label{appendix:Metrics Comparison}
\begin{figure}[htbp]
    \centering
    \subfigure[Point\_1Hazard]{
        \centering
        \parbox{0.31\textwidth}{
            \includegraphics[width=0.31\textwidth]{fig/goal1_Reward_Performance.png} \\
            \includegraphics[width=0.31\textwidth]{fig/goal1_Cost_Performance.png} \\
            \includegraphics[width=0.31\textwidth]{fig/goal1_Cost_Rate_Performance.png}
        }
        \label{fig:point-hazard-1}
    }
    \subfigure[Point\_4Hazard]{
        \centering
        \parbox{0.31\textwidth}{
            \includegraphics[width=0.31\textwidth]{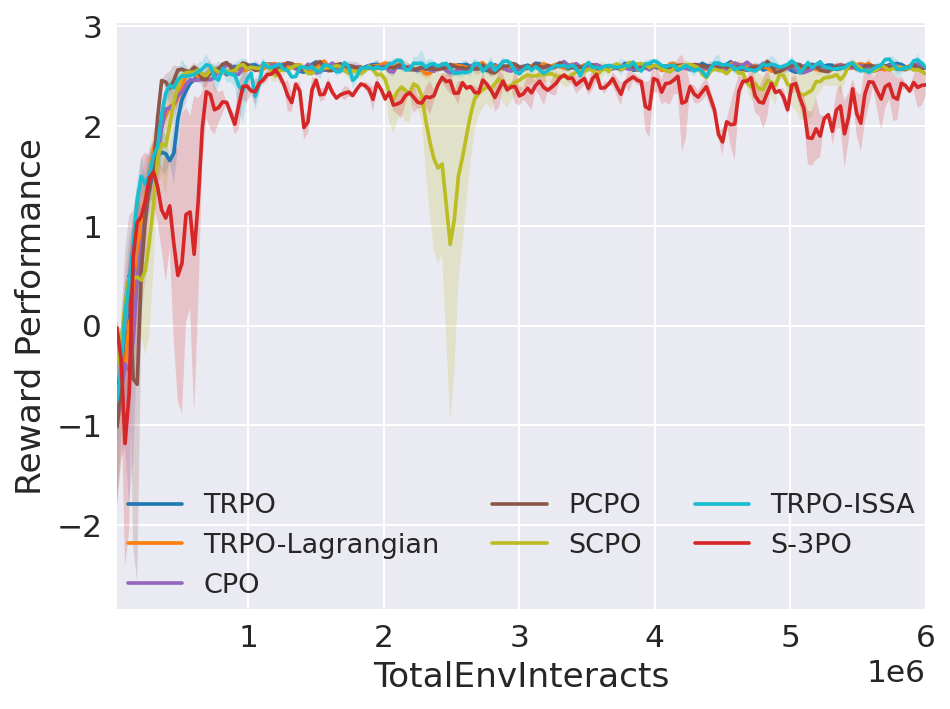} \\
            \includegraphics[width=0.31\textwidth]{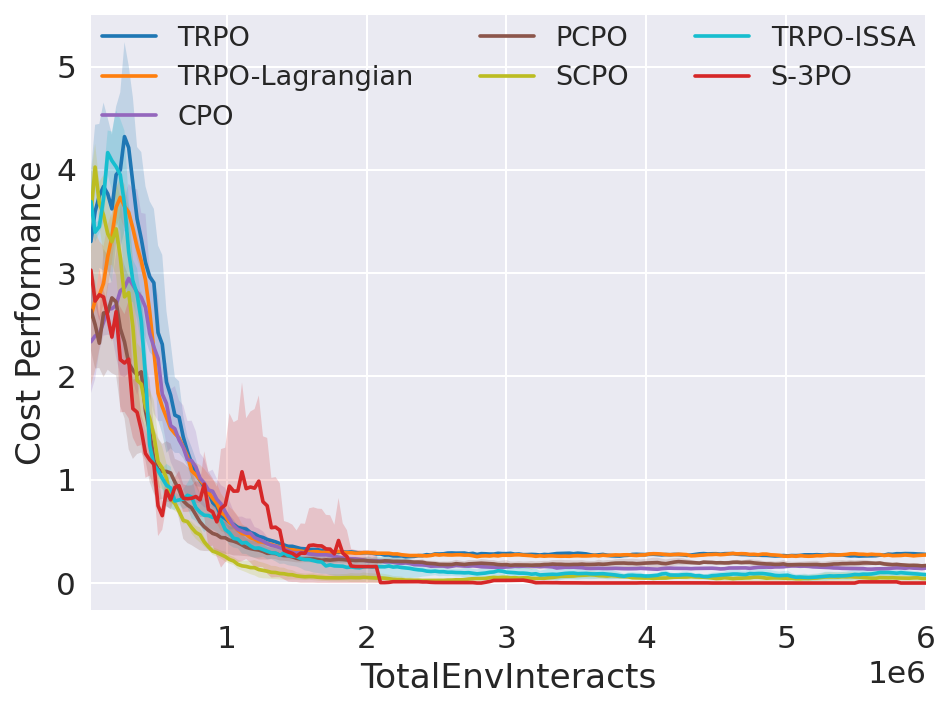} \\
            \includegraphics[width=0.31\textwidth]{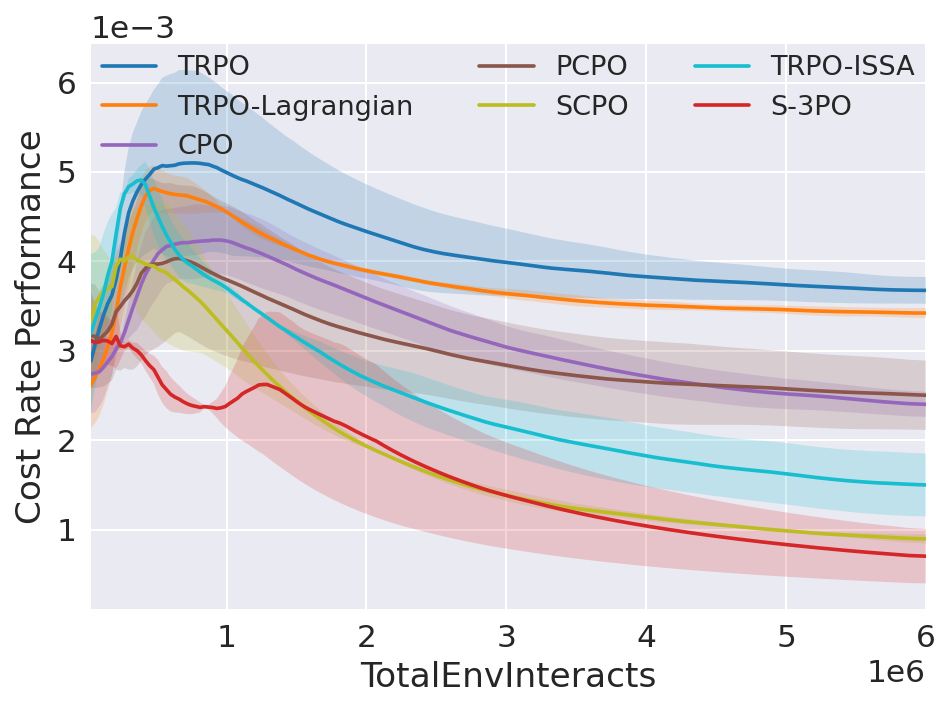}
        }
        \label{fig:point-hazard-4}
    }
    \subfigure[Point\_8Hazard]{
        \centering
        \parbox{0.31\textwidth}{
            \includegraphics[width=0.31\textwidth]{fig/goal8_Reward_Performance.png} \\
            \includegraphics[width=0.31\textwidth]{fig/goal8_Cost_Performance.png} \\
            \includegraphics[width=0.31\textwidth]{fig/goal8_Cost_Rate_Performance.png}
        }
        \label{fig:point-hazard-8}
    }
    \caption{Evaluation performance of Point\_Hazard} 
    \label{fig:exp-point-hazard-eval}
\end{figure}

\begin{figure}
    \centering
    \subfigure[Point\_1Hazard]{
        \centering
        \parbox{0.31\textwidth}{
            \includegraphics[width=0.31\textwidth]{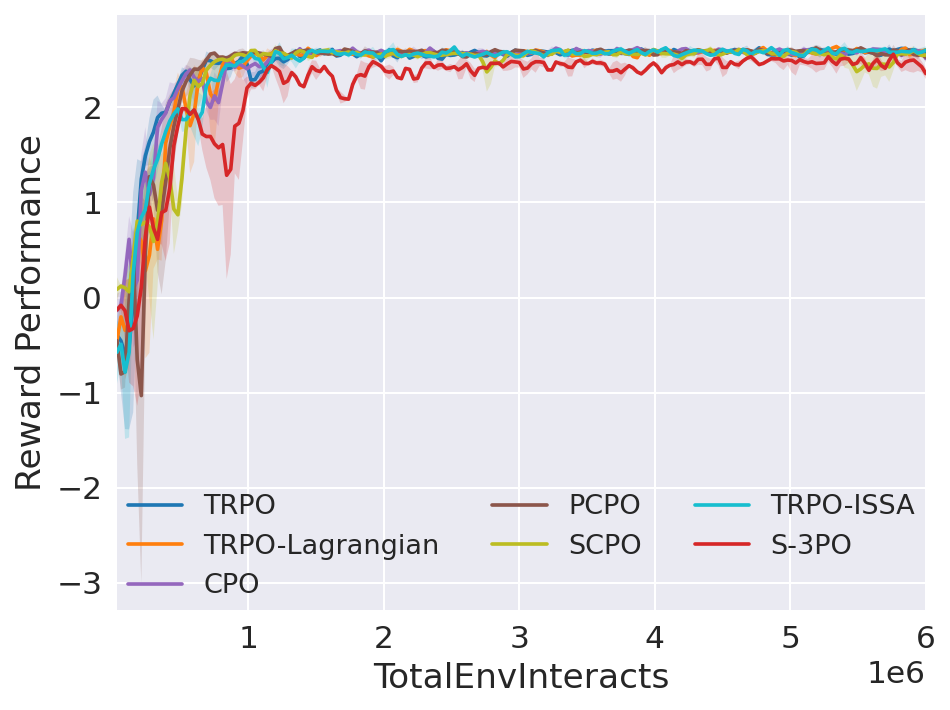} \\
            \includegraphics[width=0.31\textwidth]{fig/goal1_Cost_Performance_Train.png} \\
            \includegraphics[width=0.31\textwidth]{fig/goal1_Cost_Rate_Performance_Train.png}
        }
        \label{fig:point-hazard-1-train}
    }
    \subfigure[Point\_4Hazard]{
        \centering
        \parbox{0.31\textwidth}{
            \includegraphics[width=0.31\textwidth]{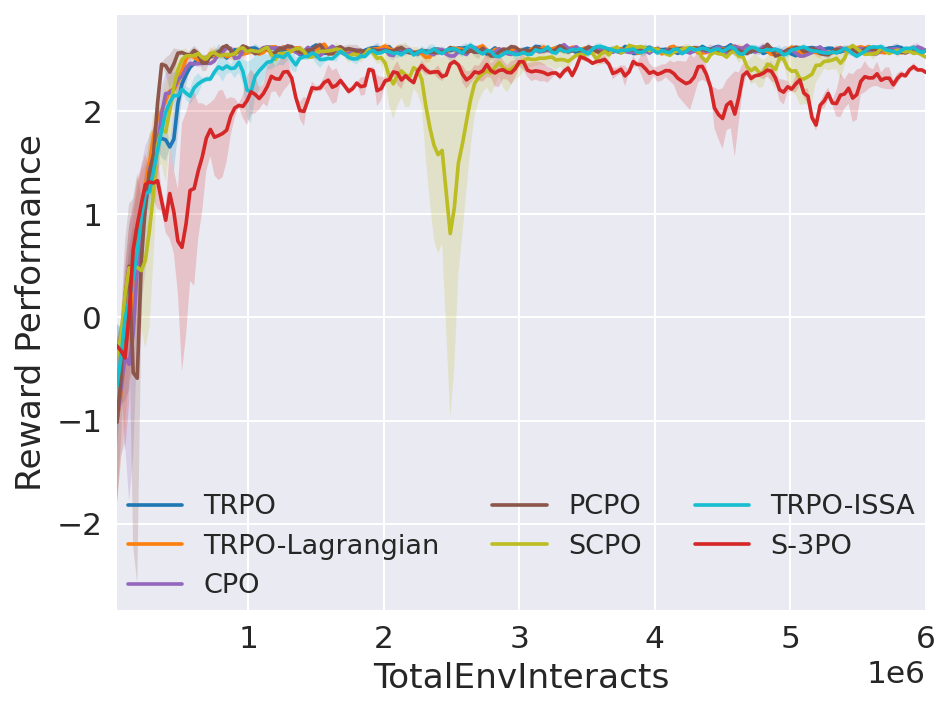} \\
            \includegraphics[width=0.31\textwidth]{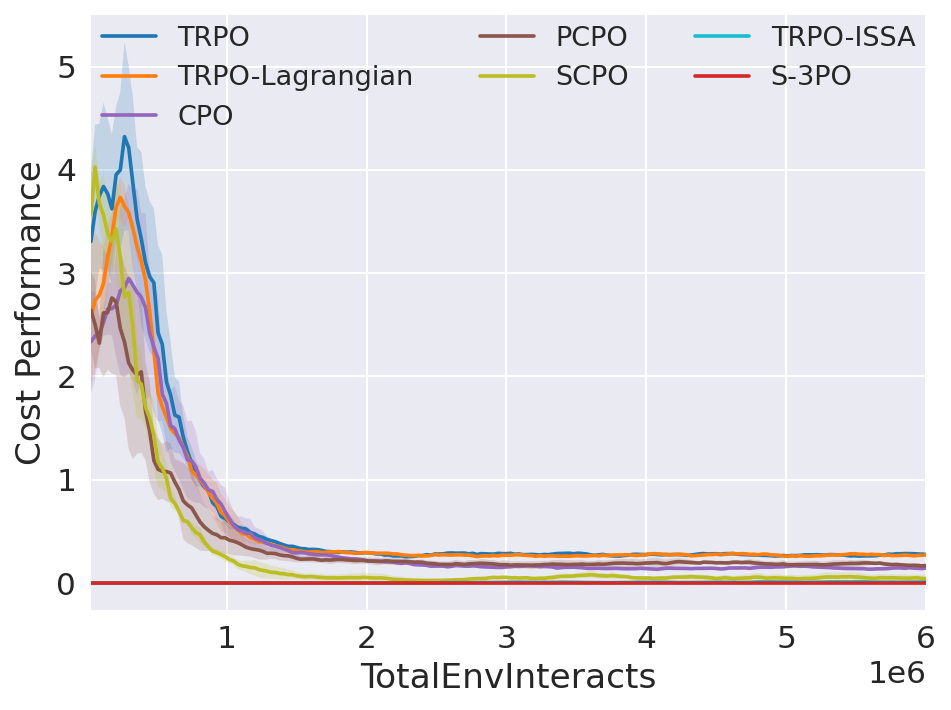} \\
            \includegraphics[width=0.31\textwidth]{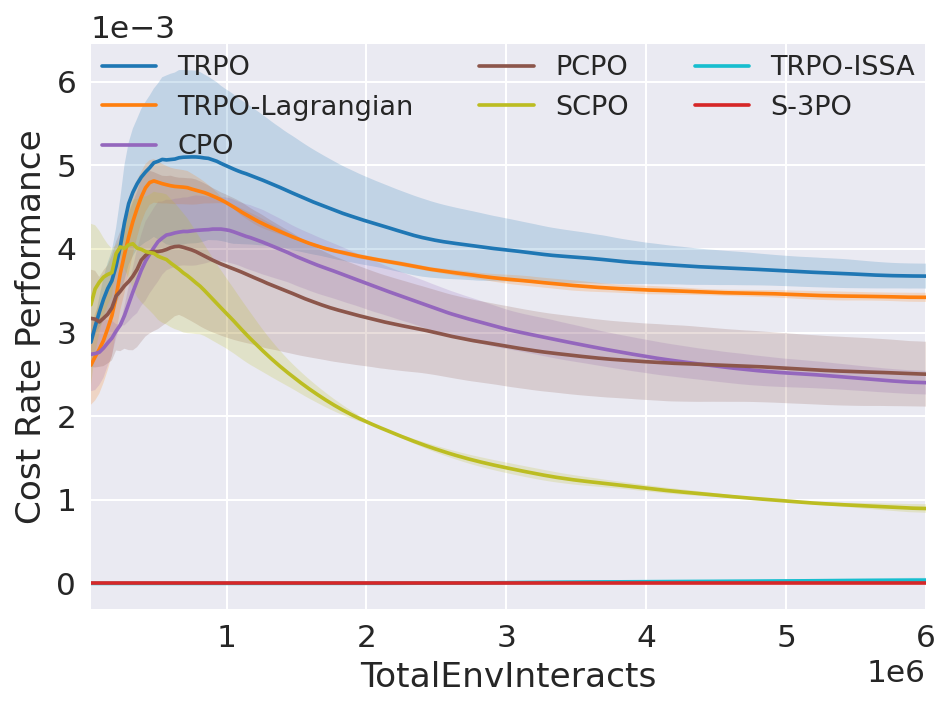}
        }
        \label{fig:point-hazard-4-train}
    }
    \subfigure[Point\_8Hazard]{
        \centering
        \parbox{0.31\textwidth}{
            \includegraphics[width=0.31\textwidth]{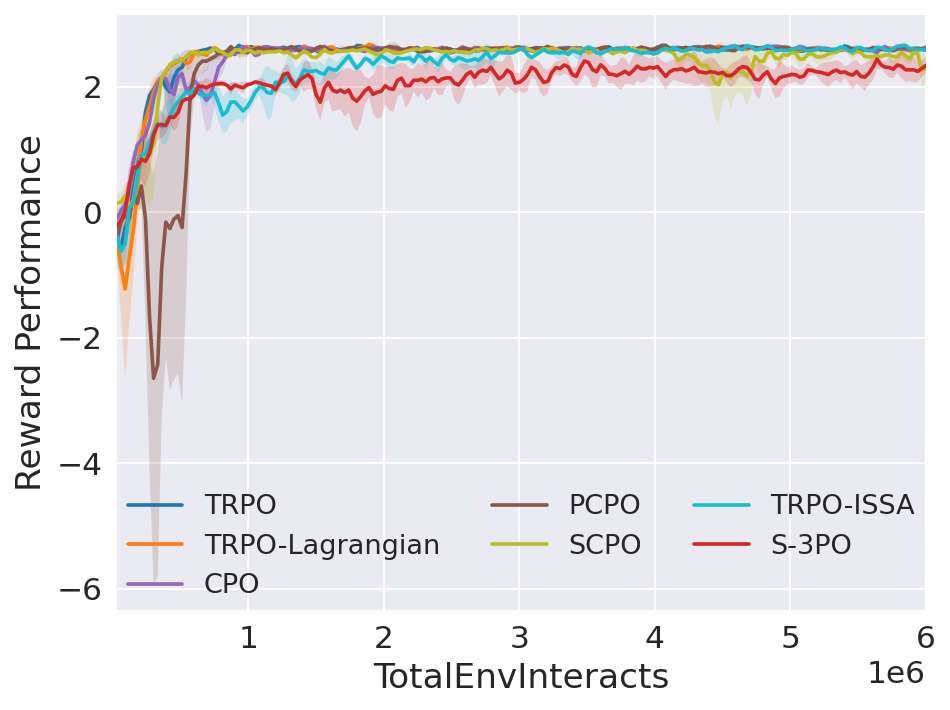} \\
            \includegraphics[width=0.31\textwidth]{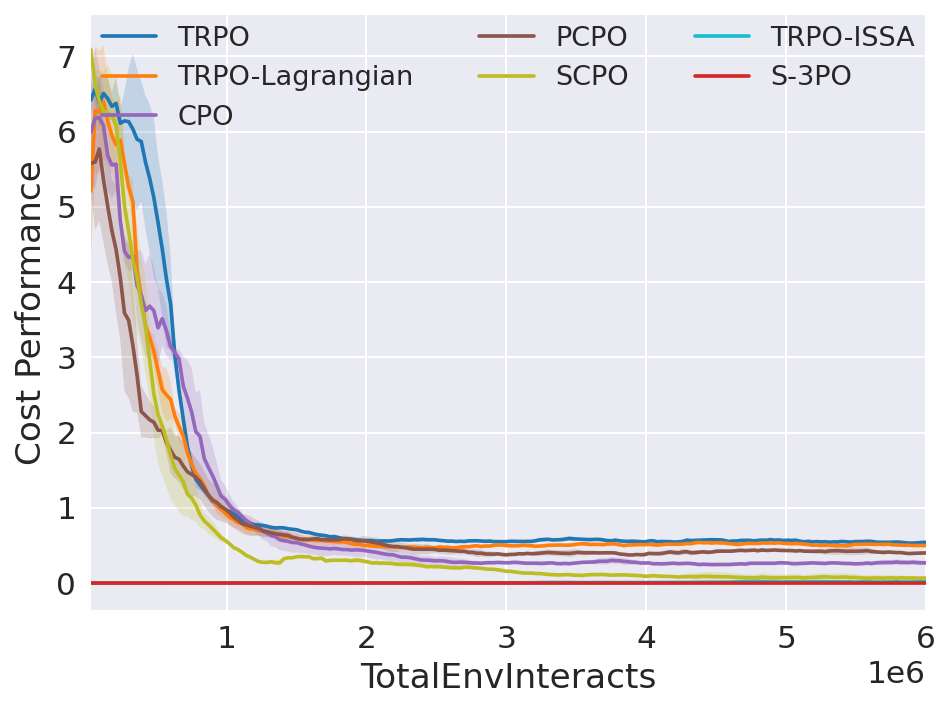} \\
            \includegraphics[width=0.31\textwidth]{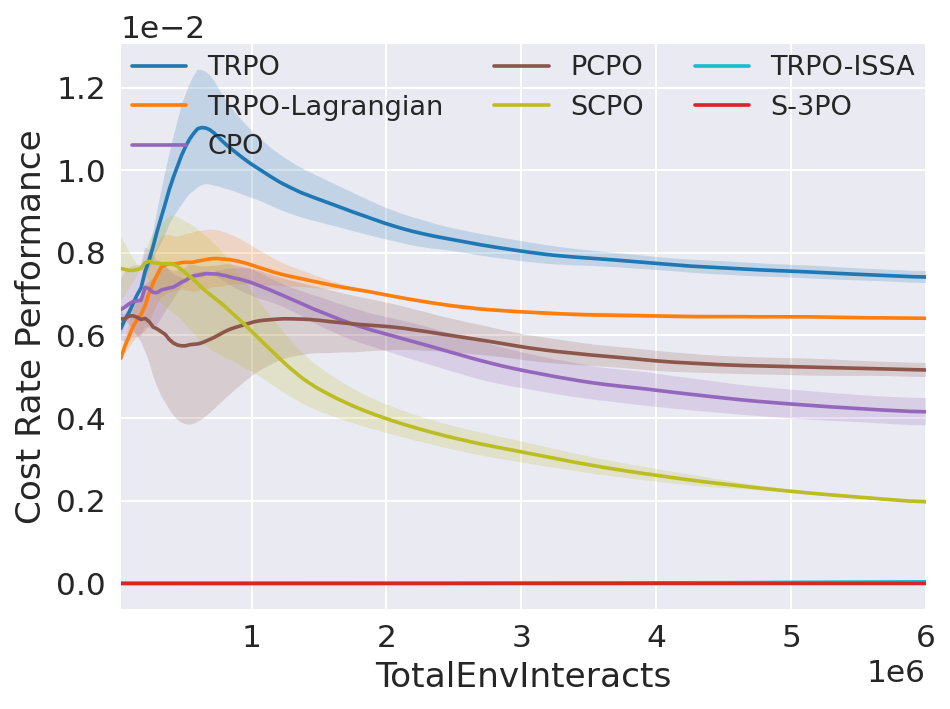}
        }
        \label{fig:point-hazard-8-train}
    }
    \caption{Training performance of Point\_Hazard} 
    \label{fig:exp-point-hazard-train}
    \vspace{-5\baselineskip}
\end{figure}

\begin{figure}
    \centering
    \subfigure[Point\_1Hazard]{
        \centering
        \parbox{0.31\textwidth}{
            \includegraphics[width=0.31\textwidth]{fig/goal1_ISSA_Trigger_Times.png}
        }
        \label{fig:goal-point-1hazard-ISSA}
    }
    \subfigure[Point\_4Hazard]{
        \centering
        \parbox{0.31\textwidth}{
            \includegraphics[width=0.31\textwidth]{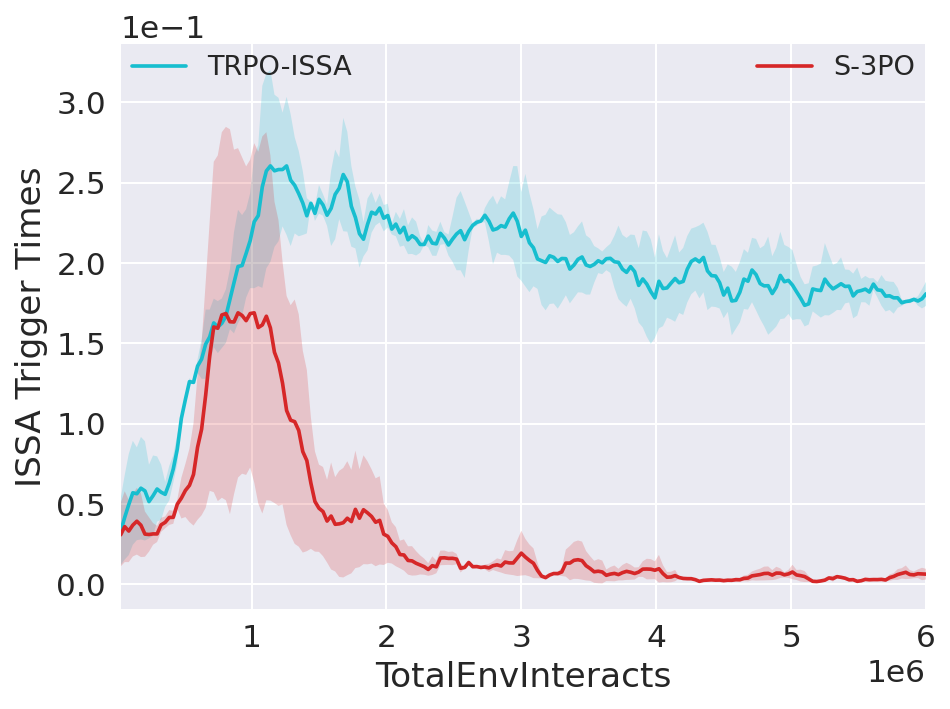}
        }
        \label{fig:goal-point-4hazard-ISSA}
    }
    \subfigure[Point\_8Hazard]{
        \centering
        \parbox{0.31\textwidth}{
            \includegraphics[width=0.31\textwidth]{fig/goal8_ISSA_Trigger_Times.png}
        }
        \label{fig:goal-point-8hazard-ISSA}
    }
    \caption{ISSA performance of Point\_Hazard}
    \label{fig:exp-point-hazard-ISSA}
\end{figure}

\begin{figure}
    \centering
    \subfigure[Point\_1Pillar]{
        \centering
        \parbox{0.31\textwidth}{
            \includegraphics[width=0.31\textwidth]{fig/goal1_pillar_Reward_Performance.png} \\
            \includegraphics[width=0.31\textwidth]{fig/goal1_pillar_Cost_Performance.png} \\
            \includegraphics[width=0.31\textwidth]{fig/goal1_pillar_Cost_Rate_Performance.png}
        }
        \label{fig:point-pillar-1}
    }
    \subfigure[Point\_4Pillar]{
        \centering
        \parbox{0.31\textwidth}{
            \includegraphics[width=0.31\textwidth]{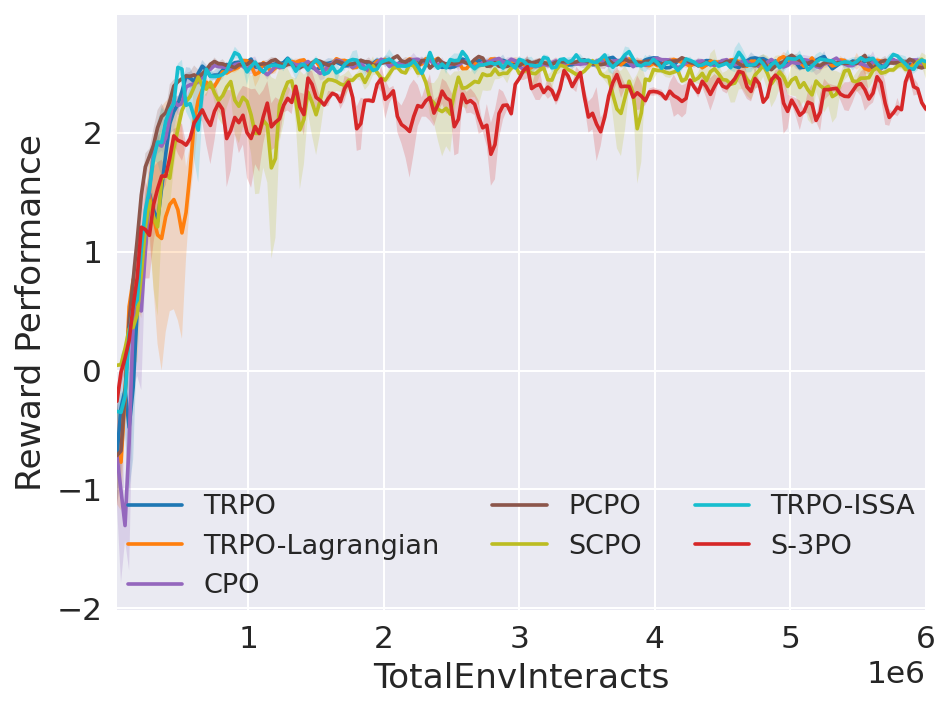} \\
            \includegraphics[width=0.31\textwidth]{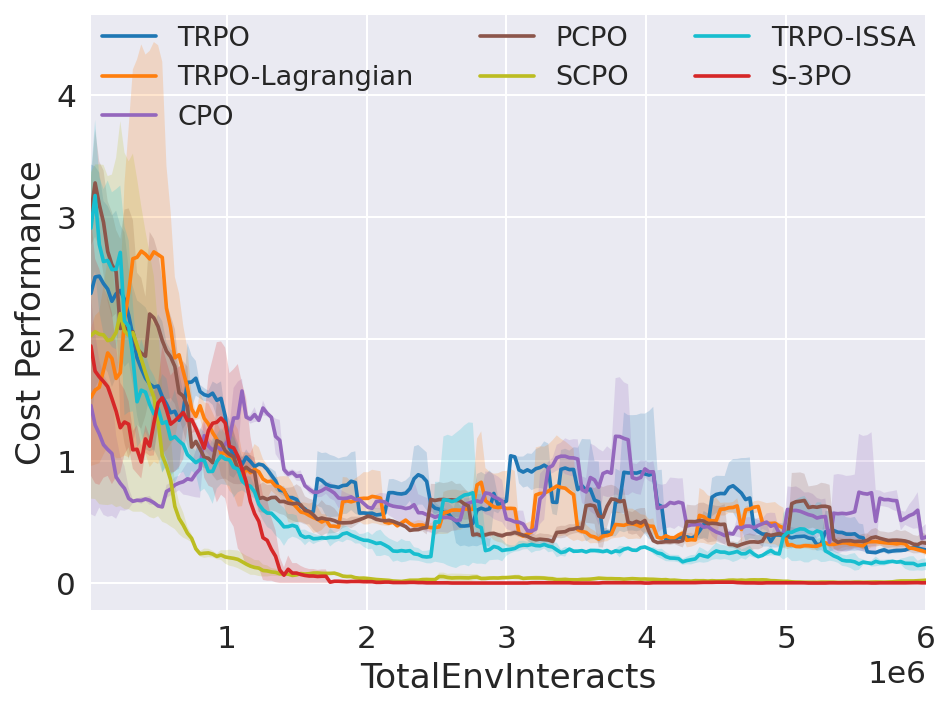} \\
            \includegraphics[width=0.31\textwidth]{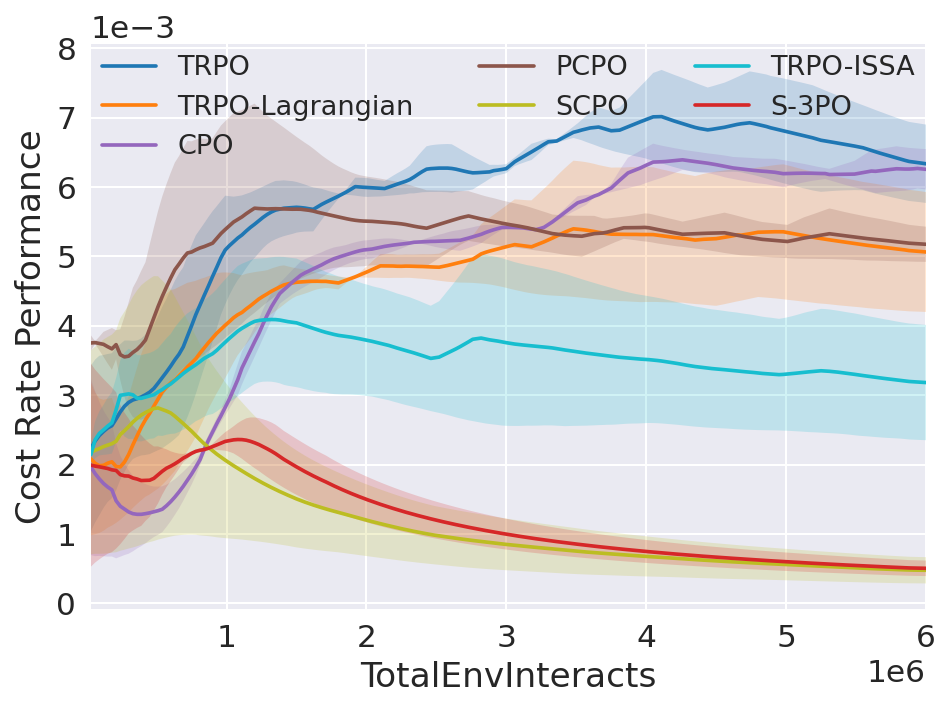}
        }
        \label{fig:point-pillar-4}
    }
    \subfigure[Point\_8Pillar]{
        \centering
        \parbox{0.31\textwidth}{
            \includegraphics[width=0.31\textwidth]{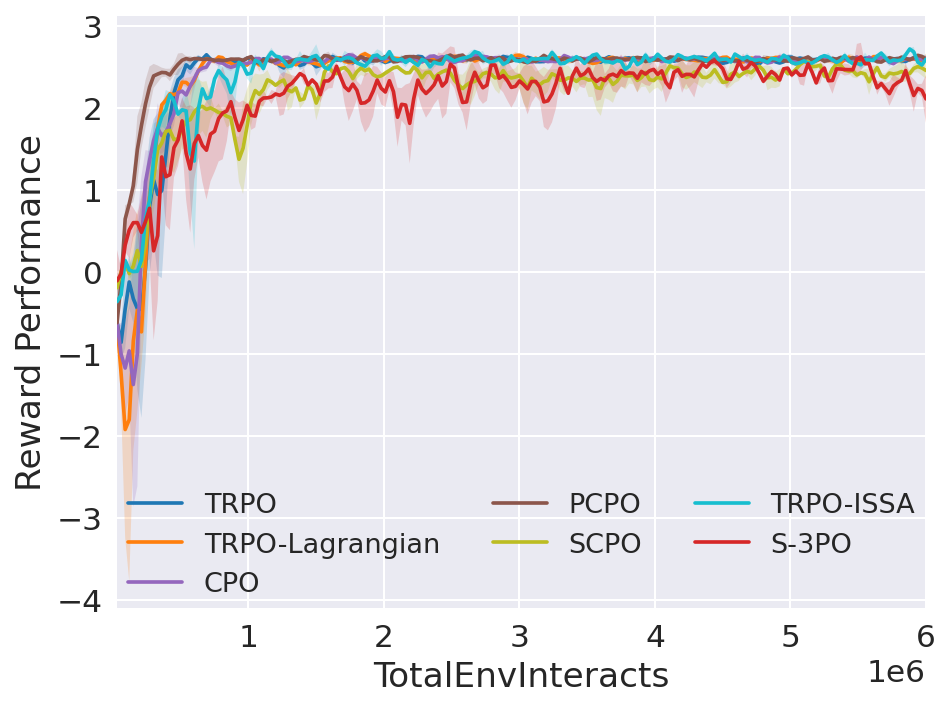} \\
            \includegraphics[width=0.31\textwidth]{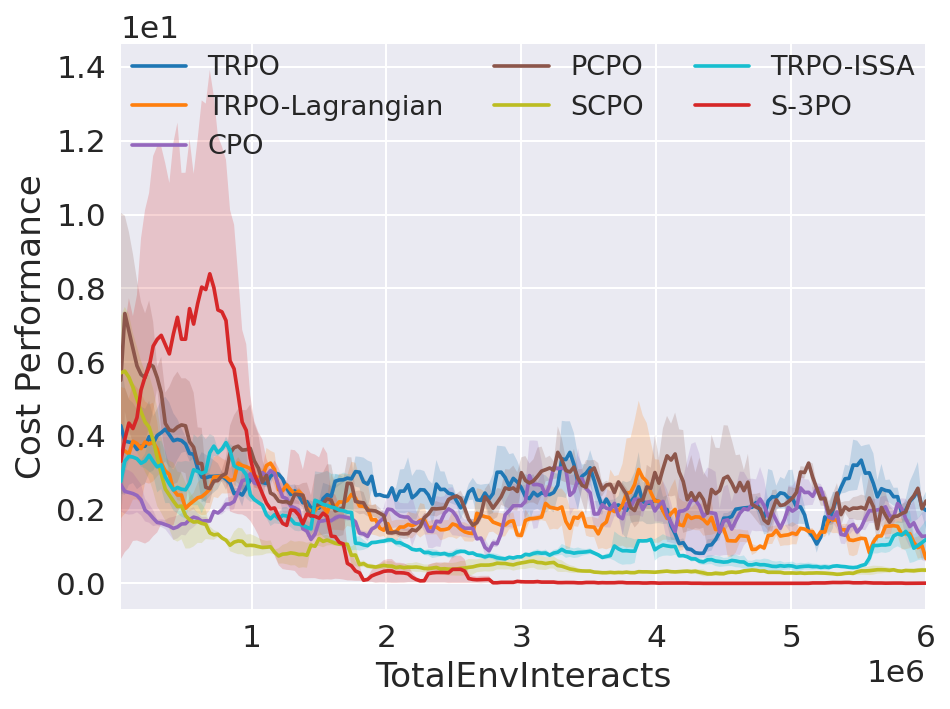} \\
            \includegraphics[width=0.31\textwidth]{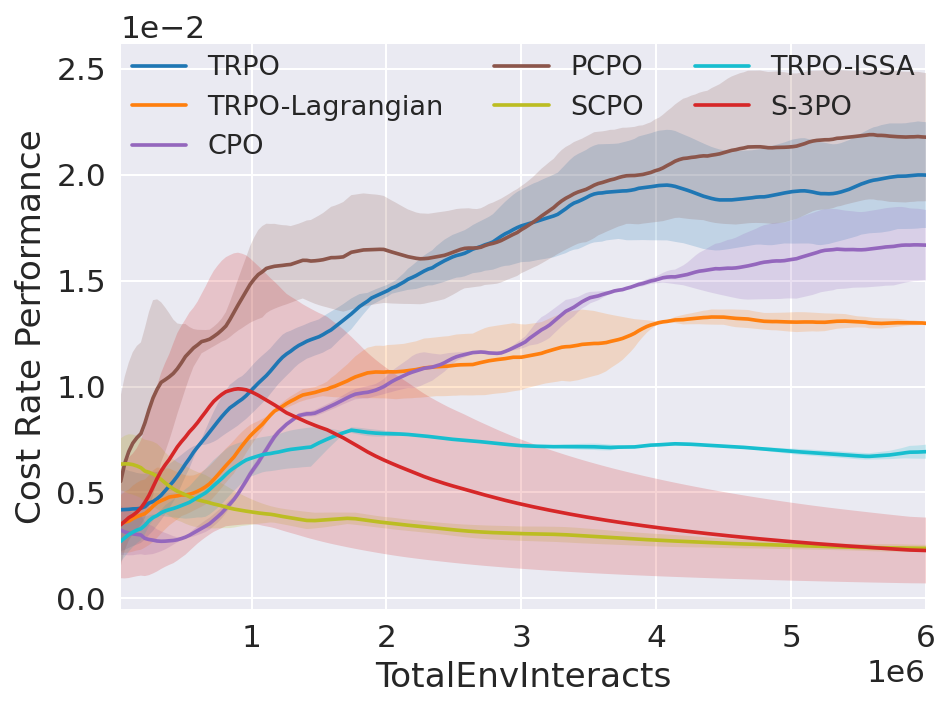}
        }
        \label{fig:point-pillar-8}
    }
    \caption{Evaluation performance of Point\_Pillar}
    \label{fig:exp-point-pillar-eval}
\end{figure}

\begin{figure}
    \centering
    \subfigure[Point\_1Pillar]{
        \centering
        \parbox{0.31\textwidth}{
            \includegraphics[width=0.31\textwidth]{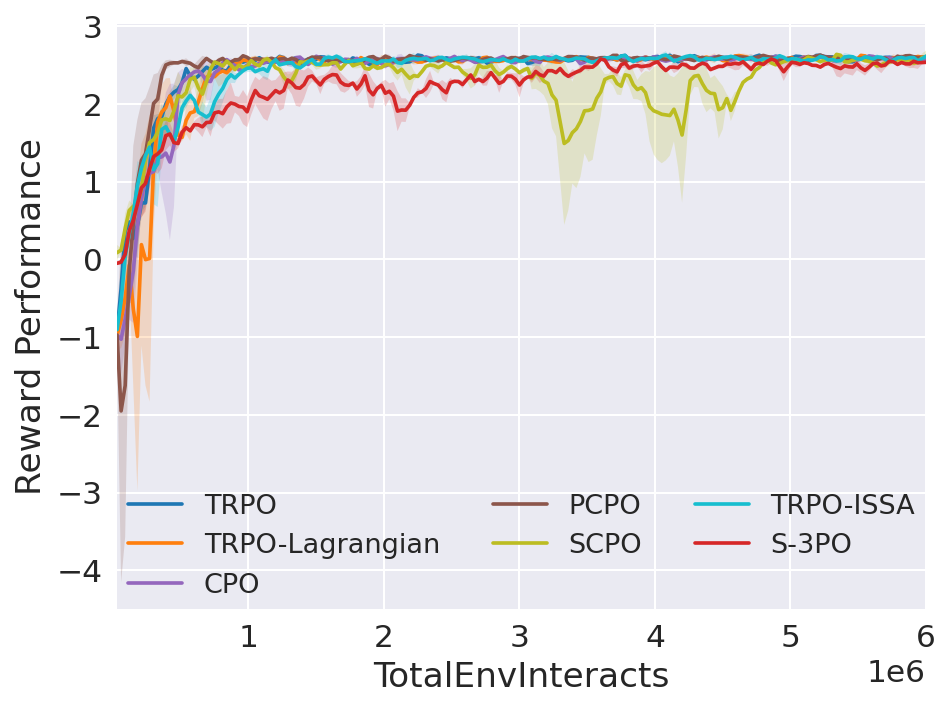} \\
            \includegraphics[width=0.31\textwidth]{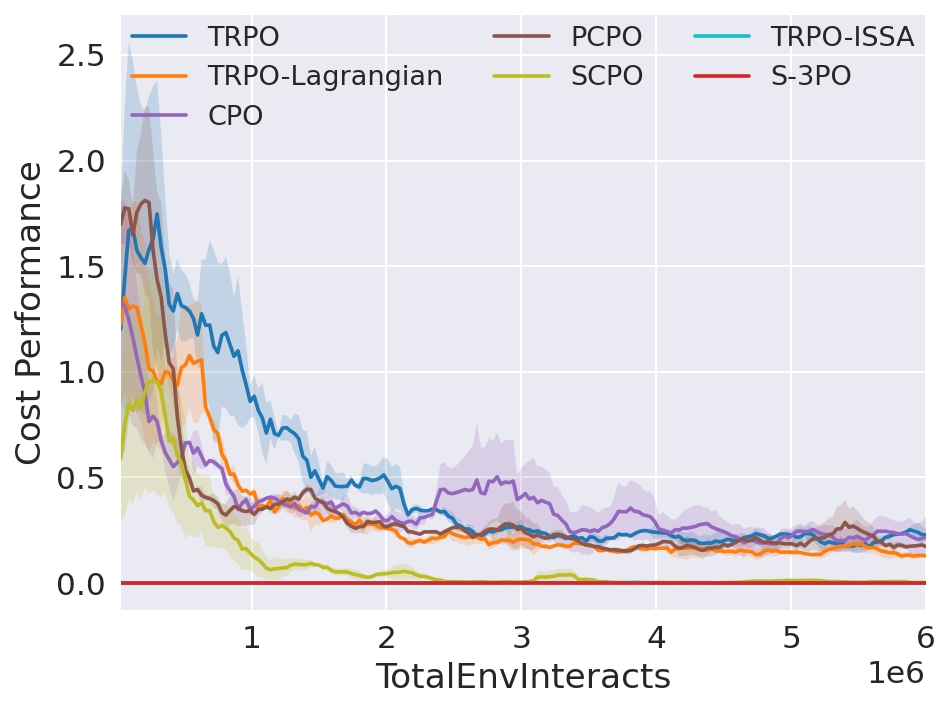} \\
            \includegraphics[width=0.31\textwidth]{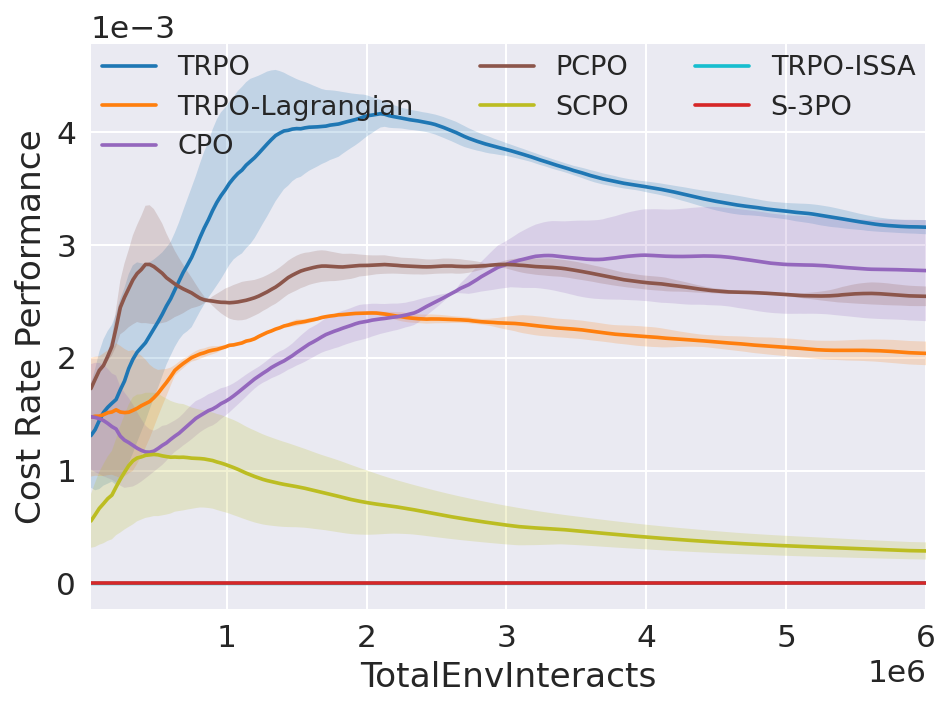}
        }
        \label{fig:point-pillar-1-train}
    }
    \subfigure[Point\_4Pillar]{
        \centering
        \parbox{0.31\textwidth}{
            \includegraphics[width=0.31\textwidth]{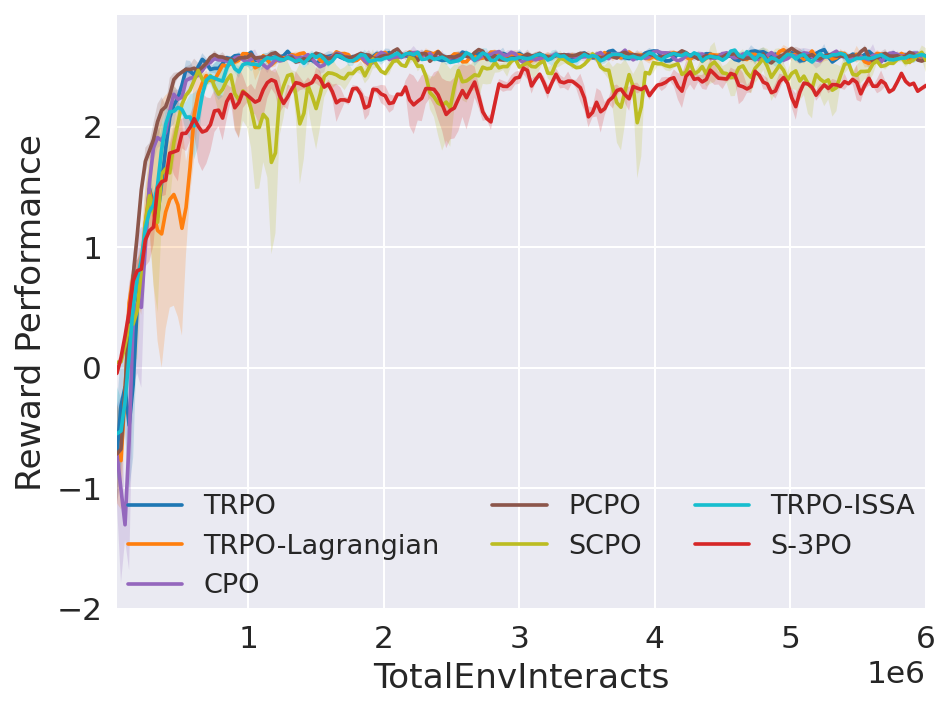} \\
            \includegraphics[width=0.31\textwidth]{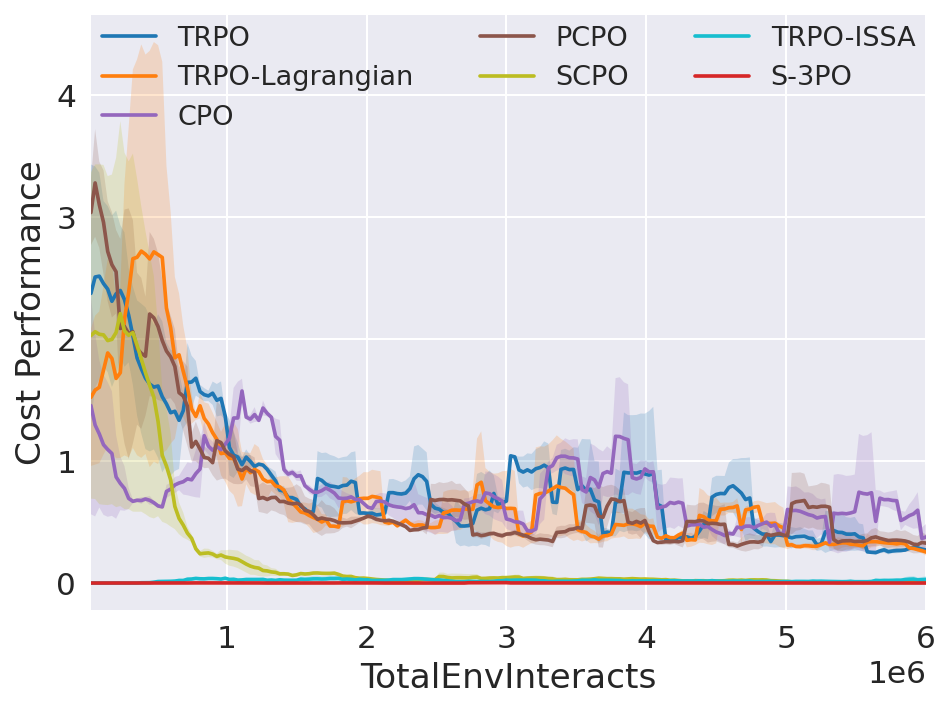} \\
            \includegraphics[width=0.31\textwidth]{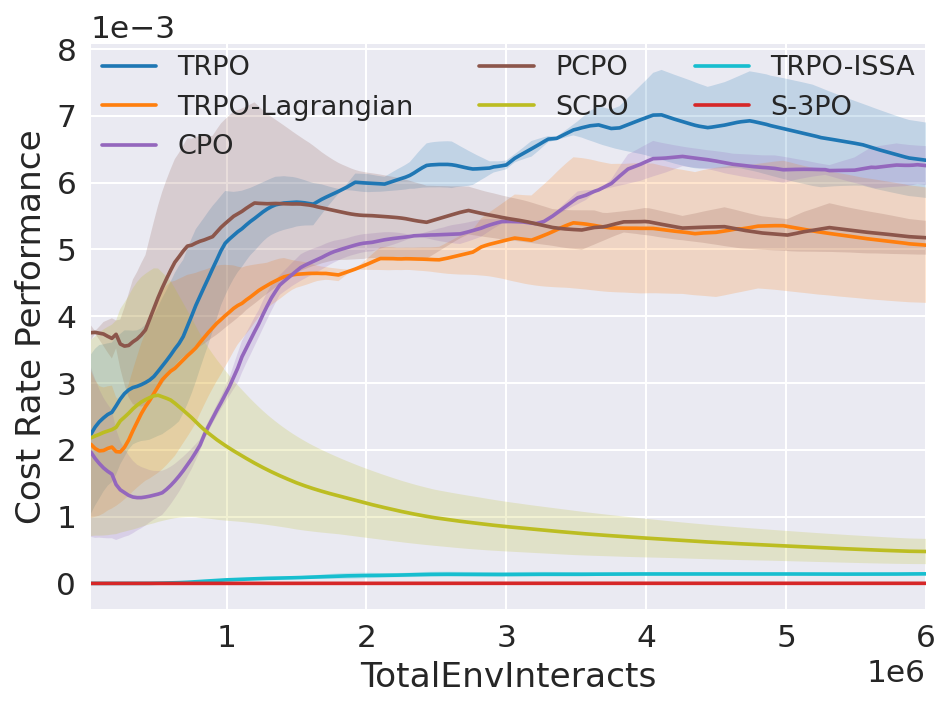}
        }
        \label{fig:point-pillar-4-train}
    }
    \subfigure[Point\_8Pillar]{
        \centering
        \parbox{0.31\textwidth}{
            \includegraphics[width=0.31\textwidth]{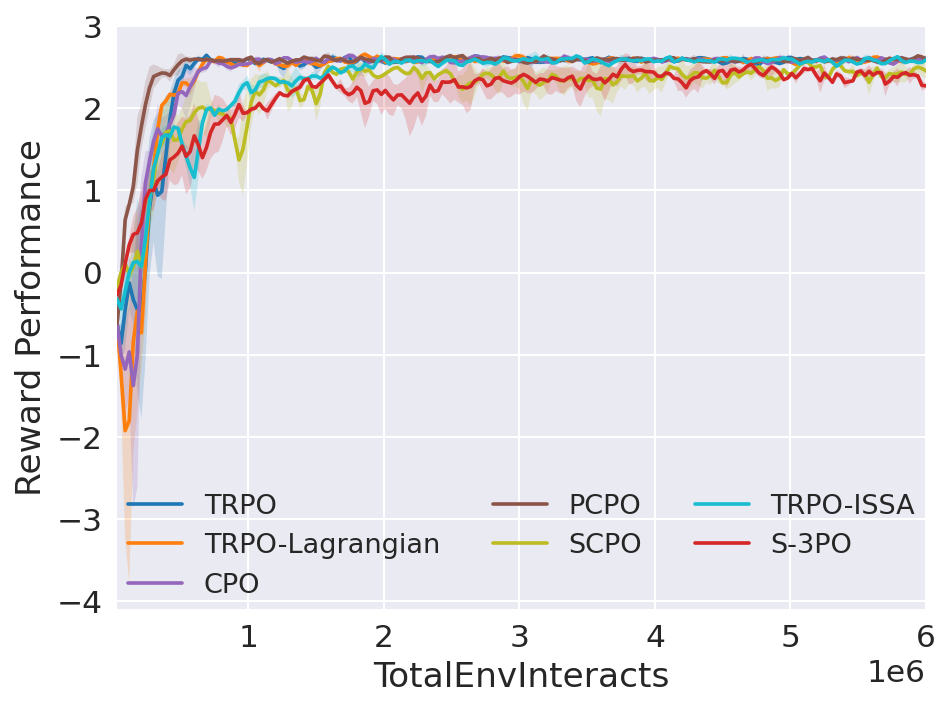} \\
            \includegraphics[width=0.31\textwidth]{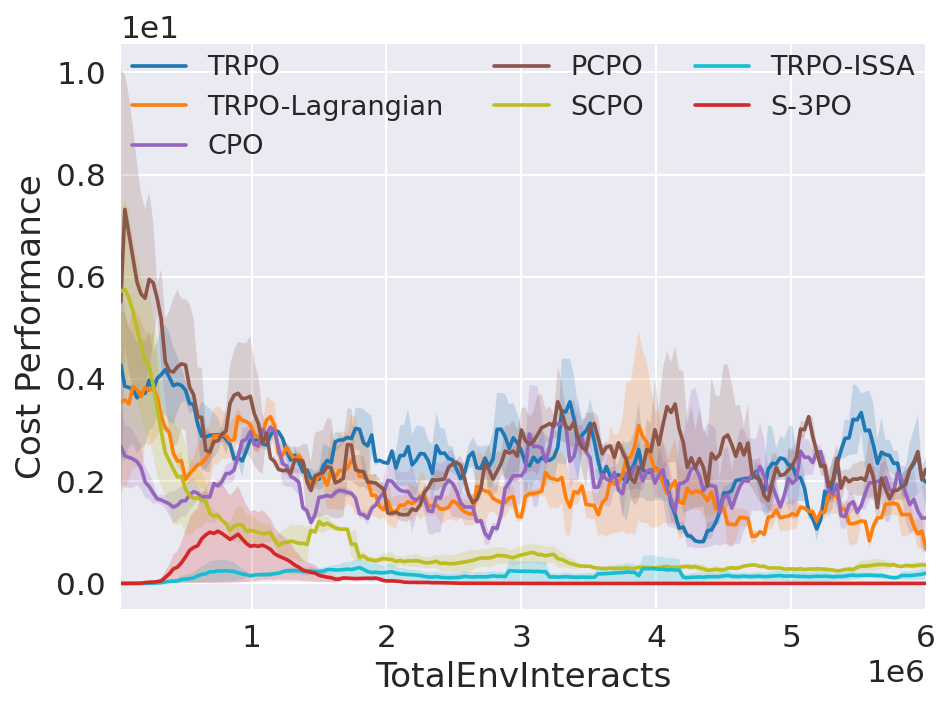} \\
            \includegraphics[width=0.31\textwidth]{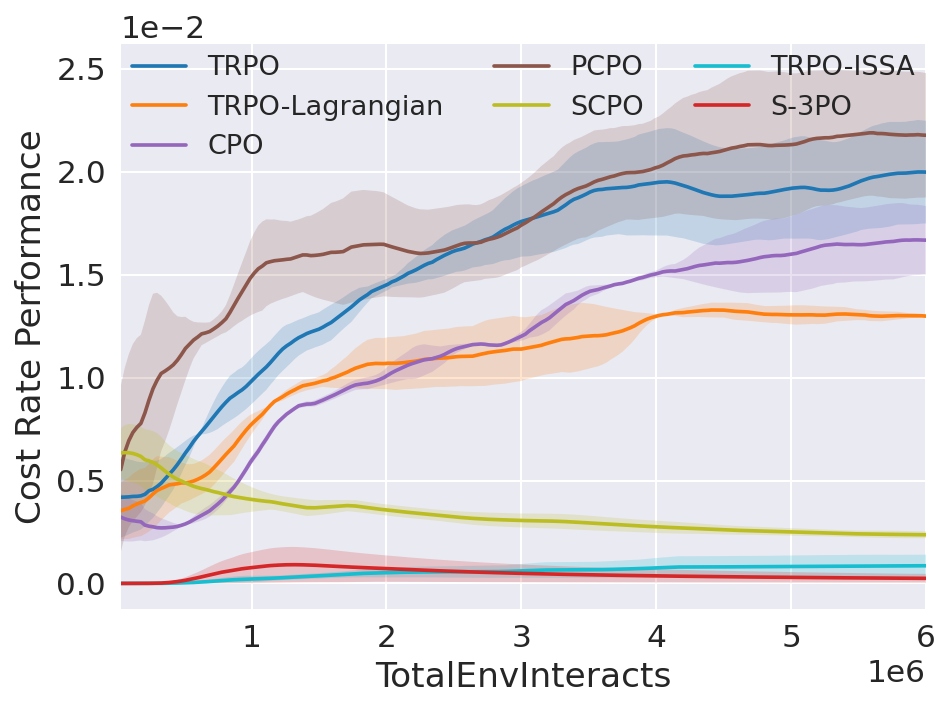}
        }
        \label{fig:point-pillar-8-train}
    }
    \caption{Training performance of Point\_Pillar}
    \label{fig:exp-point-pillar-train}
    \vspace{5\baselineskip}
\end{figure}

\begin{figure}
    \centering
    \subfigure[Point\_1Pillar]{
        \centering
        \parbox{0.31\textwidth}{
            \includegraphics[width=0.31\textwidth]{fig/goal1_pillar_ISSA_Trigger_Times.png}
        }
        \label{fig:goal-point-1pillar-ISSA}
    }
    \subfigure[Point\_4Pillar]{
        \centering
        \parbox{0.31\textwidth}{
            \includegraphics[width=0.31\textwidth]{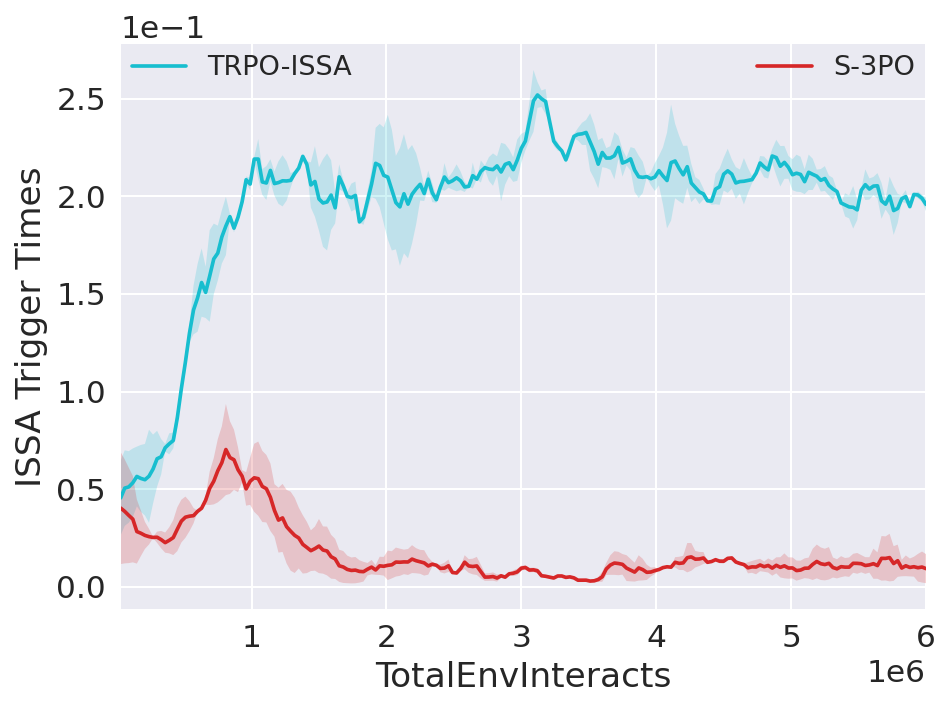} 
        }
        \label{fig:goal-point-4pillar-ISSA}
    }
    \subfigure[Point\_8Pillar]{
        \centering
        \parbox{0.31\textwidth}{
            \includegraphics[width=0.31\textwidth]{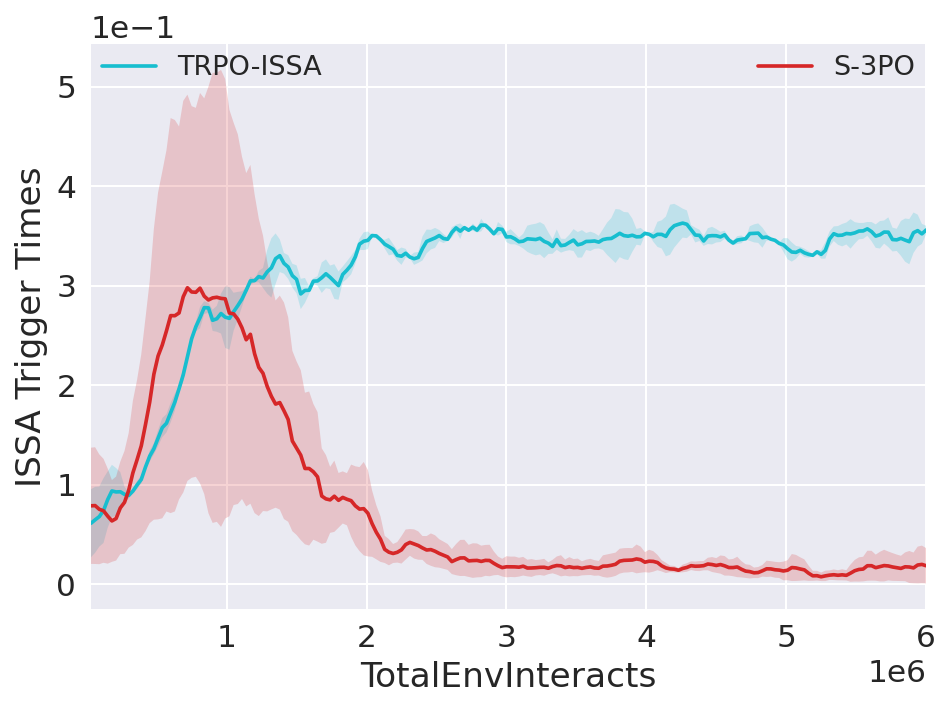}
        }
        \label{fig:goal-point-8pillar-ISSA}
    }
    \caption{ISSA performance of Point\_Pillar}
    \label{fig:exp-point-pillar-ISSA}
\end{figure}

\begin{figure}
    \centering
    \subfigure[Swimmer\_1Hazard]{
        \centering
        \parbox{0.31\textwidth}{
            \includegraphics[width=0.31\textwidth]{fig/swimmertiny1_Reward_Performance.png} \\
            \includegraphics[width=0.31\textwidth]{fig/swimmertiny1_Cost_Performance.png} \\
            \includegraphics[width=0.31\textwidth]{fig/swimmertiny1_Cost_Rate_Performance.png}
        }
        \label{fig:swimmer-hazard-1}
    }
    \subfigure[Ant\_1Hazard]{
        \centering
        \parbox{0.31\textwidth}{
            \includegraphics[width=0.31\textwidth]{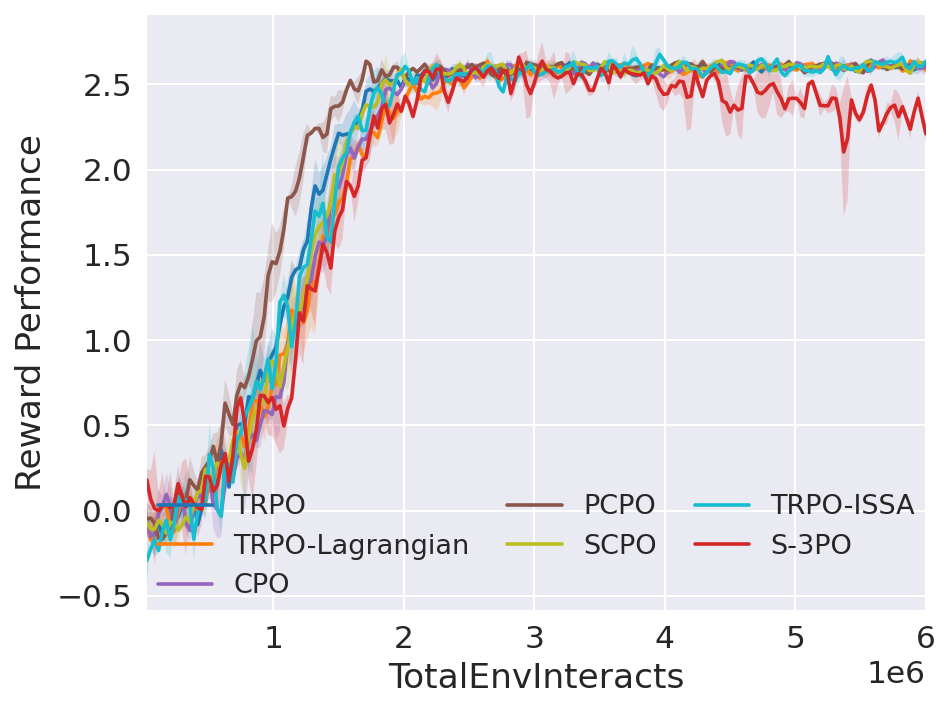} \\
            \includegraphics[width=0.31\textwidth]{fig/anttiny1_Cost_Performance.png} \\
            \includegraphics[width=0.31\textwidth]{fig/anttiny1_Cost_Rate_Performance.png}
        }
        \label{fig:ant-hazard-1}
    }
    \subfigure[Walker\_1Hazard]{
        \centering
        \parbox{0.31\textwidth}{
            \includegraphics[width=0.31\textwidth]{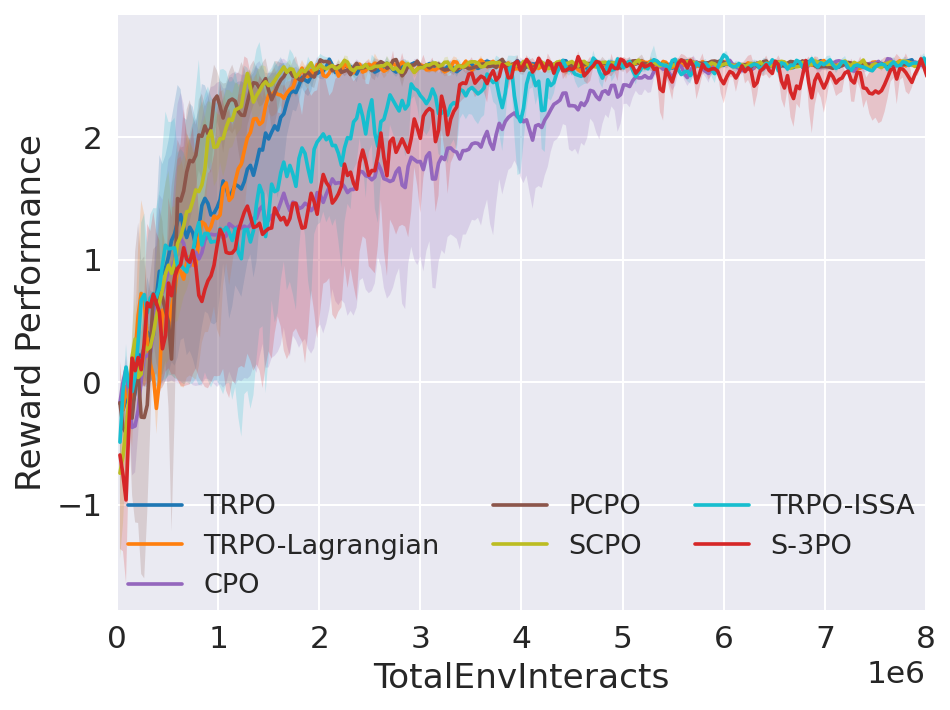} \\
            \includegraphics[width=0.31\textwidth]{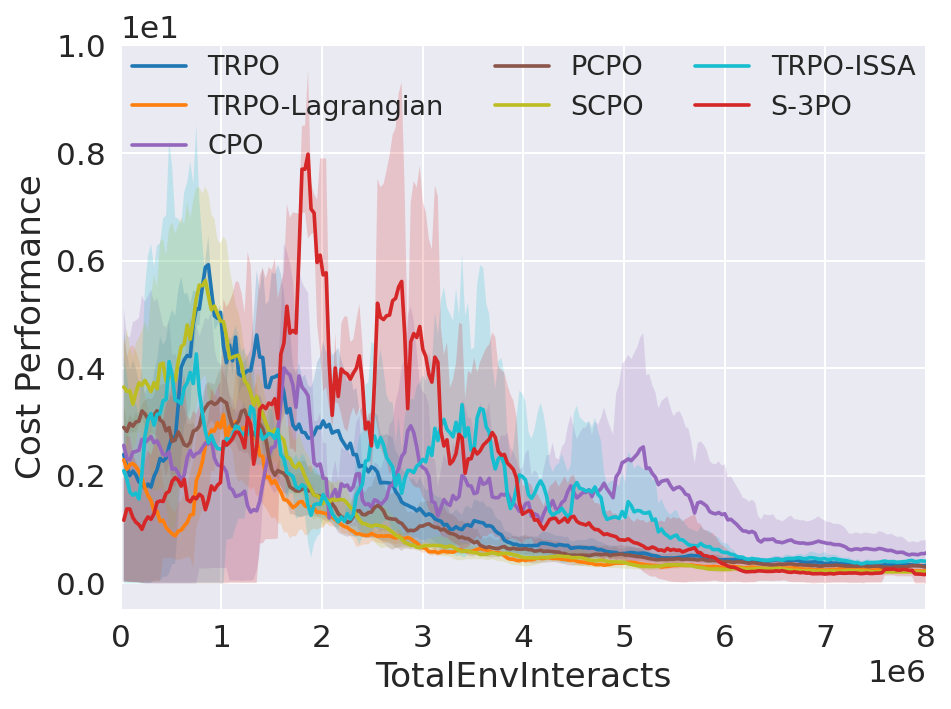} \\
            \includegraphics[width=0.31\textwidth]{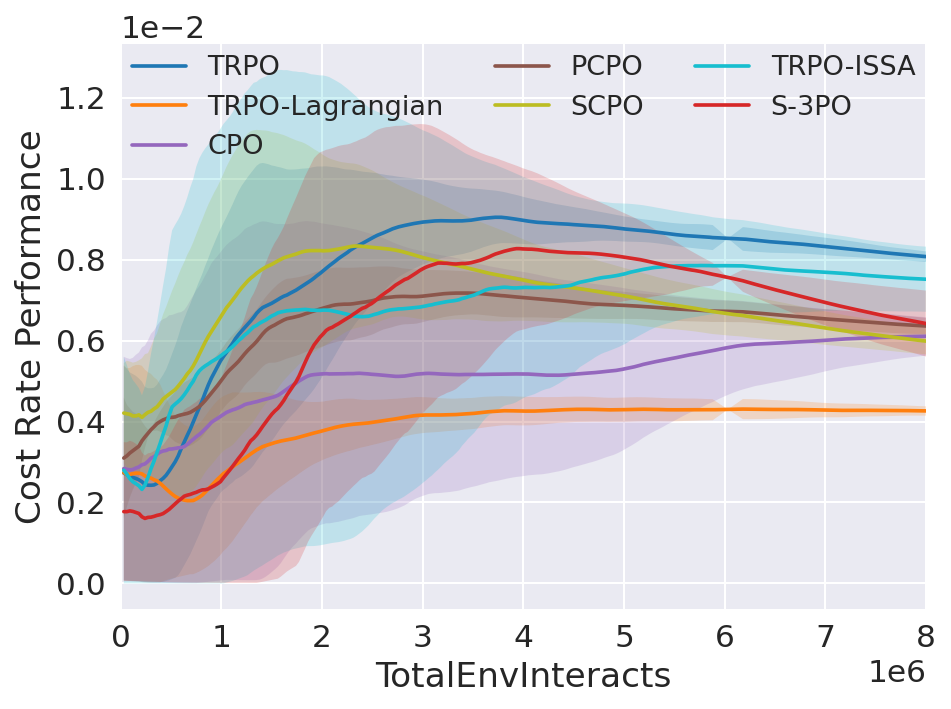}
        }
        \label{fig:walker-hazard-1}
    }
    \caption{Evaluation performance of link robots}
    \label{fig:exp-link-robot-eval}
\end{figure}

\begin{figure}
    \centering
    \subfigure[Swimmer\_1Hazard]{
        \centering
        \parbox{0.31\textwidth}{
            \includegraphics[width=0.31\textwidth]{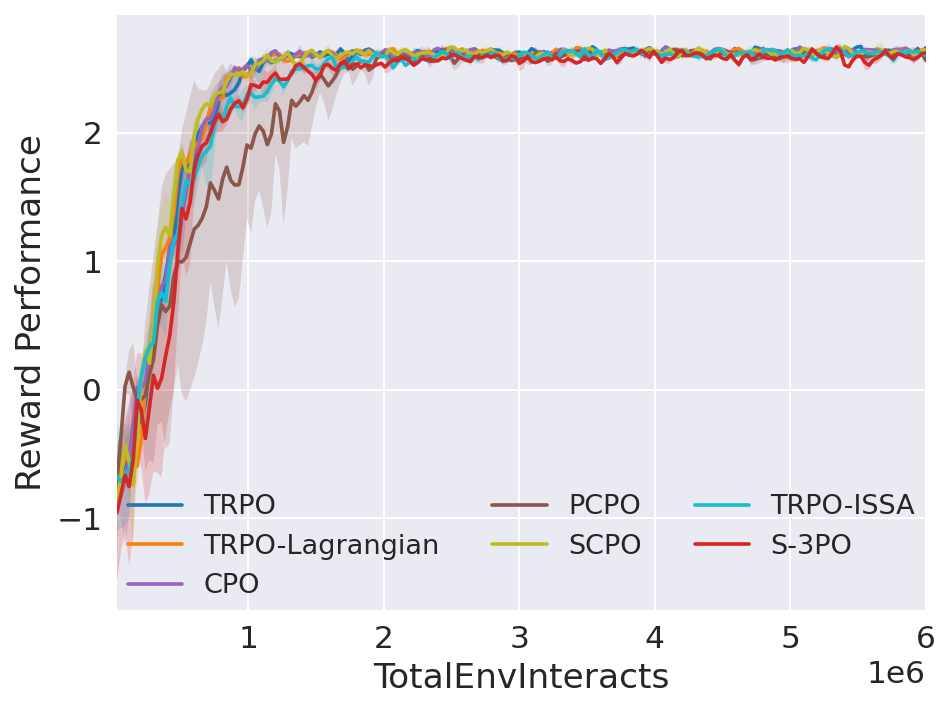} \\
            \includegraphics[width=0.31\textwidth]{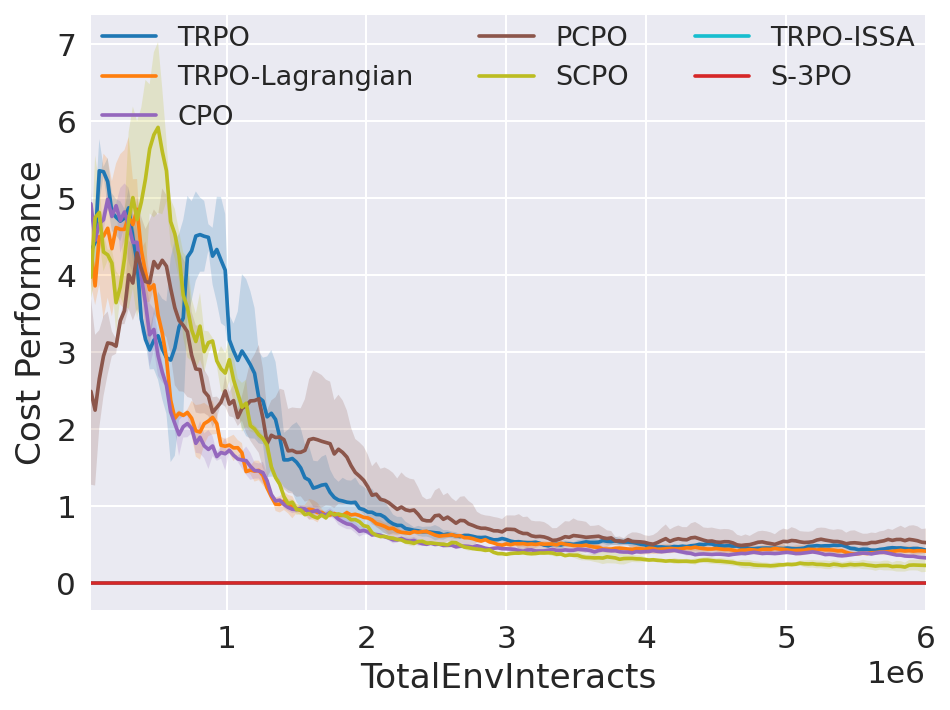} \\
            \includegraphics[width=0.31\textwidth]{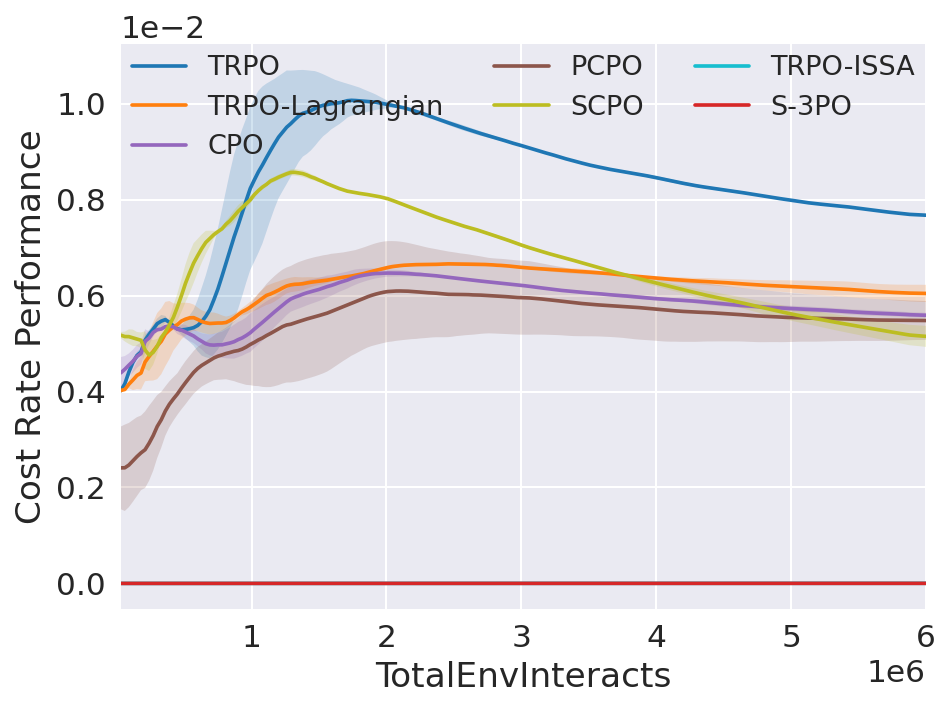}
        }
        \label{fig:swimmer-hazard-1-train}
    }
    \subfigure[Ant\_1Hazard]{
        \centering
        \parbox{0.31\textwidth}{
            \includegraphics[width=0.31\textwidth]{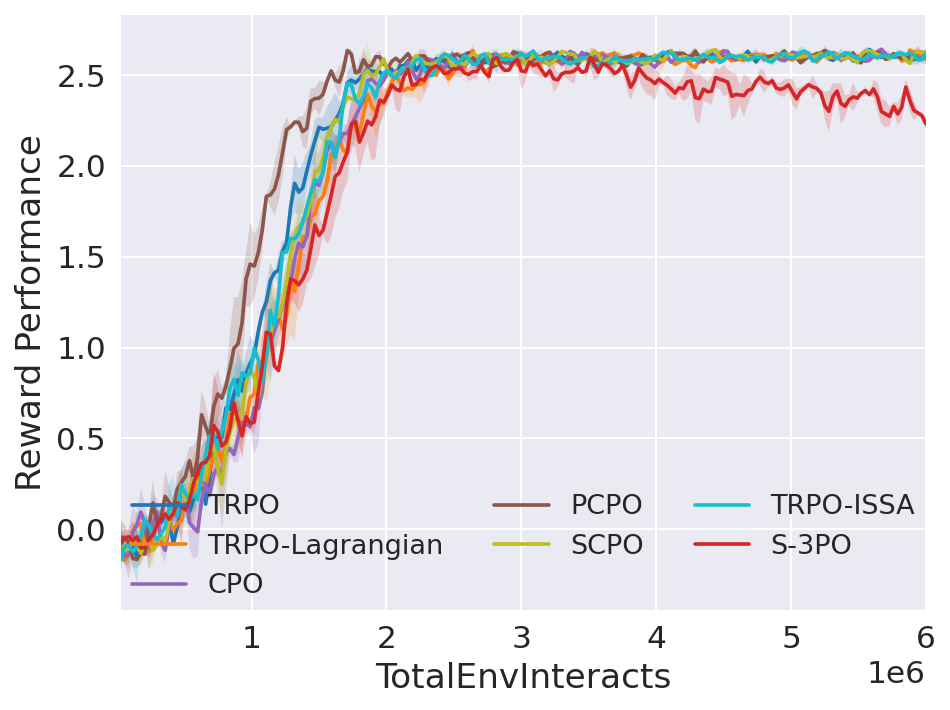} \\
            \includegraphics[width=0.31\textwidth]{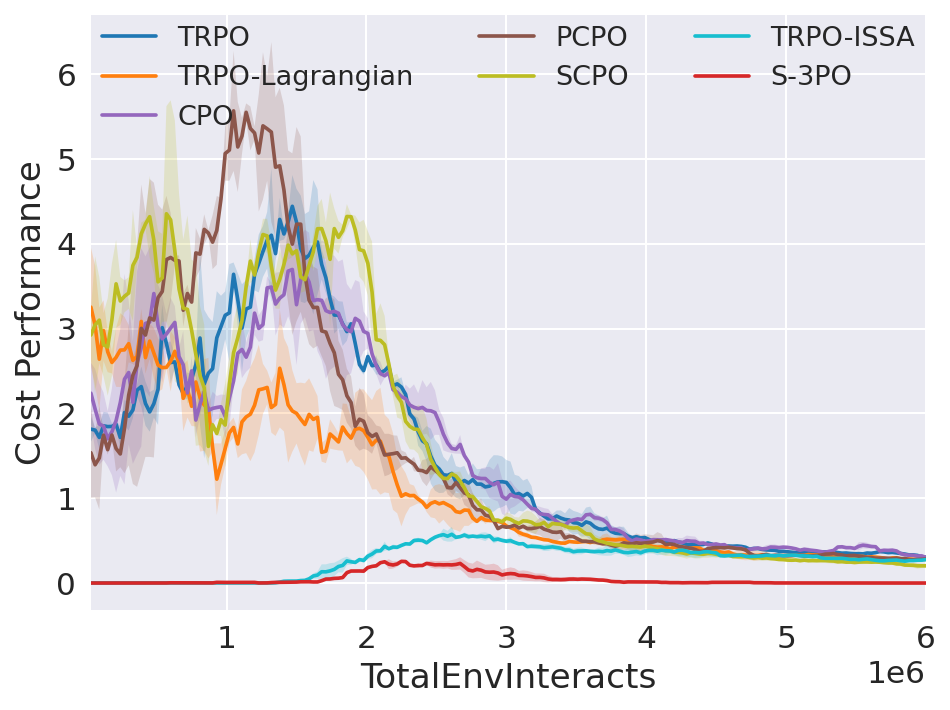} \\
            \includegraphics[width=0.31\textwidth]{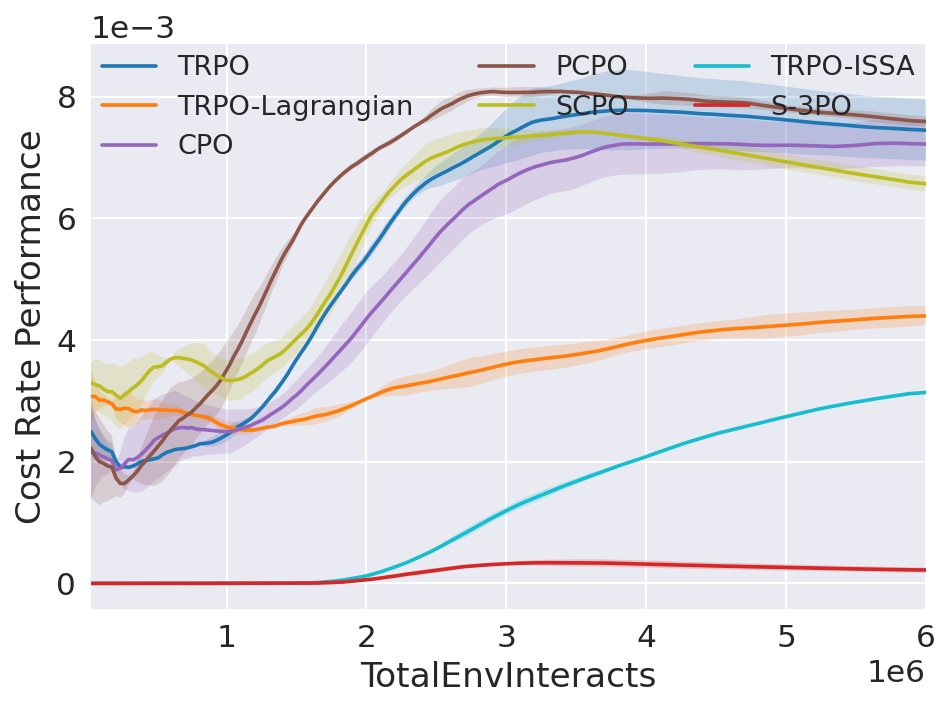}
        }
        \label{fig:ant-hazard-1-train}
    }
    \subfigure[Walker\_1Hazard]{
        \centering
        \parbox{0.31\textwidth}{
            \includegraphics[width=0.31\textwidth]{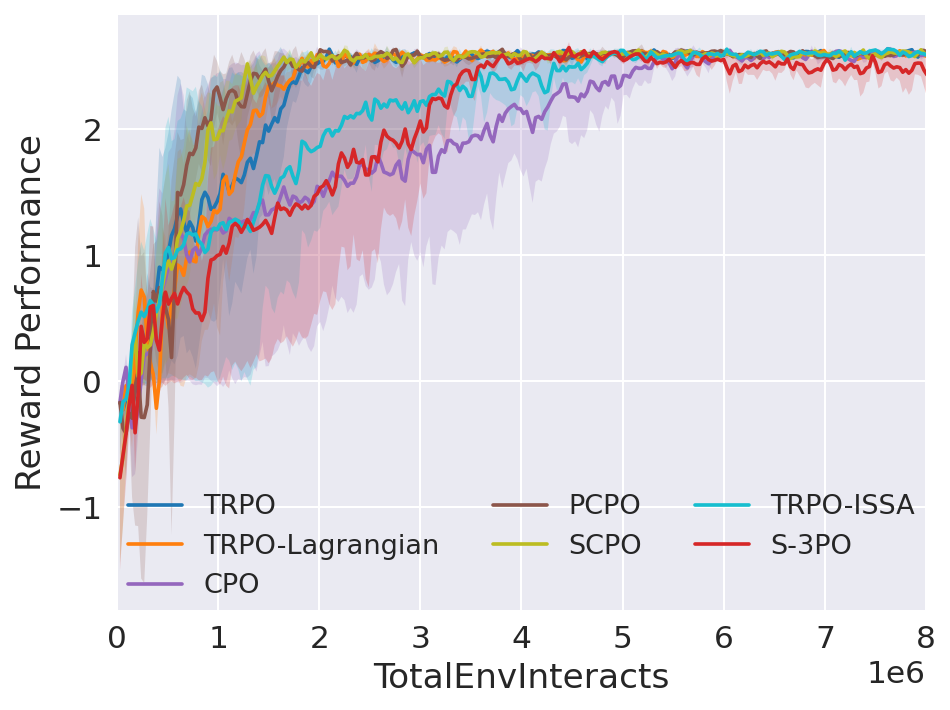} \\
            \includegraphics[width=0.31\textwidth]{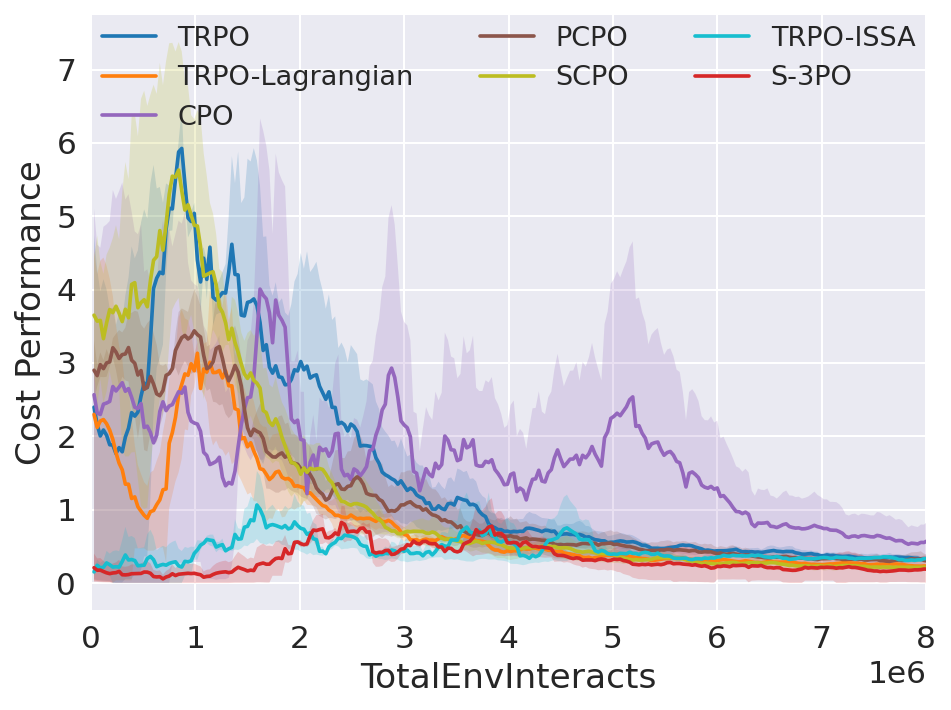} \\
            \includegraphics[width=0.31\textwidth]{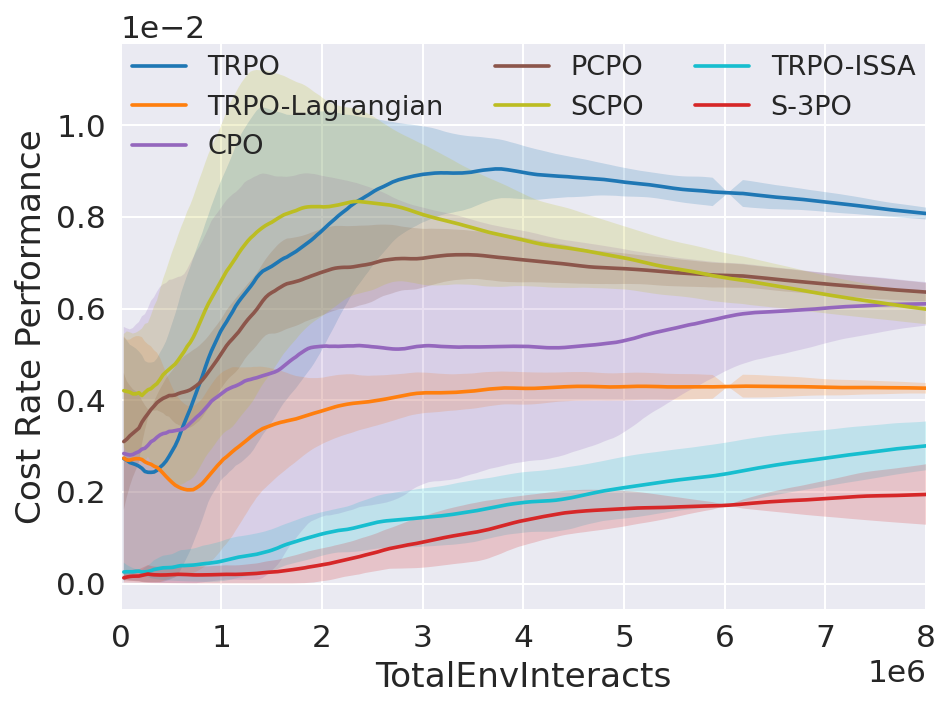}
        }
        \label{fig:walker-hazard-1-train}
    }
    \caption{Training performance of link robots}
    \label{fig:exp-link-robot-train}
\end{figure}

\begin{figure}
    \centering
    \subfigure[Swimmer\_1Hazard]{
        \centering
        \parbox{0.31\textwidth}{
            \includegraphics[width=0.31\textwidth]{fig/swimmertiny1_ISSA_Trigger_Times.png}
        }
        \label{fig:swimmer-hazard-1-ISSA}
    }
    \subfigure[Ant\_1Hazard]{
        \centering
        \parbox{0.31\textwidth}{
            \includegraphics[width=0.31\textwidth]{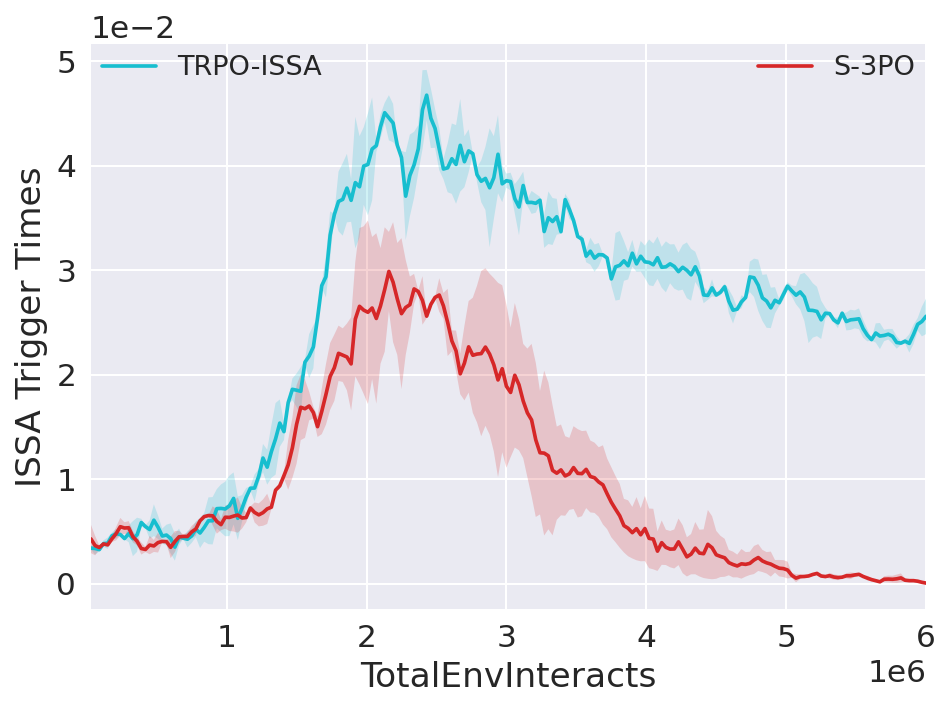}
        }
        \label{fig:ant-hazard-1-ISSA}
    }
    \subfigure[Walker\_1Hazard]{
        \centering
        \parbox{0.31\textwidth}{
            \includegraphics[width=0.31\textwidth]{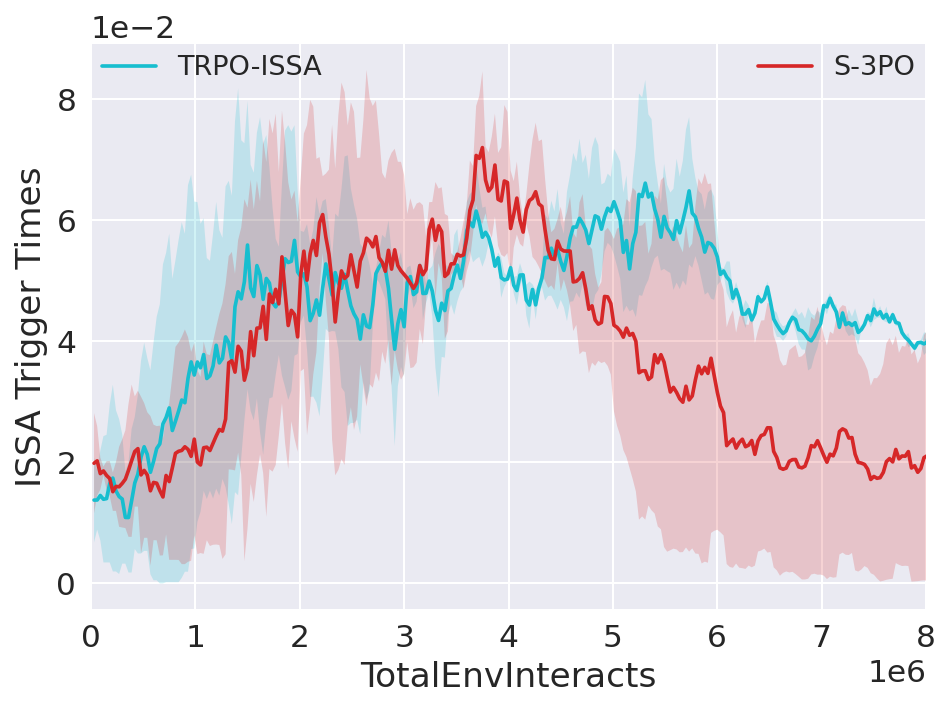}
        }
        \label{fig:walker-hazard-1-ISSA}
    }
    \caption{ISSA performance of link robots}
    \label{fig:exp-link-robot-ISSA}
\end{figure}

In \Cref{tab: point_hazard,tab: point_pillar,tab: link_robot_hazard}, we report all the $9$ results of our test suites by three metrics:
\begin{itemize}
    \item The average episode return $J_r$.
    \item The average episodic sum of costs $M_c$.
    \item The average cost over the entirety of training $\rho_c$.
\end{itemize}
Both the evaluation performance and training performance are reported based on the above metrics. Besides, we also report the ISSA triger times as ISSA performance of TRPO-ISSA and S-3PO. All of the metrics were obtained from the final epoch after convergence. Each metric was averaged over two random seed. The evaluation performance curves of all experiments are shown in \Cref{fig:exp-point-hazard-eval,fig:exp-point-pillar-eval,fig:exp-link-robot-eval}, the training performance curves of all experiments are shown in \Cref{fig:exp-point-hazard-train,fig:exp-point-pillar-train,fig:exp-link-robot-train} and the ISSA performance curves of all experiments are shown in \Cref{fig:exp-point-hazard-ISSA,fig:exp-point-pillar-ISSA,fig:exp-link-robot-ISSA}




\begin{table}[p]
\caption{Metrics of three \textbf{Point\_Hazard} environments obtained from the final epoch.}
\centering
\resizebox{!}{2cm}{
\begin{tabular}{c|ccc|ccc|c}
\toprule
\multicolumn{8}{c}{Point\_1Hazard} \\
\cline{1-8}\\[-1.02em]
\textbf{Algorithm} & \multicolumn{3}{c|}{Evaluation Performance} &\multicolumn{3}{c|}{Training Performance} &\multicolumn{1}{c}{ISSA Performance}\\
\cline{2-7}\\[-1.02em]
& $\bar{J}_r$ & $\bar{M}_c$ & \multicolumn{1}{c|}{$\bar{\rho}_c$} & $\bar{J}_r$ & $\bar{M}_c$ & \multicolumn{1}{c|}{$\bar{\rho}_c$}&\\
\cline{1-8}\\[-1.02em]
TRPO & 2.5738 & 0.5078 & 0.0082 & 2.5738 & 0.5078 & 0.0082 & -\\
TRPO-Lagrangian & \textbf{2.6313} & 0.5977 & 0.0058 & \textbf{2.6313} & 0.5977 & 0.0058 & -\\
CPO & 2.4988 & 0.1713 & 0.0045 & 2.4988 & 0.1713 & 0.0045 & -\\
PCPO & 2.4928 & 0.3765 & 0.0054 & 2.4928 & 0.3765 & 0.0054 & -\\
SCPO & 2.5457 & 0.0326 & 0.0022 & 2.5457 & 0.0326 & 0.0022 & -\\
TRPO-ISSA & 2.5113 & 0.0000 & 0.0000 & 2.5981 & 0.0000 & 0.0000 & 0.2714\\
S-3PO & 2.4157 & \textbf{0.0000} & \textbf{0.0000} & 2.2878 & \textbf{0.0000} & \textbf{0.0000} & \textbf{0.0285}\\

\bottomrule
\end{tabular}}
\vspace{20pt}

\resizebox{!}{2cm}{
\begin{tabular}{c|ccc|ccc|c}
\toprule
\multicolumn{8}{c}{Point\_4Hazard} \\
\cline{1-8}\\[-1.02em]
\textbf{Algorithm} & \multicolumn{3}{c|}{Evaluation Performance} &\multicolumn{3}{c|}{Training Performance} &\multicolumn{1}{c}{ISSA Performance}\\
\cline{2-7}\\[-1.02em]
& $\bar{J}_r$ & $\bar{M}_c$ & \multicolumn{1}{c|}{$\bar{\rho}_c$} & $\bar{J}_r$ & $\bar{M}_c$ & \multicolumn{1}{c|}{$\bar{\rho}_c$}&\\
\cline{1-8}\\[-1.02em]
TRPO & \textbf{2.6098} & 0.2619 & 0.0037 & \textbf{2.6098} & 0.2619 & 0.0037 & -\\
TRPO-Lagrangian & 2.5494 & 0.2108 & 0.0034 & 2.5494 & 0.2108 & 0.0034 & -\\
CPO & 2.5924 & 0.1654 & 0.0024 & 2.5924 & 0.1654 & 0.0024 & -\\
PCPO & 2.5575 & 0.1824 & 0.0025 & 2.5575 & 0.1824 & 0.0025 & -\\
SCPO & 2.5535 & 0.0523 & 0.0009 & 2.5535 & 0.0523 & 0.0009 & -\\
TRPO-ISSA & 2.5014 & 0.0712 & 0.0000 & 2.5977 & 0.0135 & 0.0000 & 0.1781\\
S-3PO & 2.3868 & \textbf{0.0000} & \textbf{0.0000} & 2.3550 & \textbf{0.0000} & \textbf{0.0000} & \textbf{0.0117}\\

\bottomrule
\end{tabular}}
\vspace{20pt}

\resizebox{!}{2cm}{
\begin{tabular}{c|ccc|ccc|c}
\toprule
\multicolumn{8}{c}{Point\_8Hazard} \\
\cline{1-8}\\[-1.02em]
\textbf{Algorithm} & \multicolumn{3}{c|}{Evaluation Performance} &\multicolumn{3}{c|}{Training Performance} &\multicolumn{1}{c}{ISSA Performance}\\
\cline{2-7}\\[-1.02em]
& $\bar{J}_r$ & $\bar{M}_c$ & \multicolumn{1}{c|}{$\bar{\rho}_c$} & $\bar{J}_r$ & $\bar{M}_c$ & \multicolumn{1}{c|}{$\bar{\rho}_c$}&\\
\cline{1-8}\\[-1.02em]
TRPO & 2.5535 & 0.5208 & 0.0074 & 2.5535 & 0.5208 & 0.0074 & -\\
TRPO-Lagrangian & 2.5851 & 0.5119 & 0.0064 & 2.5851 & 0.5119 & 0.0064 & -\\
CPO & \textbf{2.6440} & 0.2944 & 0.0041 & \textbf{2.6440} & 0.2944 & 0.0041 & -\\
PCPO & 2.6249 & 0.3843 & 0.0052 & 2.6249 & 0.3843 & 0.0052 & -\\
SCPO & 2.5126 & \textbf{0.0703} & 0.0020 & 2.5126 & 0.0703 & 0.0020 & -\\
TRPO-ISSA & 2.5862 & 0.0865 & 0.0000 & 2.5800 & 0.0152 & 0.0000 & 0.3431\\
S-3PO & 2.4207 & 0.1710 & \textbf{0.0000} & 2.3323 & \textbf{0.0000} & \textbf{0.0000} & \textbf{0.0337}\\

\bottomrule
\end{tabular}}

\label{tab: point_hazard}
\vspace{-5\baselineskip}
\end{table}

\begin{table}[p]
\caption{Metrics of three \textbf{Pillar\_Hazard} environments obtained from the final epoch.}
\centering
\resizebox{!}{2cm}{
\begin{tabular}{c|ccc|ccc|c}
\toprule
\multicolumn{8}{c}{Point\_1Pillar} \\
\cline{1-8}\\[-1.02em]
\textbf{Algorithm} & \multicolumn{3}{c|}{Evaluation Performance} &\multicolumn{3}{c|}{Training Performance} &\multicolumn{1}{c}{ISSA Performance}\\
\cline{2-7}\\[-1.02em]
& $\bar{J}_r$ & $\bar{M}_c$ & \multicolumn{1}{c|}{$\bar{\rho}_c$} & $\bar{J}_r$ & $\bar{M}_c$ & \multicolumn{1}{c|}{$\bar{\rho}_c$}&\\
\cline{1-8}\\[-1.02em]
TRPO & \textbf{2.6065} & 0.2414 & 0.0032 & \textbf{2.6065} & 0.2414 & 0.0032 & -\\
TRPO-Lagrangian & 2.5772 & 0.1218 & 0.0020 & 2.5772 & 0.1218 & 0.0020 & -\\
CPO & 2.5464 & 0.2342 & 0.0028 & 2.5464 & 0.2342 & 0.0028 & -\\
PCPO & 2.5857 & 0.2088 & 0.0025 & 2.5857 & 0.2088 & 0.0025 & -\\
SCPO & 2.5928 & 0.0040 & 0.0003 & 2.5928 & 0.0040 & 0.0003 & -\\
TRPO-ISSA & 2.5985 & 0.0000 & 0.0000 & 2.5909 & 0.0020 & 0.0000 & 0.3169\\
S-3PO & 2.5551 & \textbf{0.0000} & \textbf{0.0000} & 2.5241 & \textbf{0.0000} & \textbf{0.0000} & \textbf{0.0060}\\

\bottomrule
\end{tabular}}
\vspace{20pt}

\resizebox{!}{2cm}{
\begin{tabular}{c|ccc|ccc|c}
\toprule
\multicolumn{8}{c}{Point\_4Pillar} \\
\cline{1-8}\\[-1.02em]
\textbf{Algorithm} & \multicolumn{3}{c|}{Evaluation Performance} &\multicolumn{3}{c|}{Training Performance} &\multicolumn{1}{c}{ISSA Performance}\\
\cline{2-7}\\[-1.02em]
& $\bar{J}_r$ & $\bar{M}_c$ & \multicolumn{1}{c|}{$\bar{\rho}_c$} & $\bar{J}_r$ & $\bar{M}_c$ & \multicolumn{1}{c|}{$\bar{\rho}_c$}&\\
\cline{1-8}\\[-1.02em]
TRPO & 2.5671 & 0.4112 & 0.0063 & 2.5671 & 0.4112 & 0.0063 & -\\
TRPO-Lagrangian & \textbf{2.6040} & 0.2786 & 0.0050 & \textbf{2.6040} & 0.2786 & 0.0050 & -\\
CPO & 2.5720 & 0.5523 & 0.0062 & 2.5720 & 0.5523 & 0.0062 & -\\
PCPO & 2.5709 & 0.3240 & 0.0052 & 2.5709 & 0.3240 & 0.0052 & -\\
SCPO & 2.5367 & \textbf{0.0064} & 0.0005 & 2.5367 & 0.0064 & 0.0005 & -\\
TRPO-ISSA & 2.5739 & 0.1198 & 0.0001 & 2.5881 & 0.0427 & 0.0001 & 0.2039\\
S-3PO & 2.2513 & 0.0114 & \textbf{0.0000} & 2.3459 & \textbf{0.0000} & \textbf{0.0000} & \textbf{0.0116}\\

\bottomrule
\end{tabular}}
\vspace{20pt}

\resizebox{!}{2cm}{
\begin{tabular}{c|ccc|ccc|c}
\toprule
\multicolumn{8}{c}{Point\_8Pillar} \\
\cline{1-8}\\[-1.02em]
\textbf{Algorithm} & \multicolumn{3}{c|}{Evaluation Performance} &\multicolumn{3}{c|}{Training Performance} &\multicolumn{1}{c}{ISSA Performance}\\
\cline{2-7}\\[-1.02em]
& $\bar{J}_r$ & $\bar{M}_c$ & \multicolumn{1}{c|}{$\bar{\rho}_c$} & $\bar{J}_r$ & $\bar{M}_c$ & \multicolumn{1}{c|}{$\bar{\rho}_c$}&\\
\cline{1-8}\\[-1.02em]
TRPO & 2.6140 & 3.1552 & 0.0201 & 2.6140 & 3.1552 & 0.0201 & -\\
TRPO-Lagrangian & 2.6164 & 0.6632 & 0.0129 & 2.6164 & 0.6632 & 0.0129 & -\\
CPO & \textbf{2.6440} & 0.5655 & 0.0166 & \textbf{2.6440} & 0.5655 & 0.0166 & -\\
PCPO & 2.5704 & 6.6251 & 0.0219 & 2.5704 & 6.6251 & 0.0219 & -\\
SCPO & 2.4162 & 0.2589 & 0.0024 & 2.4162 & 0.2589 & 0.0024 & -\\
TRPO-ISSA & 2.6203 & 0.6910 & 0.0009 & 2.5921 & 0.0709 & 0.0009 & 0.3517\\
S-3PO & 2.0325 & \textbf{0.0147} & \textbf{0.0002} & 2.3371 & \textbf{0.0000} & \textbf{0.0002} & \textbf{0.0231}\\

\bottomrule
\end{tabular}}

\label{tab: point_pillar}
\vspace{-5\baselineskip}
\end{table}

\begin{table}[p]
\caption{Metrics of three link robots environments obtained from the final epoch.}
\centering
\resizebox{!}{2cm}{
\begin{tabular}{c|ccc|ccc|c}
\toprule
\multicolumn{8}{c}{Swimmer\_1Hazard} \\
\cline{1-8}\\[-1.02em]
\textbf{Algorithm} & \multicolumn{3}{c|}{Evaluation Performance} &\multicolumn{3}{c|}{Training Performance} &\multicolumn{1}{c}{ISSA Performance}\\
\cline{2-7}\\[-1.02em]
& $\bar{J}_r$ & $\bar{M}_c$ & \multicolumn{1}{c|}{$\bar{\rho}_c$} & $\bar{J}_r$ & $\bar{M}_c$ & \multicolumn{1}{c|}{$\bar{\rho}_c$}&\\
\cline{1-8}\\[-1.02em]
TRPO & \textbf{2.7049} & 0.3840 & 0.0076 & \textbf{2.7049} & 0.3840 & 0.0076 & -\\
TRPO-Lagrangian & 2.6154 & 0.3739 & 0.0060 & 2.6154 & 0.3739 & 0.0060 & -\\
CPO & 2.5817 & 0.3052 & 0.0056 & 2.5817 & 0.3052 & 0.0056 & -\\
PCPO & 2.5418 & 0.6243 & 0.0055 & 2.5418 & 0.6243 & 0.0055 & -\\
SCPO & 2.6432 & 0.3919 & 0.0051 & 2.6432 & 0.3919 & 0.0051 & -\\
TRPO-ISSA & 2.5826 & 0.2595 & 0.0000 & 2.5955 & \textbf{0.0000} & 0.0000 & 0.1240\\
S-3PO & 2.6032 & \textbf{0.0313} & \textbf{0.0000} & 2.6239 & 0.0001 & \textbf{0.0000} & \textbf{0.0378}\\

\bottomrule
\end{tabular}}
\vspace{20pt}

\resizebox{!}{2cm}{
\begin{tabular}{c|ccc|ccc|c}
\toprule
\multicolumn{8}{c}{Ant\_1Hazard} \\
\cline{1-8}\\[-1.02em]
\textbf{Algorithm} & \multicolumn{3}{c|}{Evaluation Performance} &\multicolumn{3}{c|}{Training Performance} &\multicolumn{1}{c}{ISSA Performance}\\
\cline{2-7}\\[-1.02em]
& $\bar{J}_r$ & $\bar{M}_c$ & \multicolumn{1}{c|}{$\bar{\rho}_c$} & $\bar{J}_r$ & $\bar{M}_c$ & \multicolumn{1}{c|}{$\bar{\rho}_c$}&\\
\cline{1-8}\\[-1.02em]
TRPO & 2.6390 & 0.3559 & 0.0074 & \textbf{2.6390} & 0.3559 & 0.0074 & -\\
TRPO-Lagrangian & 2.5866 & 0.2169 & 0.0044 & 2.5866 & 0.2169 & 0.0044 & -\\
CPO & 2.6175 & 0.2737 & 0.0072 & 2.6175 & 0.2737 & 0.0072 & -\\
PCPO & 2.6103 & 0.2289 & 0.0076 & 2.6103 & 0.2289 & 0.0076 & -\\
SCPO & 2.6341 & 0.2384 & 0.0065 & 2.6341 & 0.2384 & 0.0065 & -\\
TRPO-ISSA & \textbf{2.6509} & 0.3831 & 0.0032 & 2.6318 & 0.3516 & 0.0032 & 0.0279\\
S-3PO & 2.2047 & \textbf{0.0000} & \textbf{0.0002} & 2.2031 & \textbf{0.0000} & \textbf{0.0002} & \textbf{0.0001}\\

\bottomrule
\end{tabular}}
\label{tab: ant_hazard_1}
\vspace{20pt}

\resizebox{!}{2cm}{
\begin{tabular}{c|ccc|ccc|c}
\toprule
\multicolumn{8}{c}{Walker\_1Hazard} \\
\cline{1-8}\\[-1.02em]
\textbf{Algorithm} & \multicolumn{3}{c|}{Evaluation Performance} &\multicolumn{3}{c|}{Training Performance} &\multicolumn{1}{c}{ISSA Performance}\\
\cline{2-7}\\[-1.02em]
& $\bar{J}_r$ & $\bar{M}_c$ & \multicolumn{1}{c|}{$\bar{\rho}_c$} & $\bar{J}_r$ & $\bar{M}_c$ & \multicolumn{1}{c|}{$\bar{\rho}_c$}&\\
\cline{1-8}\\[-1.02em]
TRPO & 2.5812 & 0.2395 & 0.0075 & 2.5812 & 0.2395 & 0.0075 & -\\
TRPO-Lagrangian & 2.6227 & 0.1666 & 0.0041 & 2.6227 & 0.1666 & 0.0041 & -\\
CPO & 2.6035 & 0.3068 & 0.0062 & 2.6035 & 0.3068 & 0.0062 & -\\
PCPO & 2.5775 & 0.2414 & 0.0059 & 2.5775 & 0.2414 & 0.0059 & -\\
SCPO & 2.6352 & \textbf{0.1423} & 0.0051 & \textbf{2.6352} & \textbf{0.1423} & 0.0051 & -\\
TRPO-ISSA & \textbf{2.6419} & 0.3544 & 0.0037 & 2.5787 & 0.2060 & 0.0037 & 0.0316\\
S-3PO & 2.6117 & 0.3437 & \textbf{0.0025} & 2.6055 & 0.2665 & \textbf{0.0025} & \textbf{0.0319}\\

\bottomrule
\end{tabular}}

\label{tab: link_robot_hazard}
\vspace{-5\baselineskip}
\end{table}

\newpage
\clearpage
\vspace*{\fill}
\vspace*{\fill}

\vspace*{\fill}
\vspace*{\fill}
\clearpage

\vspace*{\fill}
\vspace*{\fill}
\clearpage

